\documentclass[onecolumn,journal]{IEEEtran}

\usepackage{arxiv}

\usepackage[T1]{fontenc}% optional T1 font encoding

\ifCLASSINFOpdf
\else
\fi
\usepackage{amsmath}
\usepackage{epsfig}%,endnotes}
\usepackage{amsthm,amsfonts}
\usepackage{amssymb,setspace,cite,color,graphicx}
\usepackage{subfigure}
\usepackage{mathtools}

\usepackage{algorithm}
\usepackage[noend]{algorithmic}

\allowdisplaybreaks[1]

\newtheorem{remark}{Remark}

\newtheorem{theorem}{Theorem}
\newtheorem{lemma}{Lemma}
\newtheorem{corollary}{Corollary}

\newtheorem{assumption}{Assumption}

\newtheorem*{proof*}{Proof}

\usepackage{bbm}
\usepackage{booktabs}       % professional-quality tables
\usepackage{microtype}
\usepackage{array,multirow}
\usepackage{diagbox}
\usepackage{enumitem}

\newcommand{\name}{\texttt{SparseSecAgg}}  % acronym
\newcommand{\namespace}{\name{ }}

\newcommand{\RNum}[1]{\uppercase\expandafter{\romannumeral #1\relax}}

\usepackage{wrapfig}

\newcounter{ALC@tempcntr}% Temporary counter for storage

\begin{document}

\title{Sparsified Secure Aggregation for Privacy-Preserving Federated Learning}

\author{\vspace{0.2in}\onehalfspacing\IEEEauthorblockN{Irem Erg{\"u}n $^1$\thanks{Equal contribution.}\quad  Hasin Us Sami $^2$$^*$ \quad 
Ba\c{s}ak~G{\"u}ler $^2$
} \vspace{0.6cm}\\
\IEEEauthorblockA{$^1$University of California, Riverside\\ Department of Computer Science and Engineering\\
Riverside, CA 92521 \\
{\em iergu001@ucr.edu}} \vspace{0.5cm}\\
\IEEEauthorblockA{$^2$University of California, Riverside\\
Department of Electrical and Computer Engineering\\
Riverside, CA 92521 \\
{\em  \quad hsami003@ucr.edu, bguler@ece.ucr.edu}}}

\maketitle
% \vspace{-0.6in}
\begin{abstract}
Secure aggregation is a popular protocol in privacy-preserving federated learning, which allows model aggregation without revealing the individual models in the clear. On the other hand, conventional secure aggregation protocols incur a significant communication overhead, which can become a major bottleneck in real-world bandwidth-limited applications. Towards addressing this challenge, in this work we propose a lightweight gradient sparsification framework for secure aggregation, in which the server learns the aggregate of the sparsified local model updates from a large number of users, but without learning the individual parameters. Our theoretical analysis demonstrates that the proposed framework can significantly reduce the communication overhead of secure aggregation while ensuring  comparable computational complexity. We further identify a trade-off between privacy and communication efficiency due to sparsification. Our experiments demonstrate that our framework reduces the communication overhead by up to $7.8\times$, while also speeding up the wall clock training time by $1.13\times$, when compared to conventional secure aggregation benchmarks.
\end{abstract}

\begin{IEEEkeywords}
Federated learning, secure aggregation, privacy-preserving distributed training, machine learning in mobile networks.
\end{IEEEkeywords}

% \singlespacing 

\section{Introduction}

Federated learning is a distributed training framework to train machine learning models over the data collected and stored at a large number of data owners (users) \cite{mcmahan2017communication}. 
Training is carried out through an iterative process coordinated by a central server, who maintains a global model. At each iteration, the server sends the current state of the global model to the users, who then update the global model by training the global model on their local datasets, creating a local model. The  local models are then aggregated by the server to update the global model for the next iteration. Finally, the updated global model is pushed back by the server to the users.

Due to its on-device learning architecture (data never leaves the device), federated learning is popular in a variety of privacy-sensitive applications, such as mobile keyboard  suggestions, remote healthcare, or product recommendations  \cite{recom, rieke2020future, xu2021federated, chen2020fedhealth, lim2020dynamic}. 
On the other hand, it has recently been shown that the local models still carry extensive information about the local datasets. In particular, a server who observes the local models in the clear can use model inversion techniques to reveal the sensitive training examples of the users  \cite{Fredrikson,nasr2019comprehensive,zhu2020deep,GeipingBD020}.

{\it Secure aggregation} protocols have emerged as a countermeasure against such privacy threats, by enabling the server to \emph{aggregate} the local models of a large number of users, without observing the local models in the clear  \cite{bonawitz2017practical,zhao2021information, JSAIT2, bell2020secure}. 
This is achieved by a process known as \emph{additive pairwise masking}, based on cryptographic secure multi-party computing (MPC, \cite{yao1982protocols})  principles. In this process, each pair of users agree on a pairwise random mask, and users mask their locally trained model by combining it with the pairwise random masks. Users then only share the \emph{masked local model} with the server, which obfuscates the true values of the local models from the server, in that the server can learn no information (in an information-theoretic sense) about the true values of the local models from the masked models.  
On the other hand, once the masked models are aggregated at the server, the pairwise masks cancel out,  allowing the server to learn the aggregate of the local models, but no further information is revealed about the local models beyond their sum. 
As such, secure aggregation provides an additional level of privacy by preventing the server from observing the local models. 
Moreover, secure aggregation is complementary to and can be combined with other privacy-preserving machine learning approaches such as differential privacy \cite{DP}, and can even benefit the latter by reducing the amount of noise required  to achieve a target privacy level  (hence improving model accuracy) \cite{jayaraman2018distributed}. As such, it has become a standard  protocol in privacy-preserving federated learning.

The major challenge against the scalability of secure aggregation protocols to large networks is their communication overhead. Conventional secure aggregation protocols require each user to send their entire model to the server, i.e., the size of the masked model is as large as the entire model, which can become a significant bottleneck in bandwidth-limited wireless environments. In conventional (non-private) federated learning, this is handled through various communication-reduction techniques such as gradient sparsification, 
% In conventional secure aggregation protocols, the dimension of the masked models is as large as the 
% Currently, there is a significant gap between the scalability of federated learning with and without secure aggregation. 
% The latter can scale to massive  networks through 
where instead of the entire model, each user only sends a few gradient (or model) parameters to the server \cite{aji2017sparse, lin2017deep, jiang2018linear, stich2018sparsified, wangni2017gradient}. 
The main sparsification techniques are random-K and top-K sparsification, where users select random or top K (in terms of the magnitude) values from their local gradients, and send the corresponding parameters along with the location indices to the server. The server then aggregates the parameters according to their locations, and updates the global model. 
Sparsification can provide substantial benefits in reducing the communication overhead in distributed learning, particularly in the bandwidth-limited wireless environments envisioned for federated learning, with minimal impact on convergence. 

These conventional gradient sparsification techniques, however, can not be applied to secure aggregation. 
This is due to the fact the coordinates of the sparsified gradient parameters often vary from one user to another, which prevents the pairwise masks from being cancelled out when the masked models are aggregated at the server. 
% As such, once the sparsified models are aggregated at the server, the pairwise masks do not cancel out. 
This in turn requires the server to learn the individual  pairwise masks to remove them from the aggregated model, which will breach user privacy, as learning the pairwise masks will reveal the local models to the server, violating a core principle of secure aggregation. Our goal is to address this challenge, in particular, we want to answer the following question, 
``\emph{How can one design a secure aggregation protocol with  gradient sparsification, where the server learns the aggregate of the sparsified local models from a large number of users, without observing them in the clear?}''.  

To address this challenge, in this work we introduce the first  secure aggregation framework with gradient sparsification, \name, that enables aggregating a fraction of $\alpha\in(0,1]$ random model parameters from each user, without learning the individual model parameters. 
To do so, we introduce a novel gradient sparsification process, termed \emph{pairwise sparsification}, where the sparsification pattern is determined via pairwise multiplicative random masks shared between each pair of users. 
Specifically, each pair of users agree on two types of random vectors, a pairwise binary multiplicative mask that  identifies the sparsification pattern, and a pairwise additive mask that hides the contents of the local models. 
Each user then locally constructs a \emph{sparsified masked model} according to the pattern specified by the pairwise binary multiplicative masks, and sends the masked model parameters and their locations (with respect to the global model) to the server. 
The proposed sparsification strategy ensures that once the sparsified masked models are aggregated at the server, the additive masks cancel out, allowing the server to learn the aggregate of the sparsified local models, but without learning their true values. 
By doing so, {\name} reduces the communication overhead of secure aggregation by having users send only a small fraction of their local models to the server at each training round. 

% , with comparable computational load at the server. 

% without additional computational 
% with comparable computational load. 
% to the server. 
% to ensure that the pairwise additive masks to cancel out after the masked models are aggregated.

In our theoretical analysis, we evaluate the performance of \namespace in terms of convergence, privacy, communication, and computational overhead, and formalize a trade-off between privacy and communication efficiency brought by sparsification. 
Specifically, stronger privacy guarantees can be achieved by increasing the number of model parameters sent from each user (hence the communication overhead). 
This in turn also increases the number of local models aggregated at the server and thus speeds up the training. For  training a model of size $d$ in a network  with $N$ users, where  up to $A \leq \gamma N  $ users are adversarial for some $\gamma\in (0, 1/2)$, with a user dropout rate of $\theta \in (0,0.5)$, 
we quantify this trade-off as $T = \alpha (1-\gamma)(1-\theta) N$, where parameter  $T$ denotes the number of honest users aggregated for any given parameter of the aggregated model, and quantifies the privacy guarantee. Larger $T$ leads to better privacy which, for standard secure aggregation is equal to $T = (1-\gamma)(1-\theta) N$ \cite{bonawitz2017practical}.  Parameter $\alpha$ quantifies the size of the sparsified models, i.e., the sparsified model of each user consists of $\alpha d$ model parameters on average, where $d$ is the total number of model parameters. A smaller $\alpha$ leads to a smaller communication overhead per user. 
In this work, our focus is on the honest-but-curious adversary setup, where the adversaries (including the server and/or the adversarial users) follow the protocol but may collude and try to learn additional sensitive information using the messages exchanged during the protocol. 
Finally, we demonstrate the theoretical convergence guarantees of \name.

In our numerical evaluations, we provide extensive experiments for image classification on the CIFAR-10 and MNIST datasets \cite{krizhevsky2009learning, xiao2017fashion}, in a network with up to $100$ users on the Amazon EC2 Cloud platform, to compare \namespace with conventional secure aggregation \cite{bonawitz2017practical} benchmarks. 
To reach the same level of test accuracy, we demonstrate that \namespace reduces the communication overhead by $7.8\times$ on the CIFAR-10 dataset and by $17.9\times$ on the MNIST dataset while also reducing the wall clock training time  compared to conventional secure aggregation.

In summary, this paper introduces a secure aggregation framework with gradient sparsification, \name, to tackle the communication bottleneck of privacy-preserving federated learning. \namespace allows the aggregation of the sparsified local models from a large number of users, without revealing their true values.  
Our specific contributions are as follows.

\begin{enumerate}
    \item We propose the first secure aggregation protocol, \name, that can leverage gradient sparsification. To do so, we introduce a novel sparsification process, where the sparsification pattern is determined by pairwise multiplicative random masks shared between the users.   
    
    \item We show that \namespace significantly reduces the communication overhead of secure aggregation, which is critical in bandwidth-limited wireless environments. 
    \item We identify the key performance metrics for privacy and communication overhead to quantify the impact of gradient sparsification on secure aggregation. 
    \item   We theoretically demonstrate a trade-off between privacy and communication efficiency. 
    Specifically, one can achieve stronger privacy 
    by increasing the communication overhead. 
    \item  
We perform extensive experiments for image classification in a network of up to $100$ users over the Amazon EC2 cloud, and  demonstrate up to $7.8\times$ reduction in the communication overhead over conventional secure aggregation. 
% and up to $6\times$ speed-up in the wall-clock training time over conventional secure aggregation. 
    
\end{enumerate}

\section{Related Work}\label{sec:related}

For conventional (non-private) federated learning \cite{mcmahan2017communication}, communication efficiency is primarily  achieved through gradient  sparsification, quantization, or compression techniques  \cite{jeon2020compressive, malekijoo2021fedzip, sattler2019robust, xu2020ternary, albasyoni2020optimal, sun2020adaptive}. 
Another line of work focuses on user selection to reduce the communication overhead of (non-private) federated learning, where at each iteration only a subset of users  participate in training \cite{li2019convergence,  cho2020client,chen2020optimal,cho2020bandit,ribero2020communication}. 
The user selection process can vary anywhere from random selection, where users are selected uniformly at random across the network, to selecting users according to how much they contribute to the training process, such as with respect to the magnitude of their gradient.   
Unlike our setup, in these works the selected users send the entire model to the server. In contrast, our focus is on reducing the  communication load per user, in particular,  the number of model parameters sent from each user, which can become a major bottleneck in emerging machine learning applications in bandwidth-limited wireless environments, where model sizes can be in the range of millions \cite{sandler2018mobilenetv2}.    
We remark that our approach is complementary to and can be combined with user sampling techniques, which is an interesting future direction.

For privacy-preserving federated learning, the  communication overhead is the major bottleneck against the scalability of secure aggregation protocols to large networks, which is in the order of  $O(N + d)$ per user, for training a model of size $d$ in a network of $N$ users \cite{bonawitz2017practical}. 
Addressing the communication overhead of secure aggregation has received significant attention in the recent years 
\cite{JSAIT2,bell2020secure}. 
Unlike our setup, these works assume that each user sends the entire model to the server, and focus on techniques that reduce the per-user  communication overhead with respect to the number of users $N$, in particular, from $O(N + d)$ to $O(\log N + d)$, by leveraging  circular \cite{JSAIT2} or graph-based  communication topologies \cite{bell2020secure}. 
Our technique is also complementary to and can be combined with these approaches. 

Another notable approach in privacy-preserving federated learning is leveraging differential privacy \cite{dwork2006calibrating, abadi2016deep, brendan2018learning, pmlr-v22-rajkumar12, pathak2010multiparty}. These approaches are based on a utility-privacy trade-off, by adding (irreversible) noise to the computations to protect the privacy of personally identifiable information (PII). The noise is calibrated to achieve a target privacy level. On the other hand, unlike secure aggregation (which is based on secure MPC principles), the additional noise is irreversible and thus may decrease the training performance. This leads to a privacy-utility trade-off, where higher noise levels increase the privacy but may also decrease the model performance. Secure aggregation protocols are complementary to differential privacy. In principle, the two can be combined to further improve the performance (model accuracy) of differential privacy protocols for federated learning, by reducing the amount of noise that needs to be added to reach a target differential privacy level \cite{jayaraman2018distributed}.

The rest of the paper is organized as follows. In Section~\ref{sec:problem}, we provide background on federated learning and secure aggregation. The system model is described in Section~\ref{sec:problemsetting}. Section~\ref{sec:sparsesecagg} introduces the \namespace framework. In Section~\ref{sec:theory}, we provide our theoretical analysis and convergence results. The experimental evaluations are demonstrated in Section~\ref{sec:experiments}.  Section~\ref{sec:Conclusion} concludes the paper. Throughout the paper, we use the following notation. $x$ represents a scalar variable, whereas $\mathbf{x}$ represents a vector. $\mathcal{X}$ refers to a set, and $[N]$ denotes the set $\{1, \ldots, N\}$.

\section{Background}\label{sec:problem}

\subsection{Federated Learning}\label{sec:FL}
Federated learning is a distributed framework for training machine learning models in mobile networks \cite{mcmahan2017communication}. The learning architecture  consists of a server and $N$ devices (users), where user $i\in[N]$ has a local dataset $\mathcal{D}_i$ with $|\mathcal{D}_i|$ data points. 
The goal is to train a model $\mathbf{w}\in\mathbb{R}^d$ of dimension $d$ to minimize a global loss function $F(\mathbf{w})$, 
\begin{equation}\label{eq:objective_fnc} 
    \min_{\mathbf{w}} F(\mathbf{w}) \text{ s.t.  } 
    F(\mathbf{w}) = \sum_{i\in[N]} \beta_i F_i (\mathbf{w}), 
\end{equation} 
where $F_i$ denotes the local loss function of user $i$ and $\beta_i$ is a weight parameter assigned to user $i$, often proportional to the size of the local datasets 
% \footnote{For ease of exposition, we present our framework for $D_i = \frac{D}{N}$ (equal-sized datasets) and note that it can be extended to the general case by locally scaling the local models without loss of generality.} 
$\beta_i = \frac{|\mathcal{D}_i|}{\sum_{i\in[N]} |\mathcal{D}_i|}$ \cite{FLSurvey}. 
% where $D = \sum_{i\in[N]} |\mathcal{D}_i|$ \cite{FLSurvey}. 
% $\beta_i = \frac{|\mathcal{D}_i|}{D}$ where $D = \sum_{i\in[N]} |\mathcal{D}_i|$ \cite{FLSurvey}. 

Training is carried out through an iterative process. At iteration $t$, the server sends the current state of the global model, represented by $\mathbf{w}^{(t)}$, to the users. 
User $i$ updates the global model by local training, where the global model is  updated on the local dataset through multiple stochastic gradient descent (SGD) steps, 
\begin{equation}
\mathbf{w}_i^{(t, j+1)} = \mathbf{w}_{i}^{(t, j)} - \eta^{(t, j)}  \nabla F_i(\mathbf{w}_i^{(t, j)}, \xi_i^{(t, j)})
\end{equation}
for $j=0,\ldots, E-1$, where $E$ is the number of local training steps, $\mathbf{w}_i^{(t,0)} \triangleq \mathbf{w}^{(t)}$ and $\eta^{(t, j)}$ is the learning rate.  $\nabla F_i(\mathbf{w}_i^{(t, j)}, \xi_i^{(t, j)})$ is the local gradient of user $i$ evaluated on a (uniformly) random sample (or a mini-batch of samples) $\xi_i^{(t, j)}$ from the local dataset $\mathcal{D}_i$. After $E$ local training steps, user $i$  forms a local model, 
\begin{align}
\mathbf{w}_i^{(t)} & \triangleq \mathbf{w}_i^{(t,E)} \\
& = \mathbf{w}^{(t)} -  \sum_{j=0}^{E-1} \eta^{(t, j)} \nabla F_i(\mathbf{w}_i^{(t, j)}, \xi_i^{(t, j)}) \label{eq:lgrad1}
\end{align}
and sends it to the server. 
Alternatively, instead of sending the local model $\mathbf{w}_i^{(t)}$, user $i$ can send the (weighted) local gradient: 
\begin{equation}\label{eq:lgrad2}
\mathbf{y}_i^{(t)} := \sum_{j=0}^{E-1} \eta^{(t, j)} \nabla F_i(\mathbf{w}_i^{(t, j)}, \xi_i^{(t, j)})
\end{equation}
to the server. The two approaches are equivalent since one can be obtained from the other. 

% {\color{blue} TO DO: Include the mathematical expression for local training. }

The local models are then aggregated by the server to update the global model,
\begin{align}
\mathbf{w}^{(t+1)} & \triangleq \sum_{i\in[N]} \beta_i \mathbf{w}_i^{(t)}  \label{aggregate} \\ 
& = \mathbf{w}^{(t)} -  \sum_{i\in[N]}  \beta_i \sum_{j=0}^{E-1}      \eta^{(t, j)} \nabla F_i(\mathbf{w}_i^{(t, j)}, \xi_i^{(t, j)}))  \\
& = \mathbf{w}^{(t)} - \sum_{i\in[N]} \beta_i \mathbf{y}_i^{(t)} 
\end{align}
after which the updated global model $\mathbf{w}^{(t+1)} $ is sent back to the users for the next iteration.

\begin{figure}[t]
\centering
\includegraphics[width=0.55\linewidth]{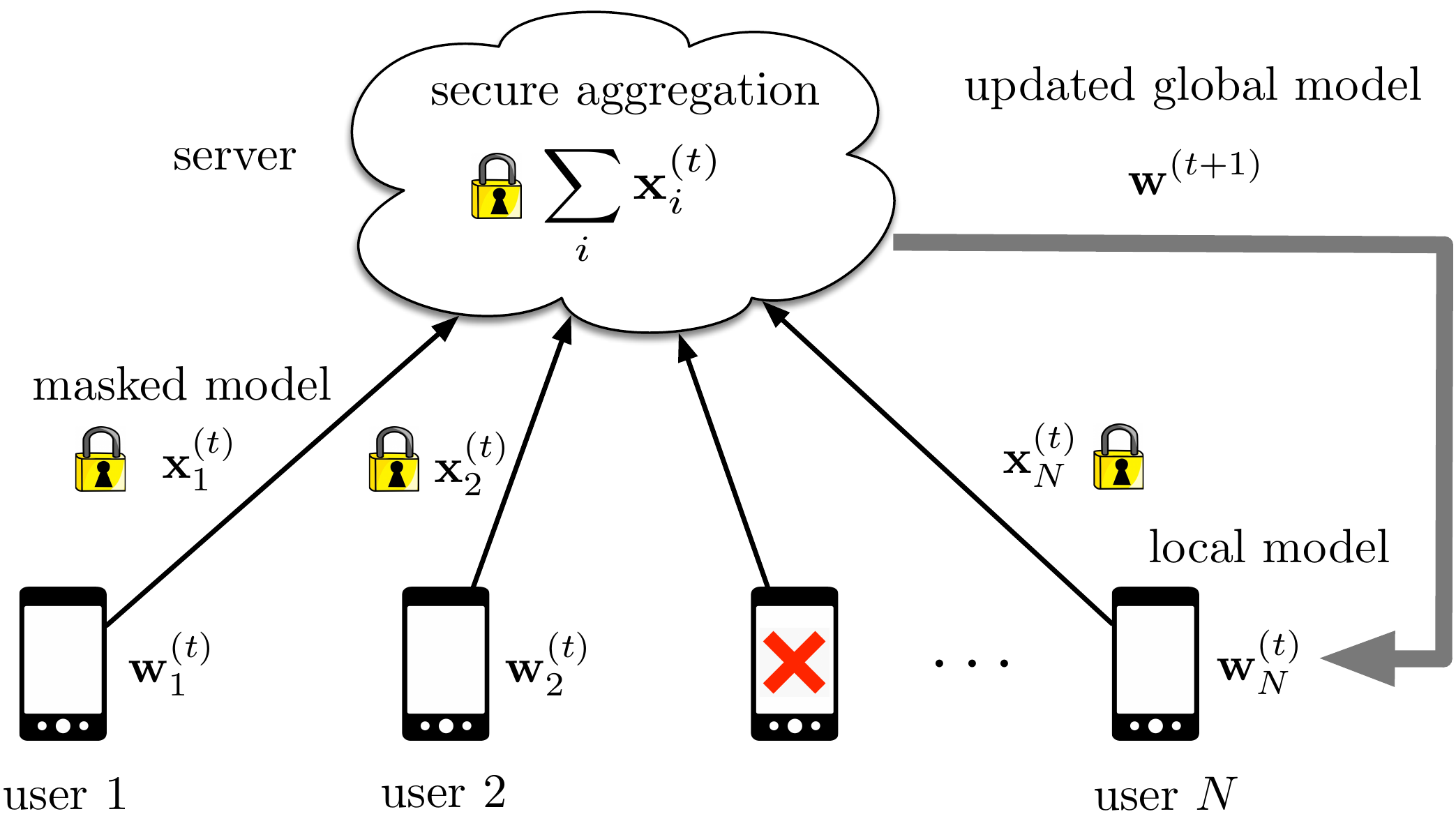} 
\caption{\footnotesize Secure aggregation in federated learning. At each iteration, the server sends the current state of the global model to the users, who then update it using their local datasets, and send a masked model to the server. 
The server aggregates the masked models and learns the sum of the true models, using which it updates the global model.}
\label{federated-original}
\vspace{-0.4cm}
\end{figure}

\subsection{Secure Aggregation}\label{Sec:secagg}
Secure aggregation is a popular protocol in privacy-preserving federated learning \cite{bonawitz2017practical}, which allows model aggregation from a large number of users without revealing the individual models in the clear. 
The goal is to enable the server to  compute the aggregate of the local models in \eqref{aggregate}, but \emph{without learning the individual local models}. 
This is achieved by a process known as \emph{additive masking} based on secure multi-party computing (MPC) principles \cite{evans2017pragmatic}, where each user masks its local model by using pairwise random keys before sending it to the server.  
The pairwise random keys are typically generated using a Diffie-Hellman type key exchange  protocol \cite{diffie1976new}. Using the pairwise keys, each pair of users $i,j\in[N]$ agree on a pairwise random seed $s_{ij}^{(t)}$. 
In addition to the pairwise seeds, user $i$ also creates a private random seed $s_i^{(t)}$, which protects the privacy of the user if a user delayed instead of being dropped, which was shown in  \cite{bonawitz2017practical}.

Using the pairwise and private seeds, user $i$ then creates a \emph{masked} version of its local model, 
\begin{equation}\label{pairwise}
\mathbf{x}_{i}^{(t)} = \mathbf{w}_i^{(t)}  +\text{PRG}(s_{i}^{(t)}) + \sum_{j:i<j} \text{PRG}(s_{ij}^{(t)}) - \sum_{j:i>j}  \text{PRG}(s_{ij}^{(t)})
\end{equation}
where PRG is a pseudorandom generator, and sends this masked model to the server. 
Finally, user $i$ secret shares $\{s_{ij}^{(t)}\}_{{j\in[N]}}$ and $s_i^{(t)}$ with every other user, using Shamir's $\frac{N}{2}$-out-of-$N$ secret sharing protocol  \cite{shamir1979share}. 
% Shamir's secret sharing embeds each seed in a random polynomial of degree $\frac{N}{2}$. The secret share sent to each user is an evaluation of that polynomial on a different point. This way, when the secret shares from at least $\frac{N}{2}+1$ users are combined, one can solve a system of linear equations that reveals the seed. However, any fewer than  $\frac{N}{2}+1$ secret shares reveals no information (in an information-theoretic sense) about the seed. 
All operations in \eqref{pairwise} are carried out in a finite field $\mathbb{F}_q$ of integers modulo a prime $q$. 
% The PRG generates uniformly random variables from $\mathbb{F}_q$. 

As secure aggregation protocols are primarily designed for wireless networks, some users may drop out from the system due to various reasons, such as poor wireless connectivity, low battery, or merely from a device being offline, and fail to send their masked model to the server. The set of dropout and surviving users at iteration $t$ are denoted by $\mathcal{D}^{(t)}$ and $\mathcal{S}^{(t)} = [N]\backslash \mathcal{D}^{(t)}$, respectively. 

In order to compute the aggregate of the user models, the server first aggregates the \emph{masked models} received from the surviving users, $\sum_{i\in \mathcal{S}^{(t)}} \mathbf{x}_i^{(t)}$. Then, the server collects the secret shares of the pairwise seeds belonging to the dropped users, and the secret shares of the private seeds belonging to the surviving users. Using the secret shares, the server reconstructs the pairwise and private seeds corresponding to dropped and surviving users, respectively, and removes them from the aggregate of the masked models, 
\begin{align}\notag
 &\sum_{i\in \mathcal{S}^{(t)}} \!\!(\mathbf{x}_i^{(t)} \!-\!   \text{PRG}(s_{i}^{(t)}) )  \!-\!\!\! \sum_{i\in\mathcal{D}^{(t)}} \!\!\big(\!\sum_{j:i<j} \text{PRG}(s_{ij}^{(t)}) \!-\!\!\! \sum_{j:i>j} \text{PRG}(s_{ji}^{(t)}) \big) \notag \\
 & \qquad \qquad = \sum_{i\in\mathcal{S}^{(t)}} \mathbf{w}_i^{(t)}
\end{align} 
after which all of the random masks cancel out and the server learns the sum of the true local models of all surviving users. 
Figure~\ref{federated-original} demonstrates this process.

\section{Problem Formulation}\label{sec:problemsetting}

The communication overhead of sending the local models in \eqref{aggregate} from the users to the server poses a major challenge in large-scale applications, where $N$ and  $d$ can be in the order of millions. 
Gradient sparsification is a recent approach introduced to address this challenge, where instead of sending the entire model (or gradient, whose size is equal to the model), users only send a few ($K\ll d$) gradient parameters to the server  \cite{aji2017sparse, lin2017deep, jiang2018linear, stich2018sparsified, wangni2017gradient}. 
The most common sparsification techniques are rand-$K$ and top-$K$ gradient sparsification, where users select either random or top (with respect to the magnitude) $K$ values from their local gradient, and send the corresponding gradient parameters along with their locations (coordinates) to  server. 
The server then updates the global model by only using the few gradient parameters sent from the users. 

% The communication overhead of sending the masked models in \eqref{pairwise} from the users to the server poses a major challenge in large-scale applications, where $N$ and  $d$ can be in the order of millions. 
% Gradient sparsification is a recent approach to reduce the communication overhead of distributed learning, where instead of sending the entire model, users only send $K\ll d$ model parameters to the server  \cite{aji2017sparse, lin2017deep, jiang2018linear, stich2018sparsified, wangni2017gradient}. 
% The most common sparsification techniques are random $K$ and top $K$ sparsification, where users select either random or top (with respect to the magnitude) $K$ values from their local gradient, and send the corresponding model parameters along with their locations (coordinates) to  server. 

%%%%%%%%%%%%%%%%%%%%
\begin{figure} %[t!]
\centering
    \subfigure[MNIST IID]{\label{fig:overlap_IID}
    \includegraphics[width=0.35\linewidth]{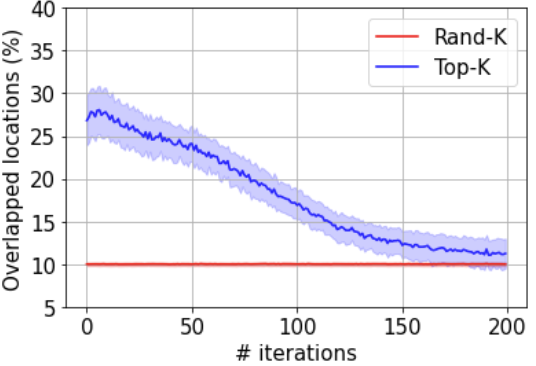}
    % \vspace{-20 pt}
    }
    \subfigure[MNIST nonIID]{\label{fig:overlap_NonIID}
    \includegraphics[width=0.35\linewidth]{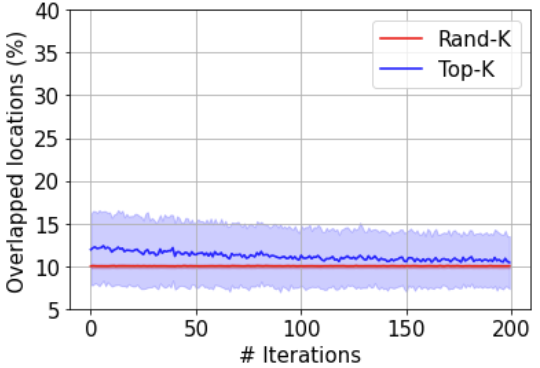}
    % \vspace{-20 pt}
    }
     
\vspace{-6 pt}

\caption{Average percentage of overlappping gradient locations between each pair of users for rand-K and top-K sparsification, respectively. Training is done for an image classification task on the MNIST dataset  with $N=30$ users and $K = \frac{d}{10}$. The IID and non-IID data distributions are implemented according to \cite{mcmahan2017communication}. 
} 
\label{fig:overlap}
\vspace{-0.2cm}\end{figure}

Gradient sparsification has become popular in reducing the communication load in distributed training due to its practicality and substantial bandwidth efficiency. 
However, none of these sparsification techniques can be applied with secure aggregation, as the locations of the $K$ parameters often differ from one user to another. 
We demonstrate this phenomenon in Fig.~\ref{fig:overlap}, where we implement  federated learning (from Section~\ref{sec:FL}) with rand-K and top-K gradient sparsification, for an image classification task on the MNIST dataset with $N=30$ users, where $K = \frac{d}{10}$. We then measure the percentage of overlapping gradient locations (coordinates) between each pair of users for rand-K and top-K gradient sparsification, respectively, and report the average across all users. The shaded areas represent one standard deviation from the mean. We consider both the  IID (independent identically distributed) and non-IID data distribution settings across the users, as given in  \cite{mcmahan2017communication}.

For rand-$K$ sparsification, in both IID and non-IID settings, only around $10\%$ of the gradient locations overlap on average, between each pair of users. This is consistent with the theoretical expectation, where the expected number of overlapping locations is $\frac{K}{d}$ as each user selects $K$ gradient locations uniformly random from $d$ locations, independently from other users.  For top-k sparsification, in the IID setting, 
only around $30\%$ of the gradient locations overlap at the initial round of training between each pair of users. 
As the training progresses, the overlap decreases to around $10\%$. This effect is even more severe in the non-IID setting, where the average overlap is further reduced to around $12\%$ throughout training.

As a result, if gradient sparsification is naively applied with a secure aggregation protocol (described in Section~\ref{Sec:secagg}), the pairwise masks will not cancel out, requiring the server to reconstruct all of the pairwise masks. 
% {\color{red} TO DO: Add figure that shows the average overlap is small between the users (mean and variance), one plot for rand k and another for top k (can be on the same figure), can use MNIST non-iid, N=25,50,75,100.} 
This in turn will lead to a substantial communication and computational overhead. More importantly, doing so will allow the server to remove the random masks from each masked model in \eqref{pairwise}, revealing the individual local models to the server, violating the core principle of secure aggregation.

%%%%%%%%%%%%%%%%%%%%

Towards addressing this challenge, in this work we introduce \emph{sparsified secure aggregation}, where the server learns the aggregate of \emph{sparsified local models} from a large number of users, without learning the individual model parameters. We consider a network with $N$ users where user $i$ holds a local model $\mathbf{w}_i^{(t)}$ of dimension $d$. The goal is to reduce the communication overhead of secure aggregation by aggregating, instead of the entire model, only a small fraction of the local model parameters from each user, while ensuring provable convergence guarantees for training and protecting the privacy of individual users.

\vspace{0.1cm}
\noindent
{\bf Threat model. } Our focus is on an honest-but-curious adversary model (also known as passive adversaries), where adversarial parties follow the protocol truthfully, but try to infer  privacy-sensitive information using the messages exchanged throughout the protocol. We assume that out of $N$ users, up to $A \leq \gamma N$ users are adversarial for some $\gamma\in(0, 1/2)$, who may collude with each other and/or the server to learn the local models of honest users.

\vspace{0.1cm}
\noindent
{\bf Key performance metrics. }
We evaluate the performance of a sparsified secure aggregation protocol according to the following key parameters:
\begin{enumerate} 
    \item \emph{Privacy:} 
    The privacy guarantee, $T$, quantifies the number of honest users whose local model updates  are aggregated at a given location of the global model, with probability approaching to $1$ as $N\rightarrow \infty$. A higher value of $T$ represents better privacy, that is, even if the adversaries collude with each other and/or the server, they can only learn the \emph{sum} of local updates from $T$ users, and no further information (in an information-theoretic sense) is revealed beyond that. 
    % they cannot learn any further information (in an information-theoretic sense) than the sum of the local models. 
    % the most meaningful information that they can extract is the aggregate (sum) of $T$ local models. 
    For the conventional secure aggregation protocol described in Section~\ref{Sec:secagg}, $T=N$, as the entire local model is aggregated from each user. 
    % , since the local models from all $N$ users are aggregated at any given location of the global model. 

    \item \emph{Compression ratio: } The compression ratio $\alpha\in(0,1]$ is defined as the fraction of the (masked) parameters sent from each user to the server (each user sends $\alpha d$ masked parameters, as opposed to the entire model of size $d$), with probability approaching $1$ as $d\rightarrow \infty$. 
    As $\alpha$ becomes smaller ($\alpha \ll 1$), users send fewer parameters to the server (as opposed to the entire model), which reduces the communication overhead. On the other hand, a smaller $\alpha$ may also increase the number of training iterations required to reach a target training accuracy. 

%     \item \emph{Communication and computation overhead:} 
%   The communication/computation overhead  refers to the asymptotic time complexity (runtime) of  communication/computation to aggregate the local models, with respect to the number of users and  parameters sent from each user. 

    \item \emph{Computation overhead:} 
   The computation overhead  refers to the asymptotic time complexity (runtime) of  computation to aggregate the local model updates, with respect to the number of users and  parameters sent from each user. For efficiency, the computation overhead should be comparable (in the same order) to conventional secure aggregation. 
   
    \item \emph{Robustness to user dropouts:} 
% Users may drop out from the network due to various reasons such as poor wireless connectivity or low battery.  
We assume that each user may drop out from the system with probability  $\theta\in(0,0.5)$, independent from other users, which we refer to as the user dropout rate.
% We assume that up to $\theta N$ users may drop out from the system, where $\theta\in[0,0.5)$ is the user dropout rate. 
In real-world settings, the dropout rate often varies between $0.06$ and $0.1$ \cite{bonawitz2019towards}. The robustness guarantee is the maximum user dropout rate that a protocol can tolerate beyond which the aggregate of the local model updates cannot be computed correctly. 
\end{enumerate}

In this work, we present the first sparsified secure aggregation protocol, \name, towards addressing the communication bottleneck of secure aggregation. 
Our framework consists of the following key components:
\begin{enumerate} 
    
    % \item \emph{Generation of random masks:} 
    \item \emph{Random mask generation:} 
    Each pair of users initially agree on two random seeds. The random seeds are then used for generating two pairwise random masks \footnote{Here, we utilize two different uses of the word \textit{mask}: The first one is the cryptographic definition, where the masks are used to hide the input, and the second one is the signal processing definition, where the masks are used to filter the input.}. 
    The first one is a pairwise additive mask, where each element is generated uniformly random from a finite field $\mathbb{F}_q$ of integers modulo a prime $q$. This mask is used for hiding the true values of the local model updates as in \eqref{pairwise}, before sending them to the server. The second one is a pairwise multiplicative mask, where each element is generated IID from a Bernoulli distribution. This mask is used to construct the sparsification pattern, and allows each pair of users to agree on a random subset of gradient locations for sparsification. Specifically, the additive masks ensure that the privacy of the local model updates are protected, while the multiplicative masks ensure that the additive masks cancel out once the sparsified gradients are aggregated. 
    % In the sequel, the former is called the \emph{pairwise binary mask}, whereas the latter is called the \emph{pairwise additive mask}, respectively. 
    
    % The first mask is a binary vector to be used for sparsification, whose coordinates are indicators of whether or not the corresponding gradient locations are selected to be sent to the server. 
    % The second mask is for hiding the gradients in an additive fashion as in \eqref{pairwise}, before sending the gradients to the server, to ensure the privacy of the local models. In the sequel, the former is called the \emph{pairwise binary mask}, whereas the latter is called the \emph{pairwise additive mask}, respectively. 

    \item \emph{Local quantization and scaling:} 
    % To ensure privacy, secure aggregation operations should be carried out 
    % To ensure privacy, secure aggregation operations are bound to finite field operations. This requires users to convert the model parameters from the domain of real numbers to a finite field. 
    % In this phase, users locally scale their model parameters and convert them from the domain of real numbers to a finite field. 
    Secure aggregation operations are bound to finite field operations, which requires the local updates to be converted from the domain of real numbers to a finite field. To do so, we leverage stochastic quantization. 
    At each iteration $t$, users $i\in[N]$ 
    % After training the model on their local dataset for $E$ local iterations, each user $i,\ i\in[N]$ 
    first scale their local gradients with respect to a scaling factor  $\frac{\beta_i}{p(1-\theta)}$, where $p=1-\left(1-\frac{\alpha}{N-1}\right)^{N-1}$. 
    % is the probability of a user selecting a given model parameter during sparsification. 
    % This scaling ensures theoretical convergence guarantees of training as we demonstrate in Section~\ref{sec:theory}. 
    % Secure aggregation operations are bound to finite field operations, which requires the model parameters to be converted from the domain of real numbers to a finite field. 
    Then, users quantize their local gradients to convert them from the domain of real numbers to the finite field $\mathbb{F}_q$. 
    % To do so, we utilize stochastic quantization. 
    % Then, users quantize their scaled model parameters via stochastic quantization. 
    % Then, they embed the model parameters in a finite field $\mathbb{F}_q$ of integers modulo a prime $q$ and add the masks who are also generated from $\mathbb{F}_q$ to the model parameters. 
    % Upon receiving the masked model updates, the server converts these parameters to the real domain before averaging. 
    The scaling factor and the stochasticity of the quantization scheme  are key components for our convergence guarantees, which we detail in Section~\ref{sec:theory}. 
    
%   \item \emph{Construction of sparsified models:}
%   \item \emph{Constructing the sparsified models:}
  \item \emph{Sparsified gradient  construction:}
% \item \emph{Masking sparsified models:} 
   In this phase, each user locally sparsifies its local gradient by leveraging the  pairwise multiplicative masks. 
   Then, each user picks the coordinates of the pairwise additive masks in accordance with the locations from the pairwise multiplicative masks and adds the corresponding values to the sparsified gradient, in order to hide the true values of the local gradient  parameters.
   Finally, each user sends the sparsified masked gradient and the corresponding parameter locations to the server.

%   \item \emph{Secure aggregation of masked models:} 

%   \item \emph{Secure aggregation of sparsified models:} 
%   \item \emph{Model aggregation and handling user dropouts:} 
 \item \emph{Secure Aggregation of Sparsified Gradients:}
     Upon receiving the sparsified masked gradients and the corresponding parameter locations, the server aggregates the sparsified masked gradients  according to the specified locations. If a user drops out before sending their update, the pairwise masks corresponding to those users  will not be cancelled out upon aggregation. 
   To handle this, the server requests the secret shares of the seeds corresponding to the dropout users who did not send their masked updates, reconstructs the pairwise masks corresponding to those users, and removes them from the aggregated gradients. 
\end{enumerate}

\section{The SparseSecAgg Framework} \label{sec:sparsesecagg}
% \label{sec:Proposed}

We now present the details of the \namespace framework. 
The process is described for one training iteration, and we omit the iteration index $t$ for ease of exposition.

\subsection{Random Mask Generation} \label{sec:masks}

\noindent
{\bf Pairwise and private additive masks. } 
\namespace leverages additive masks to hide the true content of the local model updates from the server 
during aggregation. For the generation of pairwise additive masks, each pair of users $i, j\in [N]$ agree on a pairwise secret seed $s_{ij}$ (unknown to other users and the server) by utilizing the Diffie-Hellman key exchange protocol \cite{diffie1976new}, which is then used as an input to a PRG to expand it to a random vector 
\begin{equation}\label{eq:pairwise}
\mathbf{r}_{ij} = \text{PRG}(s_{ij})
\end{equation}
of size $d$, where each element is generated uniformly at random from the finite field $\mathbb{F}_q$. 
% \noindent
% {\bf Private additive masks. }  
In addition, user $i\in[N]$ also generates a private mask 
\begin{equation}\label{eq:private}
\mathbf{r}_i := \text{PRG}(s_i)
\end{equation}
as in \eqref{pairwise}, by creating a private random seed $s_i$ and expanding it using a PRG into a random vector of dimension $d$, where each element is generated uniformly at random from $\mathbb{F}_q$. 
% In addition to the pairwise additive masks, user $i\in[N]$ also generates a private mask $\mathbf{r}_i := \text{PRG}(s_i)$ as in \eqref{pairwise}, by creating a private random seed $s_i$ and expanding it using a PRG into a random vector of dimension $d$, where each element is selected uniformly at random from $\mathbb{F}_q$. 
% As described in Section~\ref{sec:problem}, the private mask protects the privacy of the users who  are delayed (instead of dropped) and  are late in sending their masked models to the server, in which case the pairwise masks are not sufficient for privacy. 

\vspace{0.2cm}
\noindent
{\bf Pairwise multiplicative masks. } 
\namespace utilizes pairwise multiplicative masks to sparsify the local gradients. 
For this, each pair of users $i,j\in[N]$ agree on a binary vector $\mathbf{b}_{ij}\in\{0,1\}^d$, where each element $\ell\in[d]$ is generated from an IID Bernoulli random variable 
\begin{equation}\label{eq:pairwisemask}
\mathbf{b}_{ij}(\ell) = \left \{ \begin{matrix} 1 & \text{with probability} & \frac{\alpha}{N-1}\\
0 & \text{otherwise} & \end{matrix}\right .
\end{equation} 
% where the success probability $p$ is set as, 
% \begin{equation}\label{eq:success}
% p := \frac{\alpha}{N}, 
% \end{equation}
for a given $\alpha\in (0,1]$. 
Parameter $\alpha$ controls the number of parameters sent from each user (sparsity) and accordingly the communication overhead. 

For the generation of the binary vectors,   
the first step is to run another instantiation of the process described above for pairwise additive mask generation, where a vector of size $d$ is generated uniformly random from the field $\mathbb{F}_q$. 
Then, the domain of the PRG is divided into two intervals, where the size of the intervals are proportional to $\frac{\alpha}{N-1}$ and $1-\frac{\alpha}{N-1}$, respectively.  
By doing so, each pair of users $i, j\in[N]$ can agree on a binary vector   $\mathbf{b}_{ij}=\mathbf{b}_{ji}\in \{0,1\}^d$. 
The multiplicative masks (binary vectors) indicate the coordinates of which parameters are sent from each user to the server, and ensure that the additive masks cancel out once the sparsified gradients are aggregated. 

\vspace{0.2cm}
\noindent
{\bf Secret sharing. }
Finally, users secret share the seed of their additive and multiplicative masks with the other users, using Shamir's $\frac{N}{2}$-out-of-$N$ secret sharing \cite{shamir1979share}, where  each seed is embedded into $N$ secret shares, by embedding the random seed (secret) in a random polynomial of degree $\frac{N}{2}$ in $\mathbb{F}_q$. 

The secret sharing process ensures that each seed can be reconstructed from any $\frac{N}{2}+1$ shares, but any set of at most $\frac{N}{2}$ shares reveals no information (in an information-theoretic sense) about the seed. 
This ensures that the server can compute the aggregate of the local gradients even if up to $\frac{N}{2}-1$ users drop out from the network, as we describe in Section~\ref{sec:dropout}. 

\subsection{Local Quantization and Scaling}
% In this phase, user $i \in[N]$ first scales their local model $\mathbf{w}_i$ as $\frac{\beta_i}{\tilde{p}}\mathbf{w}_i$, where $\beta_i = \frac{D_i}{D}$ is as given in \eqref{eq:objective_fnc},  
% and  
% \begin{equation}
%     \label{tildep}
%     \tilde{p}\triangleq (1-(1-p)^{N-1})
% \end{equation} 
% is the probability of a model parameter being selected by user $i\in[N]$, which we demonstrate in Section~\ref{sec:theory}. 
% As we show in Section~\ref{sec:theory}, this scaling operation is critical for our  convergence guarantees of training, by ensuring the unbiasedness of the model aggregation process using the sparsified gradients. 

In this phase, users quantize their model updates to convert them from the domain of real numbers to the finite field $\mathbb{F}_q$. 
However, quantization should be performed carefully in order to ensure the  convergence of training. Moreover, the quantization should allow computations involving negative numbers in the finite field. 
% As a result, we cannot use conventional quantization techniques that represent the sign of a real number seperately from its magnitude \cite{alistarh2017qsgd}. 
We address this challenge by a scaled stochastic quantization approach as follows. 
% Doing so ensures the unbiasedness of the quantized gradients, which is required for our convergence guarantees detailed in Section~\ref{sec:theory}.

% \cite{so2021codedprivateml, so2021turbo, BREA}

% which takes as input a real number $z\in\mathbb{R}$, scales it with respect to an integer parameter $c>1$, and maps it to one of the two closest integers with a probability that is inversely proportional to the corresponding distance. 
% % Formally, the stochastic rounding function $Q_c:\mathbb{R}\rightarrow \mathbb{R}$ is defined as follows: 
% Formally, the stochastic rounding function is defined as follows: 

% The quantization process is carried out in two steps. The first step consists of a 

First, we define a scaling factor $\frac{\beta_i}{p(1-\theta)}$, where $\beta_i =  \frac{|\mathcal{D}_i|}{\sum_{i\in[N]} |\mathcal{D}_i|}$ as given in \eqref{eq:objective_fnc},  
and  
\begin{equation}
    \label{tildep}
    p\triangleq 1-\Big(1-\frac{\alpha}{N-1}\Big)^{N-1}
\end{equation} 
is the probability of a model parameter being selected by user $i\in[N]$, which we demonstrate in Section~\ref{sec:theory}. As we detail in our theoretical analysis, this scaling factor is critical for our  convergence guarantees of training, by ensuring the unbiasedness of the aggregation process using the sparsified gradients. 
Next, define a stochastic rounding function, 
\begin{equation}
   Q_c(z)= \begin{cases} 
      \dfrac{\lfloor cz \rfloor}{c} & \text{with probability } 1- (cz-\lfloor cz \rfloor)\\
      \\
      
      \dfrac{\lfloor cz \rfloor +1}{c} & \text{with probability } cz-\lfloor cz \rfloor
   \end{cases}
\end{equation}
where $\lfloor z \rfloor$ is the largest integer that is less than or equal to $z$ and the parameter $c$ is a tuning parameter that identifies the quantization level, similar rounding functions are also used in \cite{BREA, JSAIT1}. Note that $E_Q[Q_c(z)] = z$, hence the rounding process is unbiased. Utilizing a larger $c$ reduces the variance in quantization, leading to a more stable training and faster convergence. 
% Similar rounding functions are used in \cite{so2021codedprivateml, so2021turbo, BREA}. 
% We refer the reader to \cite{BREA} for a more detailed discussion about the impact of the parameter $c$ on training performance.
% Next, users multiply each quantized element with $c$ to remove the scaling.
% Finally, in the second step of the quantization process, the gradients are converted to two's complement representation to be able to represent negative numbers in a finite field. In summary, the quantized local model of user $i$ is defined as follows:

% Then, the quantized gradients are mapped to the field field $\mathbb{F}_q$, 
Then, user $i$ forms a quantized local gradient as follows,
% To be able to represent negative numbers in a finite field, two's complement representation In summary, the quantized local model of user $i$ is defined as follows:
\begin{equation}
\label{eqquant}
    \overline{\mathbf{y}}_i =\phi\Big(c\cdot Q_c\Big(\frac{\beta_i}{p(1-\theta)}\cdot \mathbf{y}_i\Big)\Big),
\end{equation}
where 
% $\mathbf{w}$ is the current state of the global model and $\mathbf{y}_i$ is the local gradient of user $i$, as shown in \eqref{eq:lgrad1} and \eqref{eq:lgrad2}, respectively, and 
the function  $\phi:\mathbb{R}\rightarrow\mathbb{F}_q$ is given by, 
% is a mapping function which embeds real numbers in the finite field $\mathbb{F}_q$, 
\begin{equation}
    \phi(z)= \begin{cases} 
     z & \text{if } z\geq0\\
      q+z & \text{if } z<0
   \end{cases}
\end{equation} 
to represent the positive and negative numbers using the first and second half of the finite field, respectively. 
% , which is also known as two's complement representation. 
Functions $Q_c(.)$ and $\phi(.)$ are applied element-wise in \eqref{eqquant}. 

\vspace{0.1cm}
\subsection{Sparsified Gradient Construction}\label{subsec:sparsified}
Using the additive and multiplicative masks, user $i\in[N]$ constructs a sparsified masked gradient  
$\mathbf{x}_i$ where the $\ell^{th}$ element is given by, 
\begin{align}
\label{construct-model}
&\mathbf{x}_i  (\ell) =  \Big (1-\prod_{j\in[N]: j\neq i} (1-\mathbf{b}_{ij}(\ell))\Big)(\overline{\mathbf{y}}_i (\ell) + \mathbf{r}_i (\ell))\notag \\
& \quad \quad \qquad + \sum_{j\in[N]:i< j} \mathbf{b}_{ij}(\ell) \mathbf{r}_{ij}(\ell) -  \sum_{j\in[N]:i >  j} \mathbf{b}_{ij}(\ell) \mathbf{r}_{ij}(\ell) 
\end{align} 
for $\ell\in [d]$. 
More specifically, 
for each non-zero element in $\mathbf{b}_{ij}$ for a given $j\in[N]$, user $i$ adds the corresponding element from $\mathbf{r}_{ij}$ to  its quantized local gradient $\overline{\mathbf{y}}_i$ if $i<j$, and subtracts it if $i>j$. 
The key property of this  process is to  ensure that once the sparsified masked gradients are aggregated at the server, the pairwise additive masks cancel out.

For each user $i\in[N]$, define a set $\mathcal{U}_i$ such that, 
\begin{equation}
\label{eq:locs}
\mathcal{U}_i = \{\ell: \mathbf{b}_{ij}(\ell)=1 \text{ for some } j\in[N], \ell\in[d]\}, 
\end{equation}
which contains the indices of the  gradient parameters to be sent from user $i$ to the server.  

User $i$ then sends all $\mathbf{x}_i(\ell)$ for which $\ell\in\mathcal{U}_i$, along with a vector holding the location indices $\ell\in\mathcal{U}_i$,  to the server. 
Sending the location information allows the server to reconstruct the sparsified masked gradient  $\mathbf{x}_i$.

\subsection{Secure Aggregation of Sparsified Gradients}\label{sec:dropout} 
Next, the server aggregates the sparsified masked gradients, 
\begin{equation}\label{eq:aggregate}
\mathbf{x} := \sum_{i\in\mathcal{S}} \mathbf{x}_i = \sum_{i\in[N]\backslash \mathcal{D}} \mathbf{x}_i
\end{equation}
where $\mathcal{D}$ is the set of users who dropped from the protocol and failed to send their masked gradients to the server, and $\mathcal{S} = [N]\backslash \mathcal{D}$ is the set of surviving users.   

Note that the pairwise masks corresponding to the dropout users as well as the private masks of the surviving users will not be cancelled out during the aggregation in \eqref{eq:aggregate}. 
To handle this, the server requests (from the surviving users), the secret shares of the pairwise seeds corresponding to the dropout users, and the private seeds  corresponding to the surviving users. 

Upon receiving a sufficient number of secret shares, the server reconstructs the corresponding random masks, and removes them from the aggregated gradients according to the locations specified by the location vector, 
\begin{align} 
    \overline{\mathbf{y}}(\ell) &= \mathbf{x}(\ell)  -  
    \sum_{i\in[N]\backslash \mathcal{D}} \mathbf{r}_i(\ell) \mathbbm{1}_i^{\ell}  
    -  \sum_{i\in\mathcal{D}} \sum_{\substack{j: i<j\\ j\in [N]\backslash\mathcal{D}}} \mathbf{r}_{ij}(\ell) \mathbbm{1}_i^{\ell} 
    +  \sum_{i\in\mathcal{D}} \sum_{\substack{j: i>j\\ j\in [N]\backslash\mathcal{D}}} \mathbf{r}_{ij}(\ell) \mathbbm{1}_i^{\ell} \label{eq:recon} \\
    & = \sum_{i\in\mathcal{S}} \overline{\mathbf{y}}_i(\ell) \label{eq:sumq}
    % \sum_{i\in\mathcal{U}_3}\mathbf{y}_i(\ell)\cdot\mathbbm{1}_i^{\ell}-\sum_{i\in\mathcal{U}_3}\mathbf{p}_i(\ell)\cdot\mathbbm{1}_i^{\ell}\\&+\sum_{i\in \mathcal{U}_3, j\in \mathcal{U}_2\setminus \mathcal{U}_3} \mathbf{s}_{ij}(\ell)\cdot \mathbf{b}_{ij}(\ell),
\end{align} 
where $\mathbbm{1}_i^{\ell}$ is an indicator random variable that is equal to 1 if and only if $\ell\subseteq \mathcal{U}_i$, and 0 otherwise. 

The sum of the quantized gradients from \eqref{eq:sumq} are then mapped back from the finite field to the real domain, 
% via  $\phi^{-1}:\mathbb{F}_q\rightarrow \mathbb{R}$, 
% \begin{equation}
% \label{eq:dequantfunc}
%     \phi^{-1}(\overline{x})= \begin{cases} 
%      \overline{x} & \text{if } 0\leq\overline{x}<\frac{q-1}{2}\\
%       \overline{x}-q & \text{if } \frac{q-1}{2}\leq\overline{x}<q
%       \end{cases}
% \end{equation}
% Specifically, the server computes 
\begin{equation}
\label{eq:dequant}
   \mathbf{w}\leftarrow \mathbf{w} - \dfrac{1}{c}\cdot{\phi^{-1}(\overline{\mathbf{y}})},
\end{equation}
where $\phi^{-1}$ is applied element-wise to $\overline{\mathbf{y}}$. 
% and scaling %with $\frac{1}{f}$ 
% are applied to $\overline{\mathbf{w}}$ in an element-wise manner. 

Finally, the updated global model $\textbf{w}$ is sent back from the server to the users for the next training iteration. 
% This concludes our protocol.  
The individual steps of our protocol are  demonstrated in Algorithm~\ref{AlgSparseSecAgg}. 

We note that in contrast to conventional secure aggregation which aggregates the \emph{local models} as described in Section~\ref{Sec:secagg}, the aggregation rule in \namespace  aggregates the \emph{local gradients}. This is to ensure  the formal convergence guarantees of our sparsified aggregation protocol as we detail in our theoretical analysis. We note, however, that in practice one can obtain the (aggregated)  local model from the (aggregated) local gradient and vice versa, and hence the two are complementary. As such, in the sequel, we refer to the aggregated local models and local gradients interchangeably when there is no ambiguity.

\begin{algorithm}[t]
\small
	\caption{\!Sparsified Secure Aggregation \!(\name)\!} \label{AlgSparseSecAgg} 
	 \textbf{Input:} Number of users N, local gradients $\mathbf{y}_i$ of users  $i\in[N]$, model size $d$, compression ratio $\alpha$, finite field $\mathbb{F}_q$.  \\
 \textbf{Output:} Aggregate of the local gradients   $\sum_{i\in\mathcal{S}}\mathbf{y}_i$ of all surviving users $\mathcal{S} = [N]\backslash\mathcal{D}$. \\
	\begin{algorithmic}[1] 
\vspace{-0.2cm}
\FOR {User $i=1,2,\ldots, N$ in parallel} 
\STATE Quantize the local gradient $\mathbf{y}_i$ according to \eqref{eqquant} to create the quantized gradient  $\overline{\mathbf{y}}_i$. 
\STATE Generate the private additive mask $\mathbf{r}_{i}$ from the finite field $\mathbb{F}_q$ according to  \eqref{eq:private}.  
% \STATE Generate the required keys for pairwise mask generation and encryption
\FOR {$j=1,2,\ldots, N\setminus\{i\}$}
\STATE Users $i$ and $j$ generate the pairwise additive mask $\mathbf{r}_{ij}$ from $\mathbb{F}_q$ according to \eqref{eq:pairwise}. 
% using the Diffie-Hellman key agreement protocol.  
% \STATE User $i$ generates the pairwise encryption keys for user $i$ and $j$, $e_{ij}$  using Diffie-Hellman key agreement 
\STATE  Users $i$ and $j$ generate the pairwise multiplicative mask  $\mathbf{b}_{ij}$ according to \eqref{eq:pairwisemask}. 
% , and then applying thresholding with respect to the value of $p$
% \STATE Each element of $\mathbf{v}_{ij}$ is subjected to thresholding such that it is converted to 1 with probability $\frac{\alpha}{N}$ and to 0 with probability $1-\frac{\alpha}{N}$. The output is the pairwise binary mask, $\mathbf{b}_{ij}$. 
\ENDFOR
% \STATE Secret share the random seeds for the pairwise and private masks $\{\mathbf{b}_{ij}\}_{j\in[N], j\neq i}$, $\{\mathbf{r}_{ij}\}_{j\in[N], j\neq i}$, and $\mathbf{r}_i$ with users $j\in[N]\backslash\{i\}$.
\STATE Secret share the random seeds for the pairwise and private masks with users $j\in[N]\backslash\{i\}$.
\STATE Construct the sparsified masked gradient  $\mathbf{x}_i$ according to \eqref{construct-model} and the corresponding locations $\mathcal{U}_i$ with respect to \eqref{eq:locs}
\STATE Send  $\{\mathbf{x}_i(\ell)\}_{\ell\in\mathcal{U}_i}$, along with the vector holding the location information $\mathcal{U}_i$ to the server. 
\ENDFOR
% re-write these later as function calls
\STATE Server aggregates the masked gradients received from the surviving users according to \eqref{eq:aggregate}. 
% Upon receiving the masked models from surviving users, 
% \STATE Server requests the secret shares of the private masks of the surviving users and the pairwise masks of the dropped users.
% \STATE Server requests the secret shares of the private masks of the surviving users and the pairwise masks of the dropped users. Using the secret shares, the server generates and removes the corresponding masks from the aggregate of the masked models as in \eqref{eq:recon}. 
\STATE Server removes the private masks of the surviving users and the pairwise masks of the dropped users from the aggregate of the masked gradients as in \eqref{eq:recon}. 
% \STATE Server requests from all surviving users, the secret shares of the private masks of the surviving users and the pairwise masks of the dropped users. 
% \STATE Using the secret shares, the server generates and removes the corresponding masks from the aggregate of the masked models as in \eqref{eq:recon}. 
% and computes the aggregate of the masked models, as in \eqref{eq:recon}.
% \STATE Using the secret shares, the server recovers the appropriate masks and computes the aggregate of the masked models, as in \eqref{eq:recon}.
\STATE Server converts the aggregated gradients from the finite field $\mathbb{F}_q$ to the real domain and updates the global model as in \eqref{eq:dequant}. 
% , and obtains the sum of the quantized local model updates of the surviving users.
% \STATE Server dequantizes the sum of the quantized local model updates of the users, as in \eqref{eq:dequant}.
	\end{algorithmic} 
\end{algorithm}

\section{Theoretical Analysis}\label{sec:theory}
% We now provide the theoretical performance guarantees of \namespace. 
We now provide our theoretical performance guarantees. 

\subsection{Key Performance Metrics}

\begin{theorem}[Compression ratio]\label{thm:sparsity}
% Consider a network with $N$ users 
% Let $p = \frac{K}{dN}$, where $N$ is the number of users and $d$ is the model size. Then, 
\namespace achieves a compression ratio of $\alpha$ with probability approaching to $1$ as the model size $d\rightarrow\infty$. 
% Then, \namespace achieves a sparsity guarantee of $k\rightarrow$ with probability approaching to $1$ as the model size $d\rightarrow$.  
% Then, \namespace achieves a sparsity guarantee of $K\rightarrow$ with probability approaching to $1$ as the model size $d\rightarrow$.  
% To train a model with dimension $d$ in a network with $N$ users, \namespace can achieve a sparsity guarantee of $$ with probability approaching to $1$ as the model size $d\rightarrow$. 
\end{theorem}
\begin{proof}
The proof is presented in Appendix~\ref{app:sparsity}. 
\end{proof}
Theorem~\ref{thm:sparsity} states that the number of parameters sent from each user is reduced from $d$ to $\alpha d$.

\begin{theorem}[Privacy]\label{thm:privacy} 
In a network with $N$ users with up to 
$A< \gamma N$ adversarial users where $\gamma\in(0,0.5)$ and a  dropout rate $\theta\in[0,0.5)$, \namespace achieves a privacy guarantee of $T = (1-e^{-\alpha}) (1-\theta) (1-\gamma)N$ with probability approaching to $1$ as the number of users $N\rightarrow\infty$. For $\alpha\ll 1$, the privacy guarantee approaches $T = \alpha (1-\theta)(1-\gamma)N$.   
% $T = (1-e^{\alpha}) (1-\gamma)N$ users, which approaches $T = \alpha (1-\gamma)N$ for $\alpha\ll 1$. 
% \namespace achieves a privacy guarantee of $T = \frac{K}{d}  (N-A)$ with probability approaching to $1$ as the number of users $N\rightarrow\infty$. 
% \namespace achieves a privacy guarantee of $T = \frac{KN}{2(d+K)}$ with probability approaching to $1$ as the number of users $N\rightarrow\infty$. 
\end{theorem}
% \begin{theorem}[Privacy]\label{thm:privacy} 
% \namespace achieves a privacy guarantee of $T =  \Big (\frac{K}{d+K} \Big )  (1-\alpha) N$ with probability approaching to $1$ as the number of users $N\rightarrow\infty$. 
% % \namespace achieves a privacy guarantee of $T = \frac{KN}{2(d+K)}$ with probability approaching to $1$ as the number of users $N\rightarrow\infty$. 
% \end{theorem}
\begin{proof}
The proof is provided in Appendix~\ref{app:privacy}. 
\end{proof}
% \begin{remark}
% Theorem~\ref{thm:privacy} states that the global model will consist of the aggregate (sum) of the local model parameters from at least $T = (1-e^{\alpha}) (1-\gamma)N$ users, and approaches $T = \alpha (1-\gamma)N$ for $\alpha\ll 1$. 
Theorem~\ref{thm:privacy} states that the global model will consist of the aggregate (sum) of the local model parameters from at least $T = \alpha (1-\theta) (1-\gamma)N$ honest users for $\alpha\ll 1$. 
% Theorem~\ref{thm:privacy} states that the global model will consist of the aggregate (sum) of the local model parameters from at least $T = \alpha  (1-\gamma)N$ users. 

% As the compression ratio $\alpha$ approaches $1$, the privacy guarantee approaches $T = (1-\gamma)N$, which is equal to the privacy guarantee of conventional secure aggregation \cite{bonawitz2017practical}, where the entire local models of all users are aggregated in the global model. 
% As $T$ grows, the server  
% \end{remark}
\begin{corollary}[Communication-privacy trade-off]
% \namespace demonstrates a 
% The sparsification of the local models demonstrate a trade-off between communication-efficiency and privacy. 
Theorem~\ref{thm:privacy} demonstrates a trade-off between privacy and communication efficiency provided by \name.  
In particular, for a sufficiently large number of users, as the compression ratio $\alpha$ increases, the privacy guarantee $T$ increases. On the other hand, a larger $\alpha$ also leads to a larger number of parameters sent from each user, thus increasing the communication overhead. 
% each user sends a larger fraction of their local models. This then increases  
\end{corollary}

\begin{theorem}[Computational Overhead] \label{thm:comp}
% The asymptotic computational overhead of \namespace is $O(dN^2)$, which is the same as conventional secure aggregation \cite{bonawitz2017practical}. 
% The asymptotic computational overhead of \namespace is $O(dN^2)$, which is the same as conventional secure aggregation \cite{bonawitz2017practical}. 
The asymptotic computational overhead of \namespace is $O(dN^2)$ for the server and $O(N+d)$ for each user, which is the same as conventional secure aggregation  \cite{bonawitz2017practical}.  
% The computation overhead of \namespace is $O(dN^2)$, the same as conventional secure aggregation. 
\end{theorem}
\begin{proof}
The only additional computations incurred in \namespace as compared to \cite{bonawitz2017practical} are due to the creation and reconstruction of the  pairwise binary multiplicative masks from \eqref{eq:pairwisemask}, whose impact on the overall computational overhead is a constant multiplicative factor of $2$. 
\end{proof}
\begin{remark}
Theorem \ref{thm:comp} states that \namespace can significantly improve the communication efficiency of secure aggregation, while ensuring comparable computational efficiency. 
\end{remark}

\begin{corollary}[Robustness against the user dropouts]
% \namespace is robust to the dropout of up to $\frac{N}{2}$ out of $N$ users. 
\namespace is robust to up to a dropout rate of $\theta < 0.5$ as $N\rightarrow \infty$. 
\end{corollary}
\begin{proof}
This immediately follows from Shamir's $\frac{N}{2}$-out-of-$N$ secret sharing \cite{shamir1979share}, which ensures that as long as there are at least $\frac{N}{2}+1$ surviving users,  
the server can collect the secret shares to reconstruct the pairwise seeds of the dropout users and the private seeds of the surviving users, reconstruct the corresponding masks, and remove them from the aggregated gradients as in \eqref{eq:recon}. Finally, with a dropout rate $\theta < 0.5$, the number of surviving users approaches $\frac{N}{2}+1$ as $N\rightarrow \infty$.  \end{proof}
\subsection{Convergence Analysis}

Finally, we provide the convergence guarantees of \name. We first state a few common technical assumptions  \cite{Stich2019localsgd,li2019convergence,Guler2021multiround} that are needed for our analysis.

\begin{assumption} \label{assumpt:1}
 $F_{1}, \cdots, F_{N}$ are all L -smooth, i.e., for all $\mathbf{v}$ and  $\mathbf{w}$, 
 \begin{equation}\label{eqassumpt:1}
F_{i}(\mathbf{v}) \leq F_{i}(\mathbf{w})+(\mathbf{v}-\mathbf{w})^{T} \nabla F_{i}(\mathbf{w})+\frac{L}{2}\|\mathbf{v}-\mathbf{w}\|_{2}^{2}, \quad \forall i \in [N]
 \end{equation}
\end{assumption}

\begin{assumption}\label{assumpt:2}
$F_{1}, \cdots, F_{N}$ are all $\mu$ -strongly convex. For all $\mathbf{v}$ and $\mathbf{w}$,
 \begin{equation}\label{eqassumpt:2}
 F_{i}(\mathbf{v}) \geq F_{i}(\mathbf{w})+(\mathbf{v}- \mathbf{w})^{T} \nabla F_{i}(\mathbf{w})+\frac{\mu}{2}\|\mathbf{v}-\mathbf{w}\|_{2}^{2}, \quad \forall i \in [N]
 \end{equation}
%  $\forall i \in [N]$.
\end{assumption}

\begin{assumption}\label{assumpt:3}
Let $\xi_{i}^{(t)}$ be a uniformly random sample from the local dataset of user $i$. Then, the variance of the stochastic gradients is bounded, 
\begin{equation}\label{eqassumpt:3}
\mathbb{E}\left\|\nabla F_{i}\left(\mathbf{w}_{i}^{(t)}, \xi_{i}^{(t)}\right)-\nabla F_{i}\left(\mathbf{w}_{i}^{(t)}\right)\right\|^{2} \leq \sigma_{i}^{2}, \quad \forall i \in [N]
\end{equation}
%  $\forall i \in [N]$.
\end{assumption}

\begin{assumption}\label{assumpt:4}
The expected squared norm of stochastic gradients is uniformly bounded, i.e., 
\begin{equation}\label{eqassumpt:4}
\mathbb{E}\left\|\nabla F_{i}\left(\mathbf{w}_{i}^{(t)}, \xi_{i}^{(t)}\right)\right\|^{2} \leq G^{2}, \quad \forall i \in [N]
\end{equation}
for $t=0, \cdots, J-1$ where $J$ is the total number of training rounds.
\end{assumption}
Let $\mathbf{w}^* := \arg \min_{\mathbf{w}} F(\mathbf{w})$ be   
the minimizer of \eqref{eq:objective_fnc}, i.e., the optimal value for the global model, and 
\begin{equation}
\Gamma=F^{*}-\sum_{i=1}^{N} \beta_{i} F_{i}^{*} \geq 0, 
\end{equation}
represent the divergence between global and local loss functions, where  $F^* := \min_{\mathbf{w}} F(\mathbf{w})$ and $F_{i}^{*} := \min_{\mathbf{w}} F_i(\mathbf{w})$. 
% represent the minimum of the global and local loss functions, respectively,   

\begin{theorem}[Convergence Guarantee]\label{thm:convergence}
Let \eqref{eqassumpt:1}~-~\eqref{eqassumpt:4} hold. Define $\nu=\max\left\{8\frac{L}{\mu},E\right\}$ and learning rate, $\eta^{(t)}=\frac{2}{\mu (\nu+t)}.$ 
Then,
\begin{equation}
\mathbb{E}\left[F\left(\mathbf{w}^{(J)}\right)\right]-F^{*} \leq \frac{2 \frac{L}{\mu}}{\nu+J}\left(\frac{B+C}{\mu}+2 L\left\|\mathbf{w}^{(0)}-\mathbf{w}^{*}\right\|^{2}\right)
\end{equation}
% where $\mathbf{w}^* := \arg \min_{\mathbf{w}} F(\mathbf{w})$ the 
% is the minimizer of \eqref{eq:objective_fnc}, i.e., the optimal value for the global model,  
% \begin{equation}
% \Gamma=F^{*}-\sum_{i=1}^{N} \beta_{i} F_{i}^{*} \geq 0, 
% \end{equation}
% measures the divergence between global and local loss functions, where  $F^* = \min_{\mathbf{w}} F(\mathbf{w})$ and $F_{i}^{*} = \min_{\mathbf{w}} F_i(\mathbf{w})$ represent the minimum of global and local loss functions, respectively,   
where
\begin{align}
B& =\sum_{i=1}^{N} \beta_{i}^{2} \sigma_{i}^{2}+6 L \Gamma+8(E-1)^{2} G^{2}
\end{align}
and
\begin{align}
%C& = \frac{d}{4c^2}+ \sum_{i=1}^{N} \left\{ \beta_{i}^2  + \sum_{j=1,j \neq i}^{N}  \beta_{i}\beta_{j} \right\}\left(\frac{1}{p}-1\right)4E^2G^2.
&C=\frac{1}{(\eta^{(t)})^{2}}\frac{Nd(1-\theta)p}{4c^2} + %\sum_{i=1}^{N} \left\{ \beta_{i}^2  + \sum_{j=1,j \neq i}^{N}  \beta_{i}\beta_{j} \right\}%
4E^2G^2\sum_{i=1}^{N} \left( \beta_{i}^2 \left(\frac{1}{(1-\theta)p}-1\right) \notag \right.\\
&\left.\hspace{2.5cm}+ \sum_{j=1,j \neq i}^{N}  \beta_{i}\beta_{j}\left(\frac{\Tilde{p}}{\left((1-\theta)p\right)^{2}}-1\right) \right)
\end{align}
such that $\Tilde{p}=(1-\theta)^{2}\left(1-2\left(1-\frac{\alpha}{N-1}\right)^{N-1}\!+\!\left(1-\frac{\alpha}{N-1}\right)^{2N-3}\right)$. 
% $\Tilde{p}$ is the probability that any two users $i,j \in [N]$ will participate together in a certain model coordinate $\ell \in [d]$ which is given as follows:
% \begin{align}
% \Tilde{p}=(1-\theta)^{2}\left(1-2\left(1-\frac{\alpha}{N-1}\right)^{N-1}+\left(1-\frac{\alpha}{N-1}\right)^{2N-3}\right) \notag 
% \end{align}

% with $p^{\prime}=(1-\theta)p$. 
% \begin{align}
% p^{\prime}=(1-\theta)p
% \end{align}
% and 
% \begin{equation}
% C= \frac{d}{4f^2}+ \sum_{i=1}^{N} \left\{ \beta_{i}^2  + \sum_{j=1,j \neq i}^{N}  \beta_{i}\beta_{j} \right\}\left(\frac{1}{p}-1\right)4E^2G^2.
% \end{equation}
\end{theorem}
\begin{proof}
The proof is presented in Appendix~\ref{app:convergence}. 
\end{proof}
\begin{remark}
Theorem~\ref{thm:convergence} states that the convergence rate is governed by  $p$, the probability of selecting any given location during the sparsification process, as defined in \eqref{tildep} and the dropout rate $\theta$. As $(1-\theta)p$ approaches $1$, $\left(\frac{1}{(1-\theta)p}-1\right)$ approaches $0$, speeding up the convergence. 
% Theorem~\ref{thm:convergence} states that the convergence rate is governed by the probability of contributing to a certain location $p$. With $p$ approaching to 1, $\left(\frac{1}{p}-1\right)$ approaches to $0$ and thus speeds up the convergence. 
\end{remark}
% \subsection{Convergence}
% \input{05_02_Convergence}

\section{Experiments} \label{sec:experiments}
 \begin{table}[t]
% \footnotesize
\small
% \centering
\caption{ Communication overhead per user per round on CIFAR-10. 
\label{table:bandwidth}}
\begin{center}
  \begin{tabular}{ |c | c | c|}
  
    \hline
    
                 \backslashbox{ N}{Protocol}
                  &\begin{tabular}{@{}c@{}}\texttt{SecAgg}\end{tabular}  & \name \\\hline
  25 &  0.66 MB & 0.08 MB\\ \hline
  50 & 0.66 MB & 0.082 MB\\ \hline
  75 & 0.66 MB & 0.083 MB\\ \hline
  100 & 0.66 MB & 0.083 MB\\ \hline
  \end{tabular}
  \end{center}
  \vspace{5 pt}
\end{table}
\begin{figure*}[t]
\centering
  \subfigure[Total communication overhead to reach the target test accuracy  ($55\%$).]{\label{fig:iid_cifar_communication}
   \includegraphics[width=.31\textwidth]{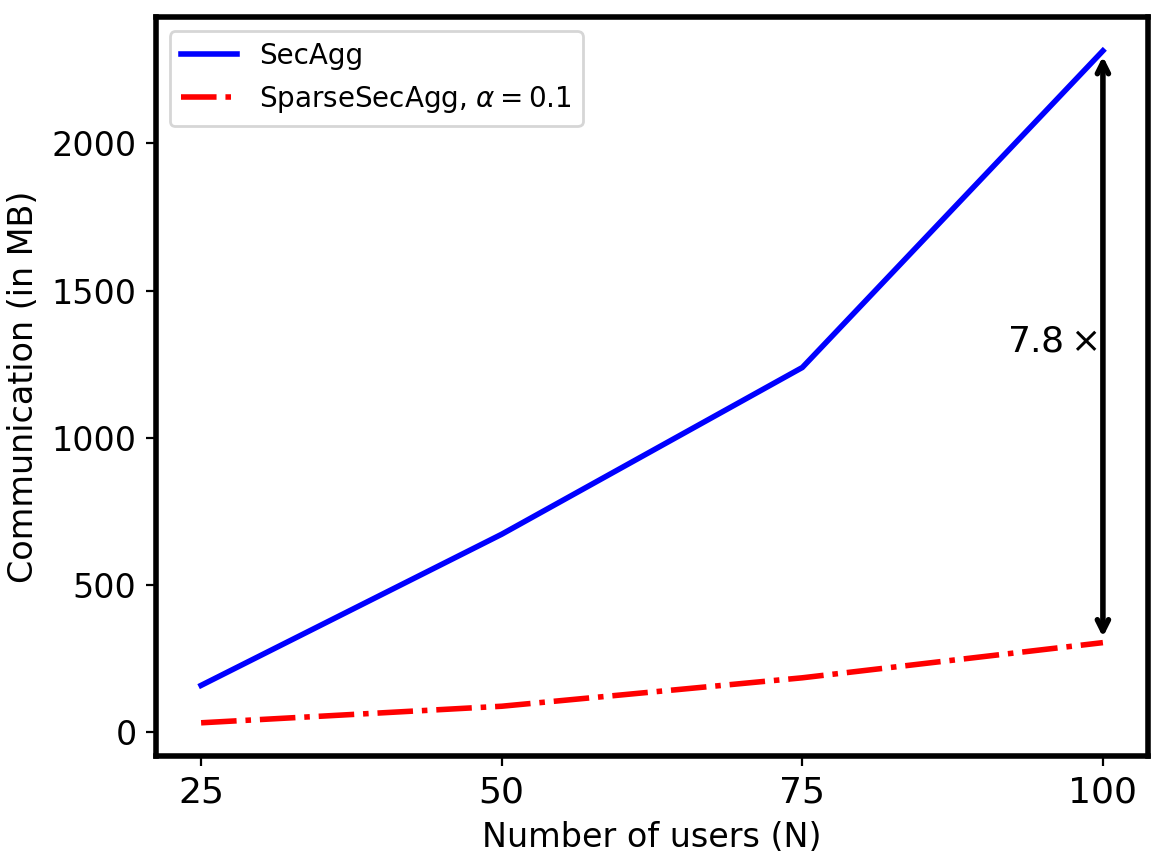}
    % \vspace{-20 pt}
    }
      \subfigure[Test accuracy vs global training rounds.]{\label{fig:iterations}
    \includegraphics[width=.305\textwidth]{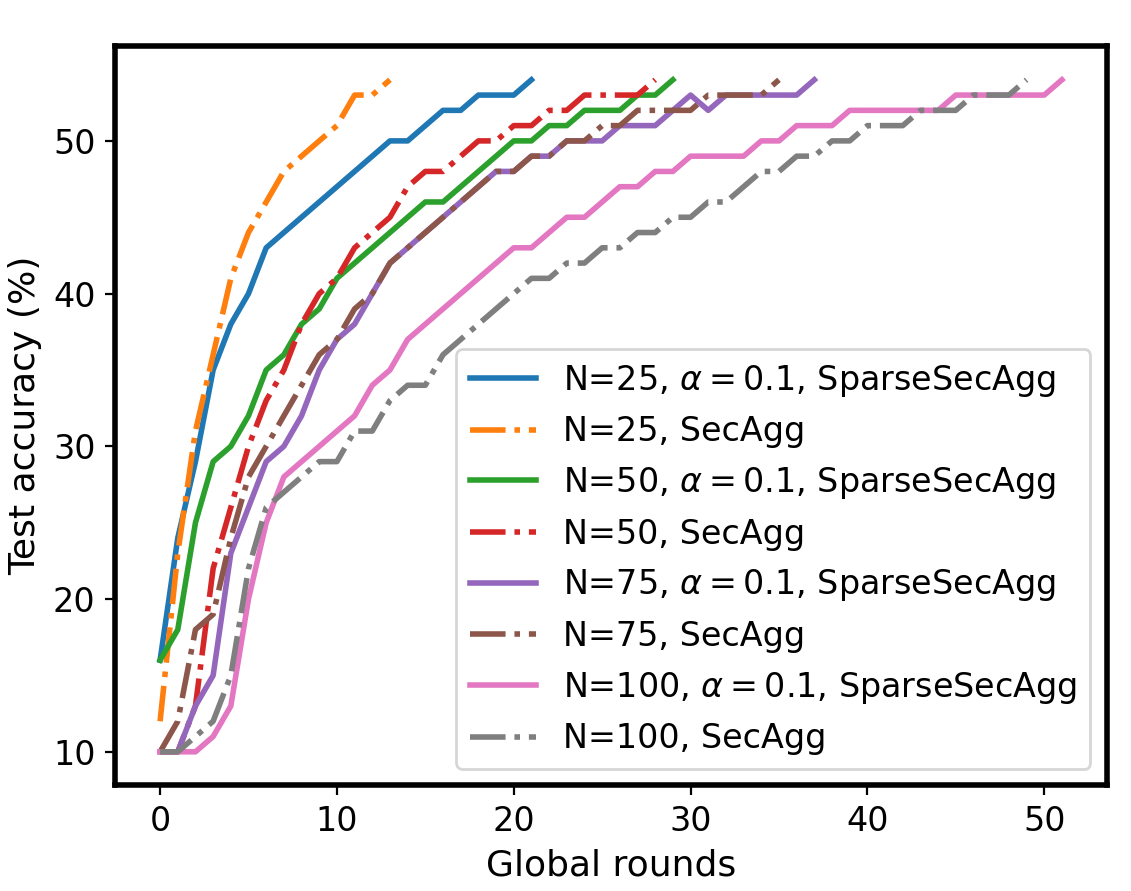}
    % \vspace{-20 pt}
    }
    \subfigure[Wall clock training time to reach the target test accuracy  ($55\%$).]{\label{fig:wallclock}
    \includegraphics[width=.315\textwidth]{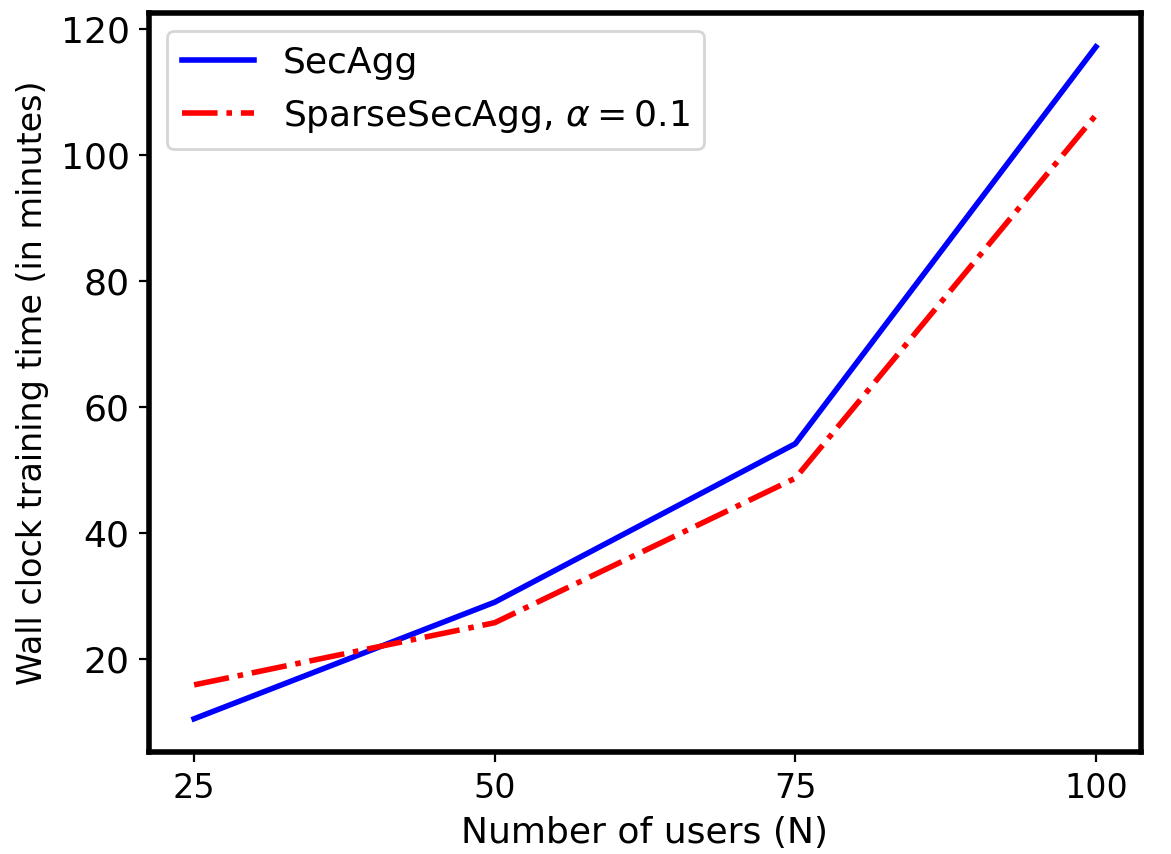}
    % \vspace{-20 pt}
    }
  
\vspace{-8 pt}
\centering
\caption{Performance comparisons on the CIFAR-10 dataset under the IID setting.}
\label{fig:perf}
\vspace{-0.2cm}\end{figure*}
\begin{figure*}[t]
\centering
 \subfigure[The average number of honest users aggregated ($T$) per model coordinate with respect to the compression ratio $\alpha$, for  $N=100$ and various dropout rates $\theta$.]{\label{fig:alphavsT}
    \includegraphics[width=.33\textwidth]{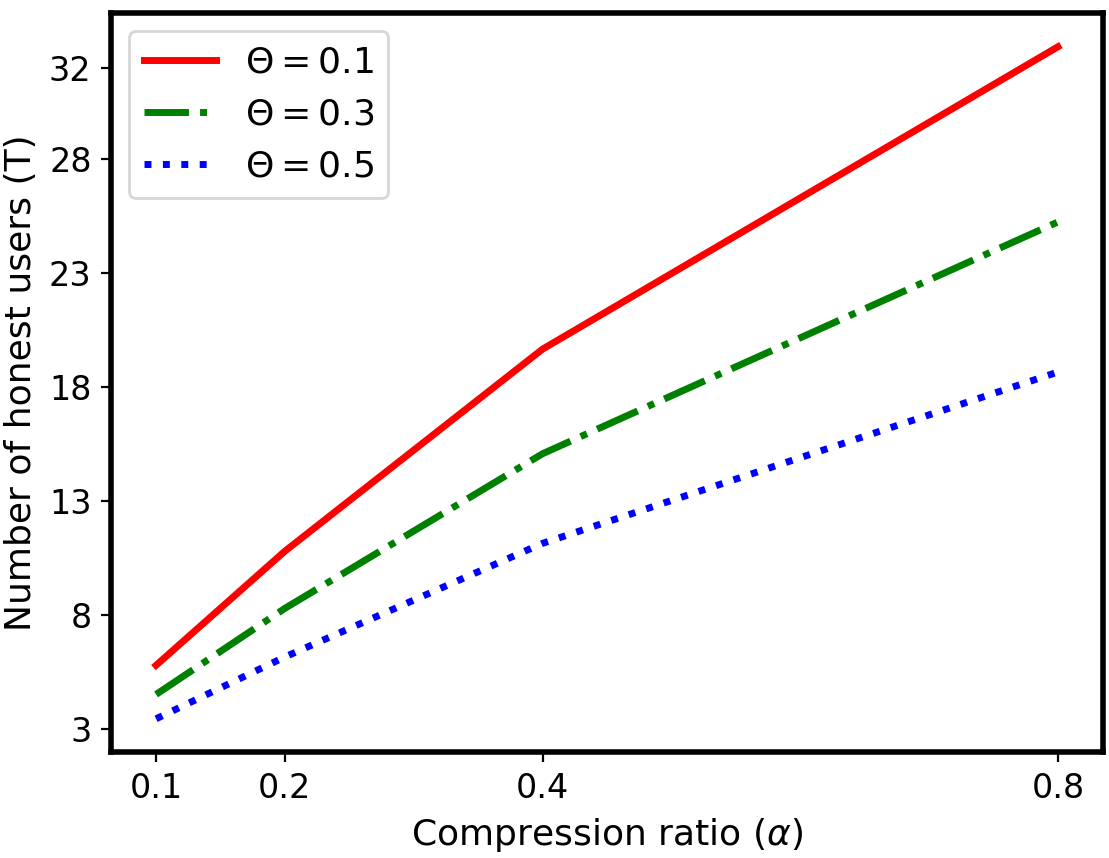}
    }
    \subfigure[The percentage of model parameters revealed.]{\label{fig:cifar_singled_out}
    \includegraphics[width=.34\textwidth]{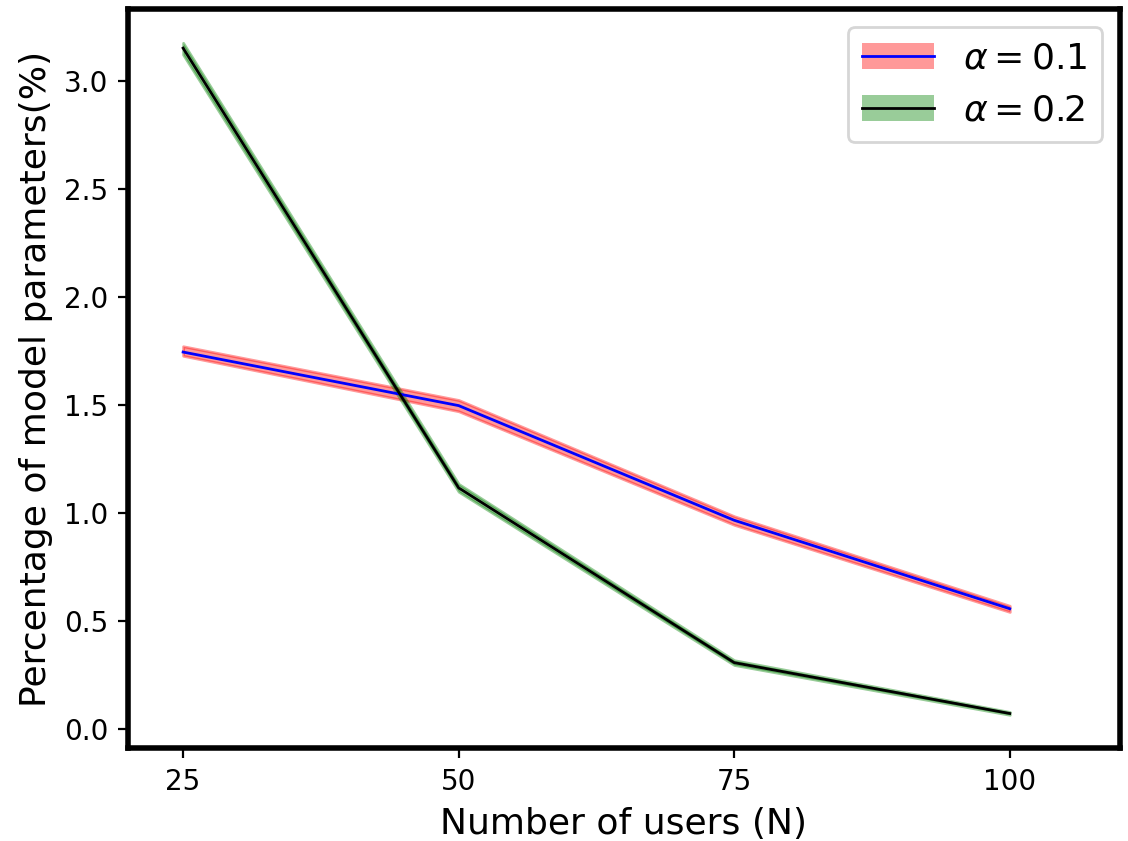}
    }
\vspace{-8 pt}
\centering
\caption{Privacy results on the CIFAR-10 dataset under the IID setting, with $A=\frac{N}{3}$ adversarial users.}
\label{fig:cif_privacy}
\vspace{-0.2cm}\end{figure*}
In our experiments, we compare the performance of \namespace with the conventional secure aggregation benchmark from \cite{bonawitz2017practical}, termed \texttt{SecAgg}, in terms of both the communication overhead and wall-clock training time. We also validate the theoretical results on the privacy guarantees offered by \namespace through experimentation.

% \subsection{Setup}

% \begin{figure}[t]
% \centering
%     \includegraphics[width=.37\textwidth]{2022-JSAC/new_figures/privacy_0.3.png}
% \vspace{-0.2cm}\caption{ \footnotesize  The percentage of total model parameters that each user is singled out at with $A=0.3N$.}
% \label{fig:privacy}
% \vspace{-0.4cm}\end{figure}

\noindent
 \textbf{Setup.} 
%  We use MNIST~\cite{lecun2010mnist} and CIFAR-10~\cite{krizhevsky2009learning} datasets for our experiments.
%  We consider image classification tasks on the MNIST~\cite{lecun2010mnist}, Fashion-MNIST \cite{xiao2017fashion}, and CIFAR-10~\cite{krizhevsky2009learning} datasets.
 We consider image classification tasks on the CIFAR-10~\cite{krizhevsky2009learning} and MNIST~\cite{lecun2010mnist} datasets, using the CNN architectures from \cite{mcmahan2017communication}.
%  We implement three different convolutional neural networks (CNNs) to evaluate the performance of our protocol. 
% We use the CNN architecture from \cite{mcmahan2017communication} in our experiments.
% and for  CIFAR-10 we utilize the mobilenetv2 CNN architecture \cite{sandler2018mobilenetv2}. 
 Unless stated otherwise, the compression ratio is set to $\alpha=0.1$, i.e., users send only one-tenth of the model parameters per round respectively, and the dropout rate is set to $\theta = 0.3$ to 
 evaluate the performance of our framework under severe network conditions.  
%  The dropout users are selected randomly. 
 We then compare the training performance of \namespace versus \texttt{SecAgg} to reach a target test accuracy.
  %unless stated otherwise. 
We then explore various compression and dropout ratios to demonstrate the impact of the compression parameter and dropout rate on the privacy guarantee. 
%  under different network conditions. 
%   We consider a dropout rate of $30\%$.
%  The dropout users are selected randomly. 
%  We run all experiments on $N+1$ Amazon EC2 \emph{m4.large} machine instances, $N$ of which operate as users and the remaining one as the server. 

 We run all experiments on Amazon EC2 \emph{m4.large} machine instances. The bandwidth of the users are set to $100$ Mbps to accurately capture the bandwidth limitations of mobile devices. 
We set the order of that finite field to $q=2^{32}-5$, which is the largest prime within $32$ bits. 
Since our main goal is to compare our protocol with the baselines under the same target accuracy and same experimental setting, our main criterion for choosing the parameters is to ensure that the model learns the given dataset, rather than to optimize the parameters to achieve state-of-the-art accuracy. 

\textbf{Performance evaluation.} 
We compare the performance of \namespace with the conventional secure aggregation in terms of wall clock training time and communication overhead to reach a target test accuracy. 
We consider federated learning settings with  $N=25,\ 50, \ 75, \ 100$ users. 
% We consider federated learning settings with various number of users  $N=20,\ 40, \ 80, \ 100$.
The datasets are distributed under both IID and non-IID settings from   \cite{mcmahan2017communication}. 
For the IID setting, the dataset is shuffled and distributed uniformly across $N$ users. For the non-IID setting, the dataset is first sorted according to the labels of each data point, and then divided into 300 shards, each shard containing samples from at most two classes. Then, each user is randomly given $300/N$ shards. This ensures the dataset sizes of the users are the same as the IID case with the same number of users. 
% , which allows us to compare the behavior of the protocol under different data sampling procedures.  
% We use SGD as the optimizer. 
% For the Fashion-MNIST experiments, we use the hyper-parameters reported in \cite{kerkouche2020compression}. 
We set the number of local training epochs to 5, batch size to 28, and the momentum parameter to 0.5. The learning rate is set to 0.01. 

\begin{figure*}[t]
\centering
   \subfigure[Total communication overhead.]{\label{fig:iid_mnist_communication}
   \includegraphics[width=.32\textwidth]{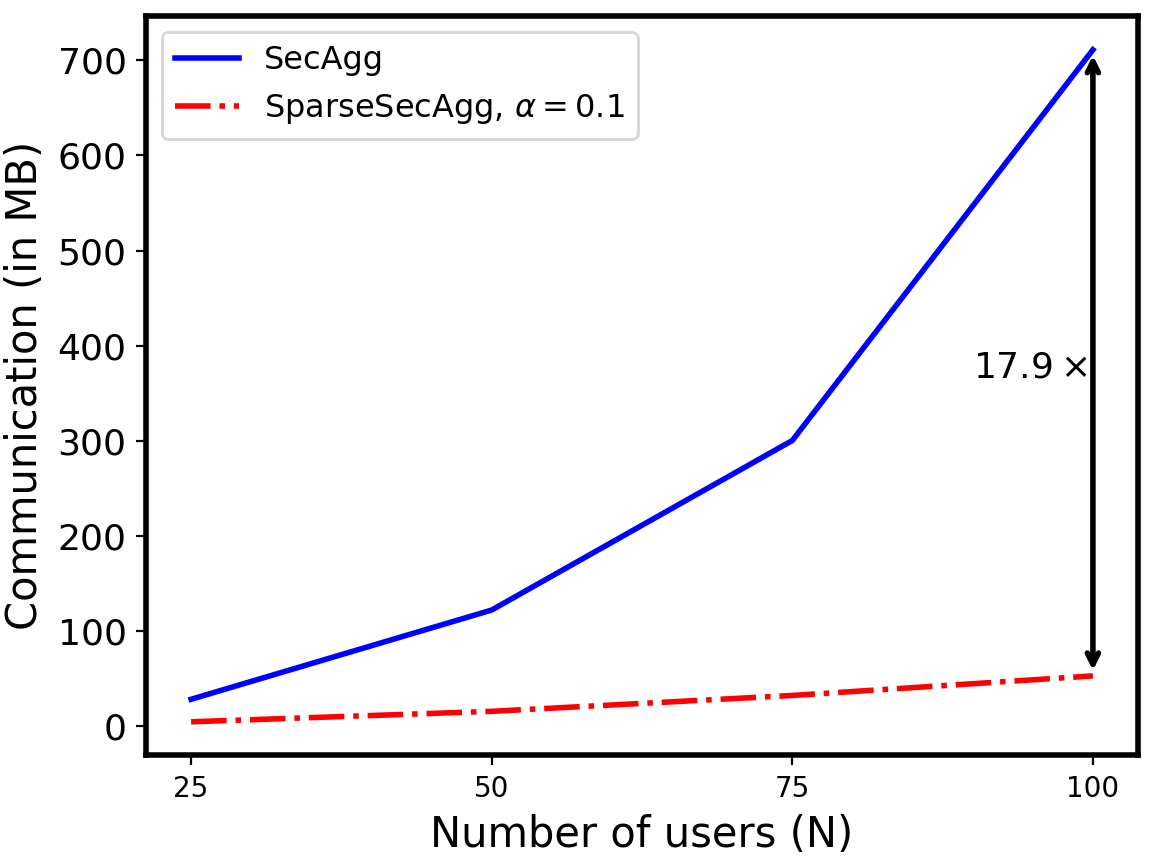}
    }
    \subfigure[Wall clock training time.]{\label{fig:wallclock_mnist}
    \includegraphics[width=.30\textwidth]{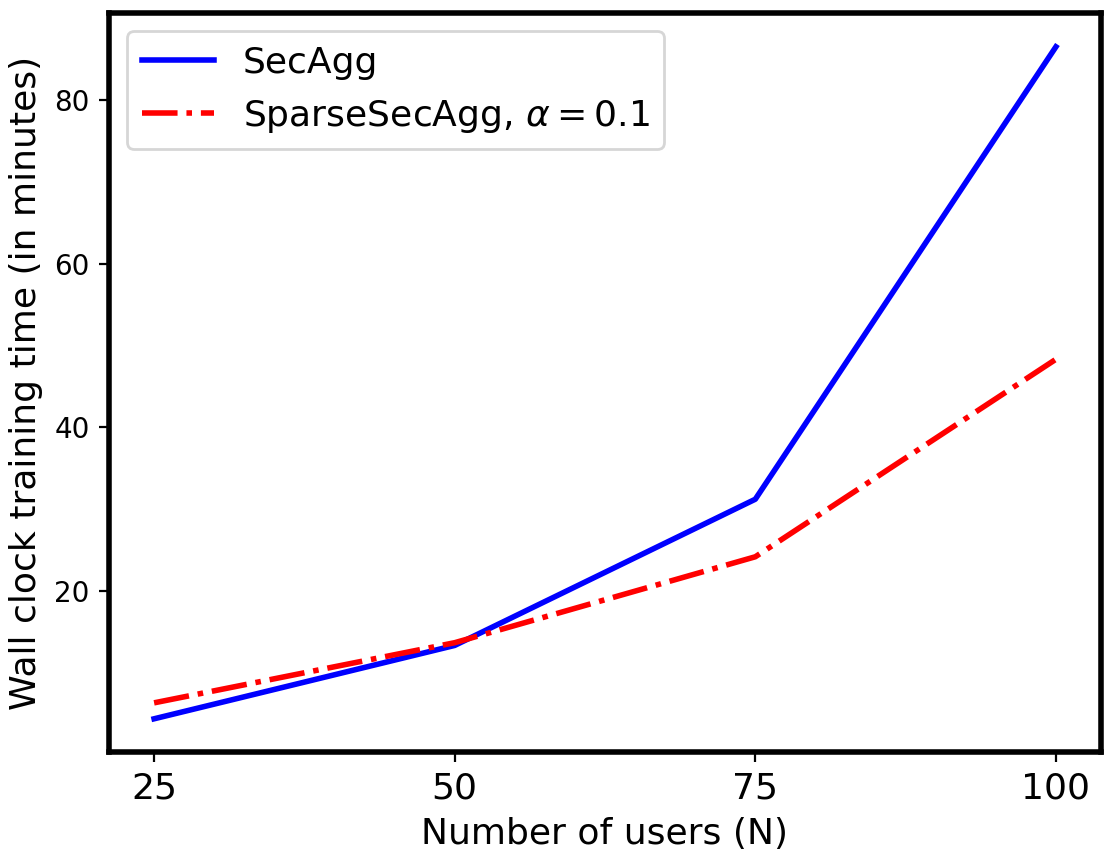}
    }
    \subfigure[Percentage of revealed model parameters of users in \namespace with $\alpha=0.1$.]{\label{fig:privacy_mnist}
    \includegraphics[width=.31\textwidth]{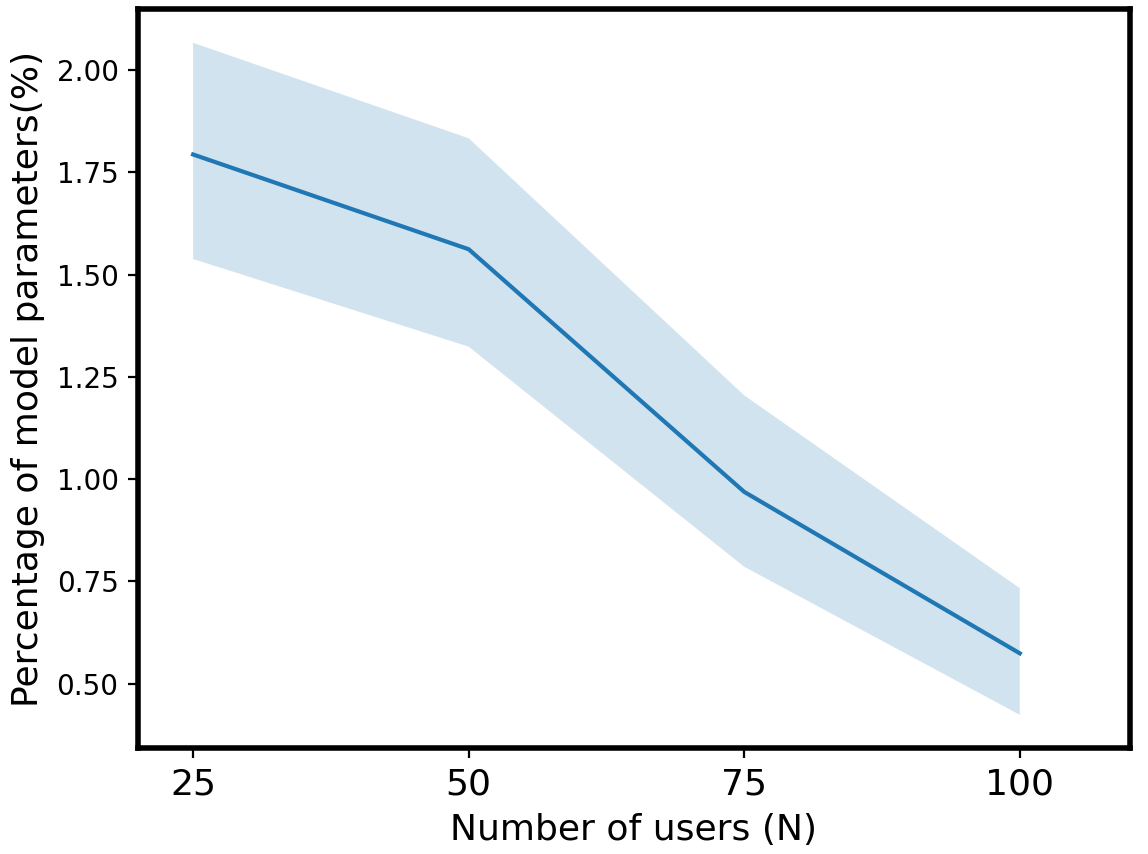}
    }
  
\vspace{-8 pt}
\centering
\caption{Results on the MNIST dataset, distributed IID across the users, with the  target test accuracy set to 97\%.}
\label{fig:perf_mnist}
\vspace{-0.2cm}\end{figure*}
\begin{figure*}[t]
\centering
    \subfigure[Total communication overhead.]{\label{fig:noniid_comm}
    \includegraphics[width=.325\textwidth]{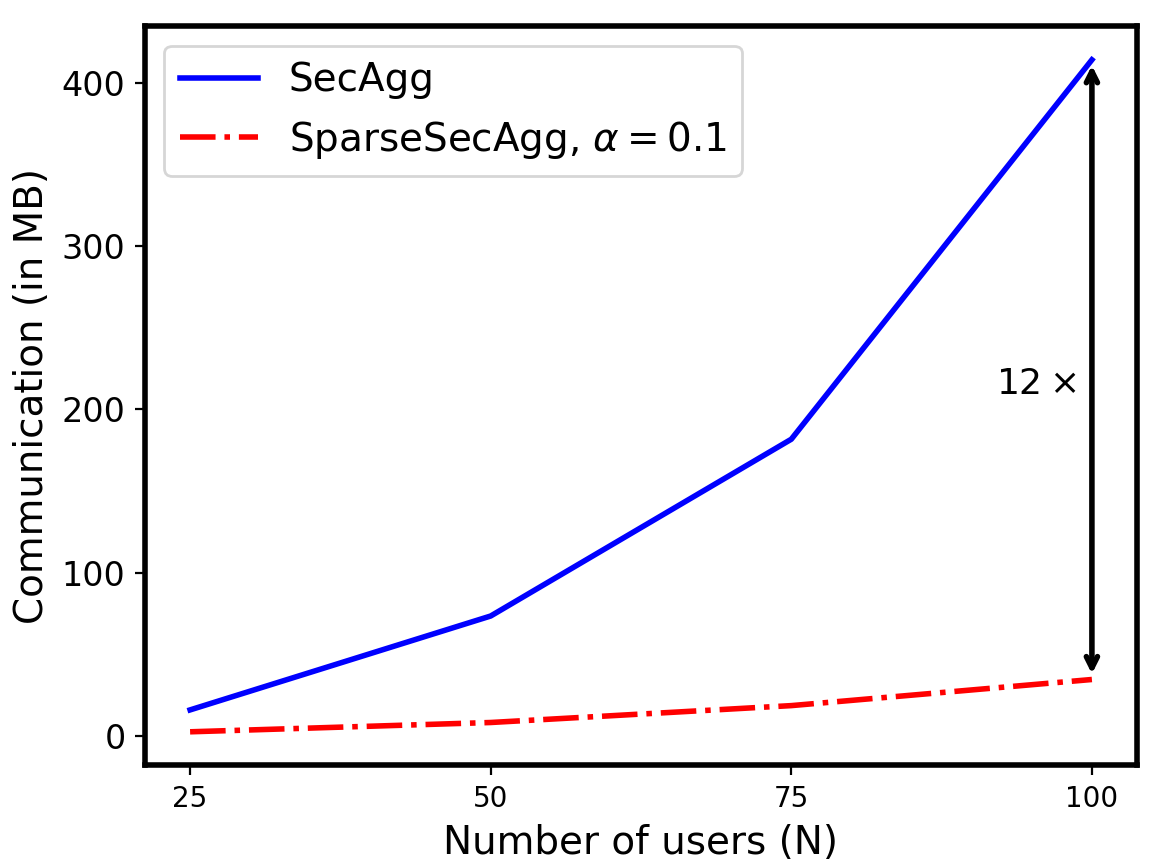}
    }
    \subfigure[ Wall clock training time.]{\label{fig:wallclock_noniid}
    \includegraphics[width=.31\textwidth]{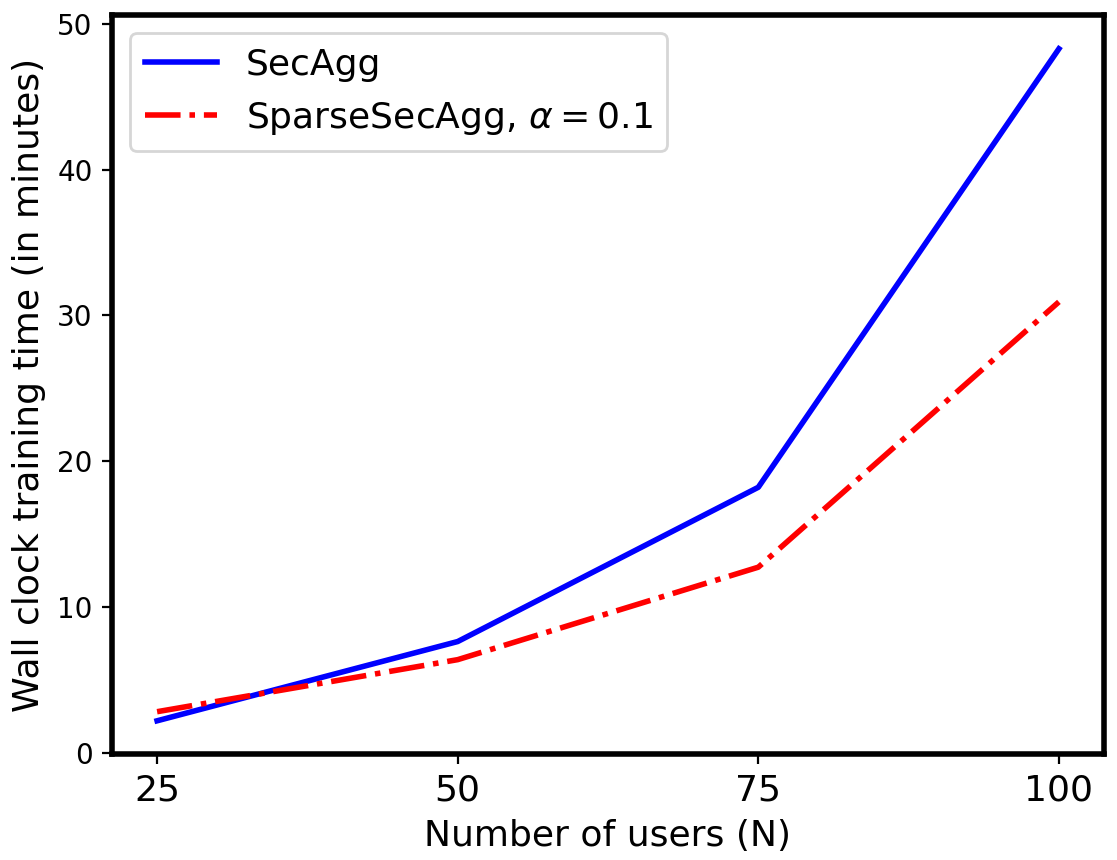}
    }

\vspace{-8 pt}
\centering
\caption{Results on MNIST dataset, distributed non-IID across the users, with target test accuracy set to 94\%.}
\label{fig:perf_mnist_noniid}
\vspace{-0.2cm}\end{figure*}

% \subsection{MNIST Experiments}

% \subsection{CIFAR-10 Experiments}
We first consider CIFAR-10 training to reach 55\% test accuracy using the CNN architecture from \cite{mcmahan2017communication} under the IID setting. 
In Table \ref{table:bandwidth}, we report the communication overhead per user per training round for \namespace ($\alpha = 0.1$) versus \texttt{SecAgg}. For \name, we report the maximum (worst-case) across all users and training rounds. We observe that the per-user communication overhead of \namespace  at any given training round  is around  $8.2\times$ smaller than \texttt{SecAgg} for $\alpha=0.1$.  This is consistent with our theoretical findings since the small increase in communication is due to the fact that the users also send the location of the model parameters to the server. 
% , hence we theoretically expect the sparsified model to be 10 times smaller than the full model. 
% , which we validate through experimentation. 
% The reason why the communication volume is one-ninth instead of one-tenth is that the users also send the model parameter locations to the server along with the parameters. 
% The small increase in the communication overhead is due to the fact that the users also send the parameter locations to the server along with the parameters. 
We use $32$ bits to represent each parameter, whereas we use one bit per parameter location (to indicate whether the corresponding parameter is selected). 
This allows us to significantly shrink the size of the location vector. 
% \footnote{Instead of using 32-bit integer values per selected parameter locations, we use one bit per parameter location to to indicate whether that model parameter was selected or not. This allows us to significantly shrink the size of the model parameter location vector and adds only a small volume to communication.}. 
% We note that the number of model parameters transmitted is proportional with the volume of communication. Therefore, the results we present in terms of megabytes is consistent with our theoretical analysis using the number of model parameters, which are described in Theorem \ref{thm:sparsity}. 

% We observe that the wall clock time for \texttt{SparseSecAgg} to reach the accuracy threshold is significantly shorter.
In Figure~\ref{fig:iid_cifar_communication}, we demonstrate the total communication overhead to reach the target accuracy, and observe that 
\namespace with $\alpha=0.1$ reduces the communication overhead by $7.8\times$ compared to \texttt{SecAgg}.
In Figure \ref{fig:iterations}, we demonstrate the convergence behavior of \namespace with $\alpha=0.1$ versus the convergence behavior of \texttt{SecAgg}. We show that even when the users are sharing one-tenth of the model update at each round, the convergence behavior of the two protocols are comparable, with \texttt{SecAgg} reaching the target accuracy only a few iterations before \namespace for the same number of users.\footnote{The increase in the number of iterations as $N$ increases is expected. The local dataset sizes of the clients shrink as $N$ grows because the dataset is divided equally to $N$ users \cite{mcmahan2017communication}.} 

In Figure \ref{fig:wallclock}, we present the wall clock training time on CIFAR-10 to reach the target test accuracy. We observe that \namespace with $\alpha=0.1$ speeds up the overall training time by $1.13\times$, hence reaches the target accuracy faster. The  main  reason  behind  the  speedup  is  the higher communication overhead of \texttt{SecAgg} per round. Hence, \namespace not only reduces the amount of data transfer per user per round, allowing users to participate to model training without being penalized due to bandwidth restrictions, but also decreases the wall clock training time. 
%Hence, \namespace is able to reach the target accuracy faster.
%This claim is further supported by the experiments we conduct on MNIST, which we detail later in this section.
% The latter is due to  \texttt{SecAgg+} reaching the target accuracy in a larger number of rounds compared to the others, even though its total communication overhead  per round is very close to \name. This is related to the fact that due to user selection, the global model is constructed using only a small part of the dataset at each round. 
% This is also demonstrated in Figure \ref{fig:iterations}, where we present the change in test accuracy until all protocols reach the target accuracy. 
  
We also experimentally demonstrate the privacy guarantees of \namespace when one-third of the users are adversarial. In Figure \ref{fig:alphavsT}, we fix $N=100$ and demonstrate the linear trade-off between privacy ($T$) and compression ratio for various dropout scenarios, and validate our theoretical findings experimentally. 

Another important implication of our theoretical analysis is that the number of local model parameters that may revealed from any given user vanishes as the number of users increase, which is important due to the probabilistic nature of our algorithm. 
In Figure \ref{fig:cifar_singled_out}, we demonstrate this phenomenon, by illustrating the percentage of the parameters which are selected by only a single honest user, hence may be revealed to the server. The solid line demonstrates the average whereas the shaded area is drawn between the minimum and maximum values, respectively. Aligned with our theoretical analysis, we observe that, for sufficiently large $N$ (i.e., $N>25$), as we increase the compression ratio, the percentage of revealed mode parameters decreases significantly, even if the users send a larger fraction of the model parameters. This demonstrates that the overlap of the model locations among honest users increase faster than the number of selected model parameters. Increasing $N$ also increases this overlap, reducing the number of revealed parameters. For instance, when $\alpha=0.2$ and $N=100$, only $0.07\%$ of the model parameters of the honest users can be singled out, making it harder for adversaries to recover any meaningful information about the dataset or the model. 

For the MNIST experiments, we consider both the IID and the non-IID settings, using the CNN architecture from \cite{mcmahan2017communication}. We set the target accuracy to 97\% for the IID setting and 94\% for the non-IID setting.
% According to \cite{sattler2019robust}, the accuracy achieved with user sampling is lower than the accuracy achieved with top-$k$ sparsification. Our results are consistent with theirs even though we utilize random$-k$ sparsificaiton. 
% In Figure \ref{fig:iter_noniid}, we demonstrate that the accuracy of \texttt{SecAgg+} fails to reach the same level of accuracy as \namespace and is less stable. The reason behind this is that the data distributions of the users selected in any two different rounds are likely to be very different from each other, possibly resulting in sudden and significant changes in the direction and/or magnitude of the gradients. \namespace reaches a higher level of accuracy within the same number of iterations. As such, \namespace is more communication efficient compared to \texttt{SecAgg+} under the same level of accuracy. 
For both of these settings, we compare the wall clock training time and communication overhead  required for \namespace with $\alpha=0.1$ and \texttt{SecAgg} to reach the target accuracy.

 In Figure \ref{fig:iid_mnist_communication}, we demonstrate that \namespace reduces the  communication overhead by $17.9\times$. In Figure \ref{fig:wallclock_mnist}, we report the wall clock time required for \texttt{SecAgg} and \namespace with $\alpha=0.1$ to reach 97\% test accuracy under the IID setting. We observe that \namespace achieves $1.8\times$ speedup over \texttt{SecAgg} for  $N=100$. In Figure \ref{fig:privacy_mnist}, we report the percentage of the model parameters selected only by a single honest user, which represents the fraction of model parameters that may be revealed to the server. 
%  , as described in Corollary \ref{cor:singled_out}. 
 The solid line demonstrates the average whereas the shaded area is drawn in between the minimum and maximum values, respectively.  We observe that these results are also consistent with our theoretical intuition and the results we report for CIFAR-10. 
% , even if communication volume of \texttt{SecAgg+} is lower than \name. 

Finally, we present the results for the MNIST dataset under the non-IID setup with a  target test accurracy 94\%. In Figure \ref{fig:noniid_comm}, we observe that \namespace reduces the communication overhead by $12\times$ compared to \texttt{SecAgg}. The improvement is further supported by the $1.2\times$ speedup in the wall clock training time to reach the target accuracy, which can be observed in Figure \ref{fig:wallclock_noniid}. We also emphasize that, for a comparable communication overhead, \namespace has only $3\%$ drop in test accuracy when the dataset is distributed in a non-IID fashion versus an IID fashion.

\section{Conclusion}\label{sec:Conclusion}
This work proposes a sparsified secure aggregation framework to tackle the communication bottleneck  of secure aggregation. 
We characterize the theoretical performance limits of the proposed framework and identify a fundamental trade-off between privacy and communication efficiency. 
Our experiments demonstrate a significant improvement in the communication overhead and wall clock training time compared to secure aggregation benchmarks.

\bibliographystyle{IEEEtran}
\bibliography{ref}
% \clearpage
\appendix

\subsection{Proof of Theorem~\ref{thm:sparsity}} \label{app:sparsity}
% \section{Tralalalala}

% \textbf{Proof of Theorem \ref{thm:sparsity}:}

% First, we provide a lemma that will be useful in our further analysis. 
% \begin{lemma}
% Let $q = 1-(1-\frac{K}{dN})^N$. Then, for any $\epsilon >0$ such that $q+\epsilon $
% \begin{equation}
% D(q+\epsilon \lVert q) \leq 2 \epsilon^2 
% \end{equation}
% where $D(\cdot\lVert \cdot)$ is the KL divergence between two Bernoulli distributions with success probability $q$ and $q+\epsilon$, respectively. 
% \end{lemma}
% We now proceed with the proof of Theorem~\ref{thm:sparsity}. 
% \begin{proof}[Proof of Theorem~\ref{thm:sparsity}]

In this section, we provide the proof of Theorem \ref{thm:sparsity}. 
As described in Section~\ref{sec:masks}, \namespace utilizes pairwise binary multiplicative masks to determine the indices of the (masked) parameters sent from each user.   
We first define a Bernoulli random variable $M_i(\ell)=1$,
\begin{equation}\label{eq:Mi}
M_i(\ell) = \left \{\begin{matrix} 1 & \text{ if }  \ell\in \mathcal{U}_i \\ 0 & \text{ otherwise } \end{matrix}\right .
\end{equation}
to represent whether the $\ell^{th}$ parameter from the local gradient  $\mathbf{y}_i^{(t)}$ of user $i$ is selected to be sent to the server. Specifically, $M_i(\ell) = 1$ if $\ell\in \mathcal{U}_i$ from \eqref{eq:locs}, and $M_i(\ell) = 0$ otherwise. 
% \begin{equation}
% M_i^{(l)} = \left \{ \begin{matrix} 1 \text {if user $i$ is selected}\\ \end{matrix}\right .
% \end{equation}

From the sparsification process described in Section~\ref{subsec:sparsified}, $M_i(\ell)=1$ as long as the pairwise binary mask $\mathbf{b}_{ij}=1$ for some $j\in[N]$, and therefore,
\begin{equation}
P[M_i(\ell) = 1] = 1-\left(1-\frac{\alpha}{N-1}\right)^{N-1} \label{eq:prob}
\end{equation}
% where $p = \frac{\alpha}{N}$, such that 
and accordingly,
\begin{equation}
\mathbb{E}[M_i(\ell)] = 1-\left(1-\frac{\alpha}{N-1}\right)^{N-1}.  
\end{equation}

% Note that, a user may drop out from the system after deciding upon the model parameters to be sent. However, the selection probability is independent of whether the user will drop out. Given user $i$ does not drop out, 
Then, the number of parameters sent from user $i$ to the server is given by $\sum_{\ell\in[d]}  M_i(\ell)$.
As the random variables $\{M_i(\ell)\}_{i\in[d]}$ are i.i.d., from Hoeffding's inequality \cite{hoeffding1994probability}, 
% Hence, from the Hoeffding-Chernoff bound \cite{hoeffding1994probability}, we have for any $\epsilon>0$, 
\begin{align}
&P\Big[\frac{1}{d} \sum_{\ell\in[d]}  M_i(\ell) 
> 1 - \Big(1-\frac{\alpha}{N-1}\Big)^{N-1} + \epsilon \Big ] \notag \\
&\qquad \qquad \qquad \leq e^{-D\big(1-\left(1-\frac{\alpha}{N-1}\right)^{N-1} + \epsilon \big \lVert 1-\left(1-\frac{\alpha}{N-1}\right)^{N-1}\big)d} \label{eq:hoeffding} \\
&\qquad \qquad \qquad \leq e^{-2\epsilon^2d}
\label{eq:hoeffding2}
\end{align}
% for any $\epsilon>0$ such that $1-(1-\frac{K}{dN})^N + \epsilon< 1$. 
for any $\epsilon>0$ such that  $\epsilon< \left(1-\frac{\alpha}{N-1}\right)^{N-1}$, where $D\big(1-\left(1-\frac{\alpha}{N-1}\right)^{N-1} + \epsilon \big \lVert 1-\left(1-\frac{\alpha}{N-1}\right)^{N-1}\big)$ denotes the KL-divergence between two Bernoulli distributions with success probability $1-\left(1-\frac{\alpha}{N-1}\right)^{N-1} + \epsilon$ and $1-\left(1-\frac{\alpha}{N-1}\right)^{N-1}$, respectively \cite{cover2006}.

% Next, note that for any $p\in(0,1)$,
% Next, note that
% Next, we use the well-known Bernoulli's inequality from mathematical analysis, which states that, 
% Next, we use a well-known inequality from mathematical analysis known as Bernoulli's inequality, which states that, 
% Next, we leverage a well-known inequality from mathematical analysis also known as Bernoulli's inequality, which states that, 
% Next, we leverage the well-known Bernoulli's inequality, which states that, 
Next, from Bernoulli's inequality, 
we have that, 
% one can observe that, 
% Next, we use the well-known Bernoulli's inequality from mathematical analysis, which states that, 
\begin{equation}
(1+x)^n\geq 1+nx. 
\end{equation} 
for any real number $x>-1$ and $n\geq 0$ \cite{ferreira2018new}. 
Since $\frac{\alpha}{(N-1)}\in(0,1]$ and $N\geq2$, we have that, 
\begin{equation}
\Big(1-\frac{\alpha}{N-1}\Big)^{N-1} \geq 1 - \alpha
\end{equation}
% \begin{equation}
% \Big(1-\frac{\alpha}{N}\Big)^{N-1} \geq \Big(1-\frac{\alpha}{N}\Big)^{N} \geq 1 - \alpha
% \end{equation}
% for any $p = \frac{K}{dN}\in(0,1]$, and therefore,  
% where $p = \frac{K}{dN}\in(0,1]$ and $N\geq1$. 
% Then,  
or equally, 
\begin{equation}\label{eq:1alphaN}
1-\Big(1-\frac{\alpha}{N-1}\Big)^{N-1} \leq \alpha. 
\end{equation}
By combining \eqref{eq:hoeffding} and \eqref{eq:1alphaN}, 
\begin{align}
&P\Big[\frac{1}{d} \sum_{\ell\in[d]}  M_i(\ell) 
> \alpha + \epsilon \Big ] \notag \\
%
% & \qquad \qquad = P\Big[\frac{1}{d} \sum_{\ell\in[d]}  M_i(\ell) 
% > \Big (\frac{\alpha}{N}\Big ) N + \epsilon \Big ] \notag \\
%
& \qquad \qquad \leq P\Big[\frac{1}{d} \sum_{\ell\in[d]}  M_i(\ell) 
> 1 - \Big(1-\frac{\alpha}{N-1}\Big)^{N-1} + \epsilon \Big ] \notag \\
%
% &\qquad \qquad \leq e^{-D\big(1-(1-\frac{\alpha}{N})^N + \epsilon \big \lVert 1-(1-\frac{\alpha}{N})^N\big)d}  \\
&\qquad \qquad \leq e^{-2\epsilon^2d} 
% & \qquad \qquad \rightarrow 0 \hspace{2cm} \text{ as }  d\rightarrow \infty 
% \label{eq:hoeffding}
\end{align}
where $e^{-2\epsilon^2d} \rightarrow 0$ as $d\rightarrow \infty$, which completes the proof.  
% Therefore, for any user $i\in[N]$, 
% \begin{equation}
% P\Big[ \sum_{\ell\in[d]}  M_i(\ell) 
% >\alpha d + \epsilon d \Big ] 
% \rightarrow 0 \text{ as } d\rightarrow \infty 
% \end{equation} 
% for any $\epsilon >0$, which completes the proof. 
% Note that, a user may drop out from the system even after deciding upon the model parameters to be sent to the server. However, the selection probability in is independent of whether the user will drop out.

Hence, the number of masked parameters sent from each user is no greater than $\alpha d$, with probability approaching to $1$ as the model size grows larger.  

% Therefore, from which, along with \eqref{eq:hoeffding}, we find that,
 
% Finally, note that 
% \begin{equation}
% e^{-2\epsilon^2d}  \rightarrow 0 \text{ as } d\rightarrow \infty \label{eq:limit}
% \end{equation}
% and therefore, from \eqref{eq:hoeffding} and \eqref{eq:limit}, 
% from which, along with \eqref{eq:hoeffding}, we find that,
% \begin{equation}
% P\Big[\frac{1}{d} \sum_{\ell\in[d]}  M_i(\ell) 
% > \frac{K}{d} + \epsilon \Big ] 
% \rightarrow 0 \text{ as } d\rightarrow \infty 
% \end{equation} 
% which completes the proof. 

% from Hoeffding-Chernoff bound, we find that: 

% note that the random variables $\{M_i^{\ell}\}_{i\in[d]}$ are i.i.d. 

% from \eqref{eq:prob}. 
% \begin{equation}
% \sum_{\ell\in[d]}  M_i(\ell). 
% \end{equation}
% In the following, we omit the user index $i$, noting that the same analysis holds for each user $i\in[N]$ due to the symmetry of the protocol. 

% From \eqref{eq:prob}, we then have,
% \begin{equation}
% E[M_i^{\ell}] = 1-(1-p)^N 
% \end{equation}
% where $p = \frac{K}{dN}$. 

% such that:
% \begin{equation}
% M_i(\ell) = \left \{ \begin{matrix} 1& \text{if user $i$ sends model parameter $\ell$ to the server} \end{matrix}\right \.
% \end{equation}
% to indicate whether user $i$ participates in the 
% \end{proof}

% \section{Proof of Privacy} 
\subsection{Proof of Theorem~\ref{thm:privacy}} \label{app:privacy}
% \input{App-Privacy-v1.tex}

% \textbf{Proof of Theorem \ref{thm:privacy}:}

% \begin{proof}
This section presents the proof of Theorem~\ref{thm:privacy}. 
% Through the proof, we assume that
% We assume that 
% From Section~\ref{sec:problem}, the privacy of a sparse secure aggregation framework is quantified by the number of honest users participating in the aggregated model. 
% We now demonstrate that the number of honest users approaches $T = \frac{k}{}$ as $N\rightarrow \infty$. 
First, we define a Bernoulli random variable $M_i(\ell)\in\{0,1\}$ as in \eqref{eq:Mi}, to denote whether or not parameter $\ell$ is selected by user $i$ to be sent to the server, where the probability $P[M_i(\ell) = 1]$ is as given in \eqref{eq:prob}. 
% , with $p = \frac{K}{dN}$. 
% From  Section~\ref{app:sparsity}, we have that:
% \begin{equation}
% P[M_i(\ell) = 1] = 1-(1-p)^N \label{eq:prob2}
% \end{equation}
% where $p = \frac{K}{dN}$.  
From \eqref{tildep}, we observe that, 
\begin{equation} \label{eq:q}
p = 1-\left(1-\frac{\alpha}{N-1}\right)^{N-1} = P[M_i(\ell) = 1]. 
\end{equation}

We then define a Bernoulli random variable $D_i$,
\begin{equation}\label{dropout RV}
 D_i = \left \{\begin{matrix} 0 & \text{ if user $i$ drops out}  \\ 1 & \text{ otherwise } \end{matrix}\right .   
\end{equation}
to represent whether user $i$ drops out during the aggregation step. Since a user may drop out with probability $\theta$, the probability that user $i$ sends the parameter corresponding to location $\ell$ to the server is given as follows:
\begin{align}
p^{\prime} &:= P[D_{i}=1, M_{i}(\ell)=1] \notag \\
&= P[D_{i}=1]P[M_{i}(\ell)=1]\notag \\
&= (1-\theta)p \label{p_prime}
\end{align}
% Next, we define, 
% \begin{equation} \label{eq:q}
% p:= 1-(1-p)^N = P[M_i(\ell) = 1]
% \end{equation}
Then, the number of users that participate in the aggregated gradient for a given location $\ell$ is,
\begin{equation}\label{eq:M}
M_D := \sum_{i\in[N]}  D_iM_i(\ell)
\end{equation}
% $\sum_{i\in[N]}  M_i^{(l)}$, 
and we define the empirical mean,
\begin{equation}\label{eq:sumM}
% \bar{M}_D^{(l)} = \sum_{i\in[N]}  M_i^{(l)},
\bar{M}_D := \frac{1}{N}\sum_{i\in[N]}D_i  M_i(\ell) 
\end{equation} 
% where $\bar{M}_D$ is a Binomial random variable with a success probability $q\in(0,1)$. 
% We omit the index $\ell$ in \eqref{eq:sumM} for simplicity, as the analysis is the same for all $l\in[d]$ due to  symmetry. 
% \begin{equation}
% \mathbb{E}[M_i^{\ell}] = 1-(1-p)^N.  
% \end{equation}
% We assume that $\frac{N}{2}$ out of $N$ users are adversarial in our further analysis, and note that the same  analysis also holds for a smaller number of adversaries. 
First, note that Shamir's $\frac{N}{2}$-out-of-$N$ secret sharing, which is employed for secret sharing the random seeds as described in Section~\ref{sec:masks} guarantees that, any set of  $A\leq\frac{N}{2}$ adversaries cannot recover the pairwise and private seeds created by honest users, even if they collude with each other and/or the server. 
As such, any set of up to $A\leq\frac{N}{2}$ adversaries cannot reveal the individual local gradients.
% As such, as long as there are $A<\frac{N}{2}$ adversaries,  cannot reveal the individual local models.
% However, adversaries can still try to remove their local models from the aggregated model, in an attempt to reveal information 
% However, adversaries can still remove their local models from the aggregated model, in an attempt to reveal information about the honest users. 
However, adversaries may still remove their local gradients from the aggregated gradient, in an attempt to reduce the number of local gradients in the aggregated gradient and thus render secure aggregation ineffective. In the sequel, we show that even after adversaries remove all their local gradients from the aggregated gradient, there will be at least $T$ local gradients belonging to the honest users. Hence, the adversaries can not observe the aggregate of fewer than $T$ local gradients. 

In our analysis, we consider the scenario where $\frac{N}{2}$ out of $N$ users are adversarial, noting that the same  analysis carries over also to a smaller number of adversaries. 
% We consider the worst case scenario where $\frac{N}{2}$ out of $N$ users are adversarial in our analysis, and note that the same  analysis also holds for a smaller number of adversaries. 
We define a binary random variable $Y_i\in\{0,1\}$ to represent whether user $i$ is honest or adversarial, 
% In particular, 
\begin{equation}
Y_i =  \left \{ \begin{matrix}
1 & \text{ if user $i$ is honest (not adversarial)} \\
0 & \text{ otherwise} \\
\end{matrix}\right .
\end{equation}
where $A = \frac{N}{2}$ users are adversarial. The adversarial users are distributed uniformly at random among the  $N$ users, hence $P[Y_i = 1] = \frac{1}{2}$ for all $i\in[N]$. 
% Note that the random variables $Y_i$ for  $i\in[N]$ are identically distributed but not independent.  
% We assume that $\frac{N}{2}$ out of $N$ users are adversarial, noting that our analysis holds with a smaller number of adversaries. 
% We assume that $\frac{N}{2}$ out of $N$ users are adversarial, noting that the analysis is the same for a smaller number of adversaries. 
% We assume that $\frac{N}{2}$ out of $N$ users are adversarial in our analysis, and note that the our analysis also holds for a smaller number of adversaries. 
% Our analysis in the sequel is given for the worst case scenario of $\frac{N}{2}$ adversaries. 

% Our goal is to quantify the number of honest users in the aggregated model. 
% The number of honest users that participate at location $\ell$ in the aggregated model is given as:
Next, we define a binary random variable, 
\begin{equation}
X_i(\ell) := D_iM_i(\ell) Y_i, 
\end{equation}
such that $X_i(\ell) = 1$ if user $i$ participates in the aggregated gradient at location $\ell$ and is an honest user. Then, 
\begin{align}
\mathbb{E}[X_i(\ell)] &= P[X_i(\ell) = 1] \\
&=P[D_i=1] P[M_i(\ell) = 1]P[Y_i = 1] \\&= \frac{p^{\prime}}{2}
\end{align} 
% In particular, $X_i^{(l)} = 1$ if user $i$ participates in the aggregated model at location $\ell$ and is an honest (non-adversarial) user. 
% $X_i^{(l)}\in\{0,1\}$ such that:
% \begin{equation}
% A_i =  \left \{ \begin{matrix}
% 1 & \text{ if user $i$ is adversarial} \\
% 0 & \text{ otherwise} \\
% \end{matrix}\right .
% \end{equation}
Then, the number of honest users that participate in the aggregated gradient at any location $\ell\in[d]$ is,
\begin{equation}
X :=\sum_{i\in[N]} X_i(\ell)=  \sum_{i\in[N]} D_iM_i(\ell)  Y_i.   
\end{equation}
We also define the empirical mean,
\begin{equation}
\bar{X} := \frac{1}{N}\sum_{i\in[N]} X_i(\ell).   
\end{equation}

For any $\epsilon_1, \epsilon_2 >0$ and $l\in[d]$, one can write
\begin{align}
% &P[\bar{X} < \frac{NK}{2d}-\delta-\epsilon]
&P[\bar{X} \leq \frac{p^{\prime}}{2}-\epsilon_1] \notag \\
&  = P[\bar{X} \leq \frac{p^{\prime}}{2}-\epsilon_1 | \bar{M}_D < p^{\prime} - \epsilon_2]  P[\bar{M}_D < p^{\prime} - \epsilon_2] \notag \\
& \quad + P[\bar{X} \leq \frac{p^{\prime}}{2}-\epsilon_1 | \bar{M}_D > p^{\prime} + \epsilon_2]  P[\bar{M}_D > p^{\prime} + \epsilon_2] \notag \\
& \quad +  P[\bar{X} \leq \frac{p^{\prime}}{2}-\epsilon_1 | p^{\prime} - \epsilon_2 \leq \bar{M}_D  \leq  p^{\prime} + \epsilon_2]   P[p^{\prime} - \epsilon_2 \leq \bar{M}_D  \leq  p^{\prime} + \epsilon_2] \\
& \leq P[\bar{M}_D \leq p^{\prime} - \epsilon_2] + 
P[\bar{M}_D \geq p^{\prime} + \epsilon_2] 
+ P[\bar{X} \leq \frac{p^{\prime}}{2}-\epsilon_1, p^{\prime} - \epsilon_2 \leq \bar{M}_D  \leq  p^{\prime} + \epsilon_2] \label{eq:bound}
\end{align}
Next, we upper bound each term on the right hand side of \eqref{eq:bound}.  
% Next, we upper bound each term from \eqref{eq:bound}.  
% First, it follows from Chernoff bound on the tail bound  of a  Binomial random variable that,
% First, the tail probability of the  Binomial random variable $\bar{X}$ with parameter $q\in(0,1)$ can be bounded as,
For the first and second terms in \eqref{eq:bound}, the tail probability of  $\bar{M}_D$ can be bounded using  Hoeffding's inequality \cite{hoeffding1994probability} as,
\begin{equation}\label{eq:bound1}
 P[\bar{M}_D \leq p^{\prime} - \epsilon_2] \leq e^{-{\epsilon_2}^2 N},  
\end{equation}
and
% Similarly, the (upper) tail probability can be bounded from the Chernoff bound as,
% Similarly, for the second term, the (upper) tail probability can be bounded as,
\begin{equation}\label{eq:bound2}
P[\bar{M}_D \geq p^{\prime} + \epsilon_2] \leq e^{-{\epsilon_2}^2 N}. 
\end{equation}
For the third term in \eqref{eq:bound}, we observe that,  
% Then, 
\begin{align}
&P[\bar{X} \leq \frac{p^{\prime}}{2}-\epsilon_1, p^{\prime} - \epsilon_2 \leq \bar{M}_D  \leq  p^{\prime} + \epsilon_2] \notag \\
& \leq \sum_{m = \lceil N(p^{\prime} - \epsilon_2) \rceil}^{\lfloor N(p^{\prime} + \epsilon_2) \rfloor} P[\bar{X} \leq \frac{p^{\prime}}{2}-\epsilon_1 , M_D = m]  \label{eq:union} \\ 
& = \sum_{m = \lceil N(p^{\prime} - \epsilon_2) \rceil}^{\lfloor N(p^{\prime} + \epsilon_2) \rfloor} P[\bar{X} \leq \frac{p^{\prime}}{2}-\epsilon_1| M_D = m]  P [M_D=m]\\ 
& \leq \sum_{m = \lceil N(p^{\prime} - \epsilon_2) \rceil}^{\lfloor N(p^{\prime} + \epsilon_2) \rfloor} P[\bar{X} \leq \frac{p^{\prime}}{2}-\epsilon_1| M_D = m] \\
& = \sum_{m = \lceil N(p^{\prime} - \epsilon_2) \rceil}^{\lfloor N(p^{\prime} + \epsilon_2) \rfloor} P[X \leq N(\frac{p^{\prime}}{2}-\epsilon_1)| M_D = m] \label{eq:sum1}
\end{align}
% \begin{align}
% &P[\bar{X} \leq \frac{q}{2}-\epsilon , q - \epsilon_2 \leq \bar{M}_D  \leq  q + \epsilon_2] \notag \\
% &\qquad \leq \sum_{m = \lceil q - \epsilon_2 \rceil}^{\lceil q + \epsilon_2 \rceil} P[\bar{X} \leq \frac{q}{2}-\epsilon , \bar{M}_D = m]  
% \end{align}
% where \eqref{eq:union} follows from the union bound of probability. Next, for a given $M=m$, we define $m$ random variables $I_{1}, \ldots, I_{m}$ to represent the random integer serving as the index of the $m$ terms for which $M_{i}^{(\ell)} = 1$. 
% where \eqref{eq:union} follows from the union bound of probability. Next, for a given $M=m$, we define $m$ random variables $I_{1}, \ldots, I_{m}$ to represent the index of the $m$ terms for which $M_{i}^{(\ell)} = 1$. 
% where \eqref{eq:union} follows from the union bound of probability. 
% For a given $M=m$, we define $m$ random variables $I_{1}, \ldots, I_{m}$ represent the index of the $m$ terms for which $M_{i}^{(\ell)} = 1$ as follows. 
We then define $m$ random variables $I_{1}, \ldots, I_{m}$ such that $I_j< I_{j'}$ for all $j>j'$, to represent the index of the $m$ terms for which $D_iM_{i}{(\ell)} = 1$. 
% \begin{align}
% (I_{1}, \ldots, I_{m}) = (i_1, \ldots, i_m) \text{ if and only if } M_{i_j}^{(l)} = 1 \forall i_j\in[m] \text{ and } i_j \neq i_{j'}.  
% \end{align}
% Let $i_0 = 0$. 
% Then, for any $j\in[M]$ and $i_j\in[N]$, define, 
% For $j\in[m]$, define, 
% \begin{equation}
% I_j = i_j \text{ if and only if } M_{i_j}^{(l)} = 1 \text{ and } i_j \neq i_{j'} 
% \end{equation}
% Define $(I_{1}, \ldots, I_{m}) = (i_1, \ldots, i_m)$ if and only if  $M_{i_j}{(\ell)} = 1$ and  $i_j \neq i_{j'}$ for all $j, j'\in[m]$.  
% As such, the random variables $I_{1}, \ldots, I_{m}$ represent the index of the $m$ terms for which $M_{i}^{(\ell)} = 1$. 
Then, using the chain rule, we can rewrite \eqref{eq:sum1} as, 
\begin{align}
& \sum_{m = \lceil N(p^{\prime} - \epsilon_2) \rceil}^{\lfloor N(p^{\prime} + \epsilon_2) \rfloor} P[X \leq N(\frac{p^{\prime}}{2}-\epsilon_1)| M_D = m] \notag \\
& \qquad = \sum_{m = \lceil N(p^{\prime} - \epsilon_2) \rceil}^{\lfloor N(p^{\prime} + \epsilon_2) \rfloor} 
\sum_{\substack{i_1, \ldots, i_m \in [N]:\\ i_j< i_{j'} \forall j'>j }}  P[\sum_{j=1}^m Y_{i_j} \!\leq\! N(\frac{p^{\prime}}{2}\!-\!\epsilon_1)| M_D \!=\! m, I_1 \!\!=\! i_1, \ldots, I_m \!\!=\! i_m] \notag \\
& \hspace{4cm} \times P[I_1 = i_1, \ldots, I_m = i_m| M_D = m] \\
& \qquad  = \sum_{m = \lceil N(p^{\prime} - \epsilon_2) \rceil}^{\lfloor N(p^{\prime} + \epsilon_2) \rfloor} 
\sum_{\substack{i_1, \ldots, i_m \in [N]:\\ i_j< i_{j'} \forall j'>j }}   \! P[\sum_{j=1}^m Y_{i_j} \leq N(\frac{p^{\prime}}{2}\!-\!\epsilon_1)] P[I_1 = i_1, \ldots, I_m = i_m| M_D = m] \label{eq:A}
% & = \sum_{m = \lceil N(q - \epsilon_2) \rceil}^{\lfloor N(q + \epsilon_2) \rfloor} 
% \sum_{\substack{\{i_j\}_{j=1}^m \in [N]\\ i_j\neq i_{j'}}} P[\sum_{j=1}^m (1-A_{i_j}) \leq N(\frac{q}{2}-\epsilon_1)| M_D = m, I_1 = i_1, \ldots, I_m = i_m] P[I_1 = i_1, \ldots, I_m = i_m| M_D = m]
% & = \sum_{m = \lceil N(q - \epsilon_2) \rceil}^{\lfloor N(q + \epsilon_2) \rfloor} 
% \sum_{\substack{i_1, \ldots, i_m \in [N]\\ i_j\neq i_{j'}}} P[\sum_{j=1}^m (1-A_{i_j}) \leq N(\frac{q}{2}-\epsilon_1)| M_D = m, I_1 = i_1, \ldots, I_m = i_m] P[I_1 = i_1, \ldots, I_m = i_m| M_D = m]
\end{align}
where \eqref{eq:A} follows from the fact that $\{Y_i\}_{i\in[N]}$ are independent from $\{D_i M_i(\ell)\}_{i\in[N]}$. 

We now bound the first term in \eqref{eq:A}.  
% For bounding the first term in the right hand side of \eqref{eq:A}, 
For this, we first note that the terms $Y_i$ for $i\in[N]$ are not independent. As such, bounds originally defined for the sum of independent random variables, such as Hoeffding's inequality, cannot immediately be applied for bounding the first term in \eqref{eq:A}. 
To address this, we utilize a relation between sampling with and without replacement on bounding the probability of sum of dependent random variables \cite[Section~5]{hoeffding1994probability}. 
% To do so, we first define $m$ random variables $Y_1, \ldots, Y_m$ such that:
% To do so, we first define $m$ random variables $Y_1, \ldots, Y_m$ such that:
% \begin{equation}\label{eq:Yjdef}
% Y_j := 1 - A_{i_j} \text{ for all } j\in[m]. 
% \end{equation}
% Note that $N/2$ users are adversarial out of $N$ users, and the adversarial users are distributed uniformly at random across the $N$ users. 
% As such, the random variables $Y_1, \ldots, Y_m$ represent sampling (without replacement) $m$ samples from a population of size $N$ that contains $\frac{N}{2}$ adversarial users. Therefore, 
% \begin{equation}
% P[Y_j = 1] = \frac{N/2}{N} = \frac{1}{2} 
% \end{equation}
% and,
% \begin{equation}
% \mathbb{E}[Y_j] = P[Y_j = 1] = \frac{1}{2}. 
% \end{equation}
% However, note that $Y_1, \ldots, Y_m$ are not independent. 

Next, we define $m$ IID binary random variables $Z_1, \ldots, Z_m$ with the same marginal distribution as $Y_{i_1}, \ldots, Y_{i_m}$. In particular, 
\begin{equation}
P[Z_j = 1] := P[Y_j = 1] = \frac{1}{2}   
\end{equation}
and
\begin{equation}
\mathbb{E}[Z_j] = \frac{1}{2}, 
\end{equation}
for all $j\in[m]$. 
Note that while $Y_{i_1}, \ldots, Y_{i_m}$ represented sampling without replacement, $Z_1, \ldots, Z_m$ represents sampling with replacement, from a population of size $N$ that contains $\frac{N}{2}$ adversarial users.
% , where half of the terms are equal to $1$ and the remaining terms are equal to $0$. 
% Then, we can use Hoeffding's inequality to bound the tail probability of the sum of $Z_1, \ldots, Z_m$. Specifically, for any $\epsilon_3 >0$,
% \begin{equation}\label{eq:boundZ}
%  P[\frac{1}{m}\sum_{j\in[m]} Z_j \leq \frac{1}{2} - \epsilon_3] \leq e^{-{\epsilon_2}^2 m}.  
% \end{equation}
It has been shown in \cite[Section~5]{hoeffding1994probability} that bounds on the sum of the latter can also be leveraged to bound the former. In particular, 
% that bounds on  $\sum_{j\in[m]} Z_j$ also provide a bound on  $\sum_{j\in[m]} Y_{i_j}$. In particular,  
% \eqref{eq:boundZ} also provides an upper bound on the tail probability of $\sum_{i\in[m]} Y_i$, 
\begin{align}
 P[\frac{1}{m}\sum_{j\in[m]} Y_{i_j} \leq \frac{1}{2} - \epsilon_3] & \leq  P[\frac{1}{m}\sum_{j\in[m]} Z_j \leq \frac{1}{2} - \epsilon_3]  \\
 & \leq e^{-{\epsilon_3}^2 m}.  \label{eq:boundY}
\end{align}
where \eqref{eq:boundY} follows from Hoeffding's inequality. 
% \begin{equation}
%   P[\frac{1}{m}\sum_{i\in[m]} Y_i \leq \frac{1}{2} - \epsilon_3]   \leq  P[\frac{1}{m}\sum_{i\in[m]} Z_i \leq \frac{1}{2} - \epsilon_3]  \label{eq:boundY}
% \end{equation}
% and note that one can always find such $\delta, \epsilon_2$, and $\epsilon_3$. 
% For $\lceil N(q - \epsilon_2) \rceil \leq m \leq \lfloor N(q + \epsilon_2) \rfloor$,  
Then, for $m \geq \lceil N(p' - \epsilon_2) \rceil$, 
\begin{align}
& P[\sum_{j\in[m]} Y_{i_j} \leq N(p^{\prime} - \epsilon_2)   (\frac{1}{2} - \epsilon_3)]  \notag  \\
& \qquad \leq P[\sum_{j\in[m]} Y_{i_j} \leq \lceil N(p^{\prime} - \epsilon_2) \rceil  (\frac{1}{2} - \epsilon_3)]    \\
  & \qquad \leq  P[\sum_{j\in[m]} Y_{i_j} \leq m(\frac{1}{2} - \epsilon_3)] \\ 
    & \qquad \leq e^{-{\epsilon_3}^2 m} \\
    & \qquad \leq e^{-{\epsilon_3}^2 \lceil N(p^{\prime} - \epsilon_2) \rceil} \label{eq:Yfirst}
\end{align}
% Next, note that, 
% % for $p = \frac{\alpha}{N}\in[0,1]$, 
% \begin{equation}
% (1-\frac{\alpha}{N-1})^{N-1} \rightarrow e^{-\alpha} \text{ as } N\rightarrow \infty
% \end{equation}
% and therefore,
% \begin{equation}
% p\rightarrow 1 - e^{-\alpha} \text{ as } N\rightarrow \infty.
% \end{equation}
% In other words, for any $\delta > 0$, there exists some $n_0$ such that, for any $N\geq n_0$, 
% \begin{equation}\label{limit}
% \left |p-(1 - e^{-\alpha} ) \right | < \delta  
% \end{equation} 
% Now, for a given $\epsilon_1>0$, select $\delta, \epsilon_2, \epsilon_3 > 0$ sufficiently small  so that,
% \begin{equation}
% \frac{\epsilon_2}{2} + \epsilon_3 (1- e^{-\alpha} + \delta - \epsilon_2) < \epsilon_1.  
% \end{equation}
% Then, from \eqref{limit}, there exists some $n_0$ such that, $p < 1 - e^{-\alpha} + \delta$ for all $N\geq n_0$, and therefore
% \begin{equation}
% \frac{\epsilon_2}{2} + \epsilon_3 (q - \epsilon_2) < \epsilon_1,
% \end{equation}
Next, select $\epsilon_1$, $\epsilon_2$, and $\epsilon_3$ such that,
\begin{equation}
\epsilon_1  > \frac{\epsilon_2}{2} + \epsilon_3 (p^{\prime} - \epsilon_2).  
\end{equation}
Then, \eqref{eq:A} can be bounded as, 
% from which we have that, 
% \begin{align}
% & P[\sum_{j\in[m]} Y_j \leq N(\frac{p^{\prime}}{2} - \epsilon_1)]  \notag \\
% & \leq P\Bigg[\sum_{j\in[m]} Y_j \leq N\bigg(\frac{p^{\prime}}{2} - \Big(\frac{\epsilon_2}{2} + \epsilon_3 (p - \epsilon_2)\Big)\bigg)\Bigg]  \\ 
% & = P[\sum_{j\in[m]} Y_j \leq N(p - \epsilon_2)   (\frac{1}{2} - \epsilon_3)] \label{eq:Yfinal}
% % & \leq P[\sum_{j\in[m]} Y_j \leq N(\frac{e^{-\frac{K}{d}} + \delta} {2} - \epsilon_1)]  \notag \\
% % & \leq P[\sum_{j\in[m]} Y_j \leq N(q - \epsilon_2)   (\frac{1}{2} - \epsilon_3)] 
% % &  = P[\sum_{j\in[m]} Y_j \leq N(q - \epsilon_2)   (\frac{1}{2} - \epsilon_3)] 
% \end{align}
% By combining \eqref{eq:Yfinal}, with \eqref{eq:Yj} and \eqref{eq:Yfirst} and using \eqref{eq:Yjdef}, \eqref{eq:A} can be bounded as,
% By combining  \eqref{eq:Yjdef}, \eqref{eq:Yj}, and \eqref{eq:Yfirst} along with \eqref{eq:Yfinal}, we can bound \eqref{eq:A} as,
\begin{align}
& \sum_{m = \lceil N(p^{\prime} - \epsilon_2) \rceil}^{\lfloor N(p^{\prime} + \epsilon_2) \rfloor} 
\sum_{\substack{i_1, \ldots, i_m \in [N]:\\ i_j< i_{j'} \forall j'>j}}   \! P[\sum_{j=1}^m Y_{i_j} \leq N(\frac{p^{\prime}}{2}\!-\!\epsilon_1)]  P[I_1 = i_1, \ldots, I_m = i_m| M_D = m] \\
&\leq  \sum_{m = \lceil N(p^{\prime} - \epsilon_2) \rceil}^{\lfloor N(p^{\prime} + \epsilon_2) \rfloor}  
\sum_{\substack{i_1, \ldots, i_m \in [N]:\\ i_j< i_{j'} \forall j'>j}}   P[\sum_{j\in[m]} Y_{i_j} \leq N(p^{\prime} - \epsilon_2)   (\frac{1}{2} - \epsilon_3)]  P[I_1 = i_1, \ldots, I_m = i_m| M_D = m] \\
&\leq  \sum_{m = \lceil N(p^{\prime} - \epsilon_2) \rceil}^{\lfloor N(p^{\prime} + \epsilon_2) \rfloor}  
\sum_{\substack{i_1, \ldots, i_m \in [N]:\\ i_j< i_{j'} \forall j'>j}}  e^{-{\epsilon_3}^2 \lceil N(p^{\prime} - \epsilon_2) \rceil}  P[I_1 = i_1, \ldots, I_m = i_m| M_D = m] \\
% &\leq  \sum_{m = \lceil N(q - \epsilon_2) \rceil}^{\lfloor N(q + \epsilon_2) \rfloor}  
% \sum_{\substack{i_1, \ldots, i_m \in [N]\\ i_j\neq i_{j'}}} \notag \\
% & \hspace{1cm}  P[\sum_{j\in[m]} (1-A_{i_j}) \leq N(q - \epsilon_2)   (\frac{1}{2} - \epsilon_3)] \notag \\ 
% & \hspace{2cm} \times P[I_1 = i_1, \ldots, I_m = i_m| M_D = m] \\
% &\leq  \sum_{m = \lceil N(p - \epsilon_2) \rceil}^{\lfloor N(p + \epsilon_2) \rfloor}   e^{-{\epsilon_2}^2 \lceil N(p - \epsilon_2) \rceil} 
% \sum_{\substack{i_1, \ldots, i_m \in [N]\\ i_j\neq i_{j'}}}   \notag \\ 
% & \hspace{2cm} \times P[I_1 = i_1, \ldots, I_m = i_m| M_D = m] \\
&\leq  \sum_{m = \lceil N(p^{\prime} - \epsilon_2) \rceil}^{\lfloor N(p^{\prime} + \epsilon_2) \rfloor}   e^{-{\epsilon_3}^2 \lceil N(p^{\prime} - \epsilon_2) \rceil}  \\
& \leq  (2N \epsilon_2)  e^{-{\epsilon_3}^2 \lceil N(p^{\prime} - \epsilon_2) \rceil} \label{eq:lastineq}
\end{align}
where the last inequality follows from,
\begin{equation}
\lfloor N(p^{\prime} + \epsilon_2) \rfloor - \lceil N(p^{\prime} - \epsilon_2) \rceil \leq 2N \epsilon_2 
\end{equation}
We will now show that \eqref{eq:lastineq} approaches $0$ as $N\rightarrow \infty$,
\begin{align}
\lim_{N\rightarrow \infty} (2N \epsilon_2)  e^{-{\epsilon_2}^2 \lceil N(p^{\prime} - \epsilon_2) \rceil} \rightarrow 0  \label{eq:conv}
\end{align}
% To do so, let $\bar{p} := 1-p$ in \eqref{eq:q}, then, 
To do so, let $\bar{p}:=1-\frac{\alpha}{N-1}$, then, \eqref{eq:lastineq} can be represented as, 
% Then, from \eqref{eq:q} and \eqref{eq:lastineq}, we have that,
% using which \eqref{eq:lastineq}, we have 
\begin{align}
(2N \epsilon_2)  e^{-{\epsilon_2}^2 \lceil N(p^{\prime} - \epsilon_2) \rceil} 
& \leq (2N \epsilon_2)  e^{-{\epsilon_2}^2 ( N(p^{\prime} - \epsilon_2) -1) } \\
& = \frac{2N \epsilon_2}{e^{{\epsilon_2}^2 \left(  N\left((1-\theta)(1-\bar{p}^{N-1}) - \epsilon_2\right) -1\right)}} \label{eq:main} \\
& \triangleq \frac{f(N)}{g(N)} \label{eq:fun}
\end{align}
% \begin{align}
% (2N \epsilon_2)  e^{-{\epsilon_2}^2 \lceil N(1-\bar{p}^N - \epsilon_2) \rceil} 
% & = \frac{2N \epsilon_2}{e^{{\epsilon_2}^2 \lceil N(1-\bar{p}^N - \epsilon_2) \rceil}} \label{eq:main} \\
% & \triangleq \frac{f(N)}{g(N)} \label{eq:fun}
% \end{align}
where we define two functions $f(N)$ and $g(N)$ in \eqref{eq:fun} to represent the numerator and denominator of \eqref{eq:main}. 
It can be observed that both $f(N)\rightarrow \infty$ and $g(N)\rightarrow \infty$ as $N\rightarrow \infty$. Then, from L'Hopital's Rule, one can find that,
\begin{align}
& \lim_{N\rightarrow \infty} \frac{f(N)}{g(N)}  
 = \lim_{N\rightarrow \infty} \frac{f'(N)}{g'(N)}=0
%& = \lim_{N\rightarrow \infty}\frac{2\epsilon_2 e^{-\epsilon_{2}^2\left(N\left(1-\bar{p}^{N-1}-\epsilon_2\right)-1\right)}}{\epsilon_{2}^{2}\left\{\left(1-\theta\right)\left(1-\bar{p}^{N-1}\left(1+N\ln\left(1-\frac{\alpha}{N-1}\right)+\frac{N\alpha}{N-1-\alpha}\right)\right)-\epsilon_2\right\}}\\
%&  \rightarrow 0 \quad \text{ as } \quad N\rightarrow \infty, 
\end{align}
where,
\begin{equation}
f'(N)=2\epsilon_2
\end{equation}
and
\begin{align}
g'(N)&= e^{\epsilon_{2}^2\left(N\left(1-\bar{p}^{N-1}-\epsilon_2\right)-1\right)}\epsilon_{2}^{2}\Big\{\left(1-\theta\right)\left(1-\bar{p}^{N-1} \notag \right.  \\
& \left.\times \left(1+N\ln\left(1-\frac{\alpha}{N-1}\right)+\frac{N\alpha}{N-1-\alpha}\right)\right)-\epsilon_2\Big\}
\end{align}
which completes the proof of \eqref{eq:conv}.

% {\color{red} TO DO: Revise the term $\frac{1}{(1-\bar{p}^N - N^2 \bar{p}^{N-1})e^{N(1-\bar{p}^N)}}$ and show it goes to 0.} 

By combining \eqref{eq:conv} with \eqref{eq:lastineq}, we find that the last term in \eqref{eq:bound} also approaches $0$ as $N\rightarrow \infty$,
\begin{equation}\label{eq:limit3}
P[\bar{X} \leq \frac{p^{\prime}}{2}-\epsilon_1, p^{\prime} - \epsilon_2 \leq \bar{M}_D  \leq  p^{\prime} + \epsilon_2] \rightarrow 0 \text{ as } N\rightarrow \infty.  
\end{equation}
Then, by combining \eqref{eq:bound1}, \eqref{eq:bound2}, and  \eqref{eq:limit3} with \eqref{eq:bound}, we have, 
\begin{align}
& P[X \leq N(\frac{p^{\prime}}{2}-\epsilon_1)] \notag \\
& \qquad = P[\bar{X} \leq \frac{p^{\prime}}{2}-\epsilon_1] \notag \\
& \qquad \leq P[\bar{M}_D \leq p^{\prime} - \epsilon_2] + 
P[\bar{M}_D \geq p^{\prime} + \epsilon_2] 
\notag \\ 
& \qquad  \quad + P[\bar{X} \leq \frac{p^{\prime}}{2}-\epsilon_1 ,  p^{\prime} - \epsilon_2 \leq \bar{M}_D  \leq  p^{\prime} + \epsilon_2] \\
& \qquad \rightarrow 0 \; \text{ as } \; N\rightarrow \infty
\end{align}
Therefore, 
% $X = \sum_{i\in[N]} X_i^{(l)}$, which represents the number of honest users that participate in the aggregated model at a given location $\ell\in[d]$, approaches $\frac{Nq}{2}$. 
$X = \sum_{i\in[N]} X_i(\ell)$, which represents the number of honest users that participate in the aggregated gradient at any given location $\ell\in[d]$, is $\frac{Np^{\prime}}{2}$ with probability approaching to $1$ as the number of users $N\rightarrow \infty$. 

Next, note that  
\begin{equation}
\frac{Np^{\prime}}{2} > \frac{N}{2} (1-\theta)(1-e^{-\alpha}) = T
\end{equation}
which follows from 
% \begin{equation}
% (1-\frac{\alpha}{N})^N\leq e^{-\alpha},  
% \end{equation}
\begin{equation}
\left(1-\frac{\alpha}{N-1}\right)^{N-1} = e^{\left(N-1\right)\ln{\left(1-\frac{\alpha}{N-1}\right)}} 
% \leq e^{N\frac{\alpha}{N})}  
< e^{-\alpha},  
\end{equation}
as $\ln{(1-\frac{\alpha}{N-1})}< -\frac{\alpha}{N-1}$ for all $\alpha\in(0,1]$ and $N\geq 2$. Finally, 
\begin{equation}
\lim_{\alpha\rightarrow 0} \frac{1-e^{-\alpha}}{\alpha} =  1 
\end{equation}
which follows from L'Hopital's Rule, and therefore, for $\alpha\ll 1$,  
% {\color{red} 
% TO DO: Complete the missing steps in (95) and (96). 
% }
% hence
% \begin{align}
% 1- (1-\frac{\alpha}{N})^N\geq 1- e^{-\alpha}  
% \end{align}

% Note that, from \eqref{limit}, for sufficiently large $N$,  
% {\color{orange}\begin{align}
% \frac{Np}{2} 
% % & = \frac{N(1-(1-p)^N)}{2} \\
% & \approx \frac{N(1-e^{-\alpha}-\delta)}{2}. 
% \end{align}}
% for any sufficiently small $\delta >0$. 
% For  $\alpha \ll 1$, the first order Taylor's series approximation to $e^{-\alpha}$ leads to,
% \begin{align}
% e^{-\alpha} \approx 1 - \alpha
% \end{align}
% and therefore,
\begin{equation}
T = 
% & = \frac{N(1-(1-p)^N)}{2} \\
% \approx \frac{N (1-(1-\frac{K}{d})} 
% \approx \frac{NK}{2d}
 \frac{N}{2} (1-\theta)(1-e^{-\alpha}) \rightarrow \frac{N\alpha}{2}(1-\theta)  
%  =  \alpha(N-A) 
\end{equation}
% where $A = \frac{N}{2}$, 
which completes our proof. 
The case for any $A<\frac{N}{2}$ follows the same steps. 
Hence, in a network of $N$ users where $A = \gamma N \leq \frac{N}{2}$ users are adversarial for some $\gamma\in(0,0.5)$, \namespace provides a privacy guarantee of $T = (1-e^{-\alpha})(1-\theta)(1-\gamma)N$, which approaches $\alpha(1-\theta)(1-\gamma)N$ as the compression ratio $\alpha$ becomes smaller.

\subsection{Proof of Theorem~\ref{thm:convergence}}\label{app:convergence}
Let $D_{i}^{(t)}$ be a Bernoulli random variable that defines whether user $i \in [N]$ drops out at time $t$ which is given as follows:
\begin{equation}
D_{i}^{(t)} = \left \{ \begin{matrix} 0 & \text{with probability} & \theta\\
1 & \text{otherwise} & \end{matrix}\right.    
\end{equation}

Let $M_{i}^{(t)}(\ell)$ be a Bernoulli random variable that defines whether a location $\ell \in [d]$ is selected by user $i$ to be sent to the server at round $t$. Therefore,
\begin{equation}
M_{i}^{(t)}(\ell) = \left \{ \begin{matrix} 1 & \text{with probability} & p\\
0 & \text{otherwise} & \end{matrix}\right.
\end{equation}
where $p$ is the probability that a location will be chosen by user $i \in [N]$ as defined in \eqref{tildep}. For simplicity, in this section, the time index $t$ is used to represent both the local and global training rounds. In particular, $t \in \left\{0,E,2E,\cdots \cdots, J\right\}$ represents a global round where $\frac{J}{E} \in \mathbb{Z}^{+}$ and any other time index represents a local training round.

Let $\tau_{E} \triangleq \left\{0,E,2E,\cdots \cdots, J\right\}$ represent the global rounds. As shown in \eqref{aggregate}, at each global round, the server aggregates the local gradients of the users and sends the updated global model back to the users. Users then synchronize their local models with the updated global model. As such, we call each global iteration $t \in \tau_E$ a synchronization step, where the local models of all the users are synchronized to the updated global model. Then the local model of $\name$ can be expressed as follows: 
\begin{align}
\mathbf{v}_{i}^{(t+1)}(\ell) &=\mathbf{w}_{i}^{(t)}(\ell)-\eta^{(t)} \nabla F_{i}^{(\ell)}\left(\mathbf{w}_{i}^{(t)}, \xi_{i}^{(t)}\right)  \\
&=\mathbf{w}_{i}^{(t_{o})}(\ell)- \sum_{k=t_{o}}^{t}\mathbf{z}_{i}^{(k)}(\ell)\\
\mathbf{w}_{i}^{(t+1)}(\ell) &=\left\{\begin{array}{ll}
\mathbf{v}_{i}^{(t+1)}(\ell)  \hspace{2.8cm} \text { if } t+1 \notin \tau_{E} \\
\mathbf{w}_{i}^{(t_{o})}(\ell)\\- \sum_{i \in [N]}D_i^{(t)}M_{i}^{(t)}(\ell)Q\left(\frac{\beta_{i}}{p^{\prime}}\sum_{k=t_{o}}^{t}\mathbf{z}_{i}^{(k)}(\ell)\right) \\ \hspace{4cm} \text { if } t+1 \in \tau_{E}
\end{array}\right.
\end{align}
$\forall \ell \in [d]$ where we define
\begin{equation}
\mathbf{z}_{i}^{(k)}(\ell) := \eta^{(k)} \nabla F_{i}^{(\ell)}\left(\mathbf{w}_{i}^{(k)}, \xi_{i}^{(k)}\right)
\end{equation}
and we use the definition of $p^{\prime}$ from \eqref{p_prime}. $\nabla F_i^{(\ell)}$ represents the $\ell^{th}$ element of the local gradient $\nabla F_i$. $t_{o}$ is the previous synchronization step such that $t_{o}=t-E+1$ where all the local models were equal (synchronized). %$\mathbf{w}_{i}^{(t_{o})}=\overline{\mathbf{w}}^{(t_{o})}$.

In our analysis, we further define two virtual sequences: 
\begin{equation}
\overline{\mathbf{v}}^{(t)}=\sum_{i=1}^{N}\beta_{i}  \mathbf{v}_{i}^{(t)}
\end{equation}
\begin{equation}
\overline{\mathbf{w}}^{(t)}=\sum_{i=1}^{N}\beta_{i} \mathbf{w}_{i}^{(t)}
\end{equation} 
We next define:
\begin{equation}
\mathbf{g}^{(t)} =\sum_{i=1}^{N}\beta_{i} \nabla F_{i}\left(\mathbf{w}_{i}^{(t)}, \xi_{i}^{(t)}\right)
\end{equation}
Therefore,
\begin{equation}
\overline{\mathbf{v}}^{(t+1)}=\overline{\mathbf{w}}^{(t)}-\eta^{(t)}\mathbf{g}^{(t)}
\end{equation}
where 
$\overline{\mathbf{w}}^{(0)}=\overline{\mathbf{v}}^{(0)}$ is the initial model for all users. Note that $\overline{\mathbf{v}}^{(t+1)}=\overline{\mathbf{w}}^{(t+1)}$ when $t+1 \notin \tau_{E}$ and $\overline{\mathbf{v}}^{(t+1)} \neq \overline{\mathbf{w}}^{(t+1)}$ when $t+1 \in \tau_{E}$. 
Next, we define:
\begin{equation}
\overline{\mathbf{g}}^{(t)}=\sum_{i=1}^{N}\beta_{i} \nabla F_{i}\left(\mathbf{w}_{i}^{(t)}\right)
\end{equation}
Therefore,
\begin{equation}
\mathbb{E} [\mathbf{g}^{(t)}]=\overline{\mathbf{g}}^{(t)}  \end{equation}

The proof relies on the following two key lemmas. 
\begin{lemma}[Unbiased Estimator]\label{lemma 1}
If $t+1 \in \tau_{E}$, then the following holds:
\begin{align}
\mathbb{E}_{D,M,Q}\left[\overline{\mathbf{w}}^{(t+1)}\right]=\overline{\mathbf{v}}^{(t+1)}
\end{align}
\end{lemma}
\begin{proof}
\begin{align}
&\mathbb{E}_{Q} \left[\sum_{i=1}^{N}D_{i}^{(t)}M_{i}^{(t)}(\ell) Q\left(\beta_{i}\frac{1}{p^{\prime}}\sum_{k=t_o}^{t}\mathbf{z}_{i}^{k}(\ell)\right)\right]\notag \\
&= \sum_{i=1}^{N}\mathbb{E}_{Q} \left[D_{i}^{(t)}M_{i}^{(t)}(\ell)Q\left(\beta_{i}\frac{1}{p^{\prime}} \sum_{k=t_o}^{t}\mathbf{z}_{i}^{k}(\ell)\right)\right] \\
&= \sum_{i=1}^{N} D_{i}^{(t)}M_{i}^{(t)}(\ell)\mathbb{E}_{Q}\left[Q\left(\beta_{i}\frac{1}{p^{\prime}} \sum_{k=t_o}^{t}\mathbf{z}_{i}^{k}(\ell)\right)\right] \\
&= \sum_{i=1}^{N}\beta_{i}D_{i}^{(t)}M_{i}^{(t)}(\ell)\frac{1}{p^{\prime}}\sum_{k=t_o}^{t}\mathbf{z}_{i}^{k}(\ell) \label{sto_quant}
\end{align}
where in \eqref{sto_quant}, we leverage the unbiasedness of stochastic quantization: 
\begin{equation}
\mathbb{E}_{Q} \left[Q(x)\right]=x \label{unbiasedness}
\end{equation}
as shown in \cite[Lemma 1]{BREA}. Thus,
\begin{align}
&\mathbb{E}_{D,M,Q}\left[\overline{\mathbf{w}}^{(t+1)}(\ell)\right]\notag\\
&=\mathbb{E}_{D,M,Q}\left[\overline{\mathbf{w}}^{(t_o)}(\ell)-\sum_{i=1}^{N}D_{i}^{(t)}M_{i}^{(t)}(\ell)Q\left(\frac{\beta_{i}}{p^{\prime}}\sum_{k=t_o}^{t}\mathbf{z}_{i}^{k}(\ell)\right)\right] \notag\\
&=\mathbb{E}_{D,M}\left[\overline{\mathbf{w}}^{(t_o)}(\ell)-\sum_{i=1}^{N}D_{i}^{(t)}M_{i}^{(t)}(\ell)\mathbb{E}_{Q}\left[Q\left(\frac{\beta_{i}}{p^{\prime}} \sum_{k=t_o}^{t}\mathbf{z}_{i}^{k}(\ell)\right)\right]\right] \notag \\
&=\mathbb{E}_{D,M}\left[\overline{\mathbf{w}}^{(t_o)}(\ell)-\sum_{i=1}^{N}D_{i}^{(t)}M_{i}^{(t)}(\ell)\frac{\beta_{i}}{p^{\prime}}  \sum_{k=t_o}^{t}\mathbf{z}_{i}^{k}(\ell)\right] \\
&=\overline{\mathbf{w}}^{(t_o)}(\ell)-\sum_{i=1}^{N}\mathbb{E}_{D}\left[D_{i}^{(t)}\right]\mathbb{E}_{M}\left[M_{i}^{(t)}(\ell)\right]\frac{\beta_{i}}{p^{\prime}}  \sum_{k=t_o}^{t}\mathbf{z}_{i}^{k}(\ell)\notag \\
&=\overline{\mathbf{w}}^{(t_o)}(\ell)-\sum_{i=1}^{N}(1-\theta)p \frac{\beta_{i}}{(1-\theta)p}  \sum_{k=t_o}^{t}\mathbf{z}_{i}^{k}(\ell)\\
&=\overline{\mathbf{w}}^{(t_o)}(\ell)-\sum_{i=1}^{N}\beta_{i} \sum_{k=t_o}^{t}\mathbf{z}_{i}^{k}(\ell)\\
&=\overline{\mathbf{v}}^{(t+1)}(\ell)
\end{align}
Therefore,
\begin{align}
\mathbb{E}_{D,M,Q}\left[\overline{\mathbf{w}}^{(t+1)}\right]=\overline{\mathbf{v}}^{(t+1)}
\end{align}
which completes the proof.
\end{proof}
\vspace{-0.3cm}
\begin{lemma}[Variance of $\overline{\mathbf{w}}^{(t+1)}$] \label{lemma 2}
If $t+1 \in \tau_{E}$, $\eta^{(t)}$ is non-increasing with $t$ and $\eta^{(t)} \leq 2\eta^{(t+E)}$ $\forall t \geq 0$, then
\begin{align}
&\mathbb{E}_{D,M,Q,\xi}\left\|\overline{\mathbf{w}}^{(t+1)}-\overline{\mathbf{v}}^{(t+1)}\right\|^2 \notag\\
&\leq \frac{Ndp^{\prime}}{4c^2} + %\sum_{i=1}^{N} \left\{ \beta_{i}^2  + \sum_{j=1,j \neq i}^{N}  \beta_{i}\beta_{j} \right\}%
4\left(\eta^{(t)}\right)^{2}E^2G^2\sum_{i=1}^{N} \left( \beta_{i}^2 \left(\frac{1}{p^{\prime}}-1\right)  \right. \left.+ \sum_{j=1,j \neq i}^{N}  \beta_{i}\beta_{j}\left(\frac{\Tilde{p}}{\left(p^{\prime}\right)^{2}}-1\right) \right)
\end{align}
\end{lemma}
\vspace{-0.3cm}\begin{proof}
\begin{align}
&\left\|\overline{\mathbf{w}}^{(t+1)}-\overline{\mathbf{v}}^{(t+1)}\right\|^2
=\sum_{\ell=1}^{d}\left|\overline{\mathbf{w}}^{(t+1)}(\ell)-\overline{\mathbf{v}}^{(t+1)}(\ell)\right|^2 \label{vector_norm}\\
&=\sum_{\ell=1}^{d}\left|\overline{\mathbf{w}}^{(t_o)}(\ell)-\sum_{i=1}^{N}D_{i}^{(t)}M_{i}^{(t)}(\ell)Q\left(\frac{\beta_{i}}{p^{\prime}} \sum_{k=t_o}^{t}\mathbf{z}_{i}^{(k)}(\ell)\right)  \right.\left.-\overline{\mathbf{w}}^{(t_o)}(\ell)+ \sum_{i=1}^{N}\beta_{i} \sum_{k=t_o}^{t}\mathbf{z}_{i}^{(k)}(\ell)\right|^2\\
&=\sum_{\ell=1}^{d}\left|\sum_{i=1}^{N}\left(D_{i}^{(t)}M_{i}^{(t)}(\ell)Q\left(\frac{\beta_{i}}{p^{\prime}} \sum_{k=t_o}^{t}\mathbf{z}_{i}^{(k)}(\ell)\right) \right. \right. \left. \left. -\beta_{i} \sum_{k=t_o}^{t}\mathbf{z}_{i}^{(k)}(\ell)\right)\right|^2\\
&\leq \sum_{\ell=1}^{d}\left(\left|\sum_{i=1}^{N}D_{i}^{(t)}M_{i}^{(t)}(\ell)Q\left(\frac{\beta_{i}}{p^{\prime}} \sum_{k=t_o}^{t}\mathbf{z}_{i}^{(k)}(\ell)\right) \right.  \right. \left.-\sum_{i=1}^{N}\frac{\beta_{i}D_{i}^{(t)}M_{i}^{(t)}(\ell)}{p^{\prime}} \sum_{k=t_o}^{t}\mathbf{z}_{i}^{(k)}(\ell)\right|^2 \notag \\
&  +\left|\sum_{i=1}^{N}\frac{\beta_{i}D_{i}^{(t)}M_{i}^{(t)}(\ell)}{p^{\prime}} \notag \right. \left. \sum_{k=t_o}^{t}\mathbf{z}_{i}^{(k)}(\ell) -\sum_{i=1}^{N}\beta_{i} \sum_{k=t_o}^{t}\mathbf{z}_{i}^{(k)}(\ell)\right|^2 \\
&+2\left(\sum_{i=1}^{N}D_{i}^{(t)}M_{i}^{(t)}(\ell) \right.  \left. Q\left(\frac{\beta_{i}}{p^{\prime}} \sum_{k=t_o}^{t}\mathbf{z}_{i}^{(k)}(\ell)\right)-\sum_{i=1}^{N}\frac{\beta_{i}D_{i}^{(t)}M_{i}^{(t)}(\ell)}{p^{\prime}}\sum_{k=t_o}^{t}\mathbf{z}_{i}^{(k)}(\ell)\right) \notag  \\
& \left. \times \left(\sum_{i=1}^{N}\frac{\beta_{i}D_{i}^{(t)}M_{i}^{(t)}(\ell)}{p^{\prime}}\sum_{k=t_o}^{t}\mathbf{z}_{i}^{(k)}(\ell)  
 -\sum_{i=1}^{N}\beta_{i} \sum_{k=t_o}^{t}\mathbf{z}_{i}^{(k)}(\ell)\right)\right) \label{sto_bound}
\end{align}
% where \eqref{vector_norm} follows from the fact that for any vector $\mathbf{a}$ of length $d$ with elements $a_1,\cdots  , a_d$, 
% \begin{equation}
% \left\|\mathbf{a}\right\|^2=a_1^{2}+ \cdots +a_d^{2}
% \end{equation}
% \begin{equation}
% \left\|\mathbf{a}\right\|^2=a_1^{2}+ \cdots \cdots +a_d^{2}
% \end{equation}
where the last term in \eqref{sto_bound} vanishes in expectation since
\begin{equation}
  \mathbb{E}_Q\left[Q\left(\frac{\beta_{i}}{p^{\prime}} \sum_{k=t_o}^{t}\mathbf{z}_{i}^{(k)}(\ell)\right)\right]=\frac{\beta_{i}}{p^{\prime}} \sum_{k=t_o}^{t}\mathbf{z}_{i}^{(k)}(\ell) 
\end{equation}
% as mentioned in \eqref{unbiasedness}.
as shown in \eqref{unbiasedness}.
% We define the term $D_{i}^{(t)}M_{i}^{(t)}(\ell)$ as follows:
We next define:
\begin{equation}
M_{D,i}^{(t)}(\ell):=D_{i}^{(t)}M_{i}^{(t)}(\ell)
\end{equation}
The first term in \eqref{sto_bound} can be bounded as follows:
\begin{align}
&\sum_{\ell=1}^{d} \left| \sum_{i=1}^{N}M_{D,i}^{(t)}(\ell)\left(Q\left(\frac{\beta_{i}}{p^{\prime}} \sum_{k=t_o}^{t}\mathbf{z}_{i}^{(k)}(\ell)\right) -\frac{\beta_{i}}{p^{\prime}} \sum_{k=t_o}^{t}\mathbf{z}_{i}^{(k)}(\ell)\right)\right| ^2 \notag\\
&=\sum_{\ell=1}^{d} \sum_{i=1}^{N}\left(\left(M_{D,i}^{(t)}(\ell)\right)^{2}\left(Q\left(\frac{\beta_{i}}{p^{\prime}} \sum_{k=t_o}^{t}\mathbf{z}_{i}^{(k)}(\ell)\right)- \frac{\beta_{i}}{p^{\prime}} \sum_{k=t_o}^{t}\mathbf{z}_{i}^{(k)}(\ell)\right)^2 \notag \right.\\
&\qquad +\sum_{j=1,j \neq i}^{N}M_{D,i}^{(t)}(\ell)\left(Q\left(\frac{\beta_{i}}{p^{\prime}} \sum_{k=t_o}^{t}\mathbf{z}_{i}^{(k)}(\ell)\right) -\frac{\beta_{i}}{p^{\prime}} \sum_{k=t_o}^{t}\mathbf{z}_{i}^{(k)}(\ell)\right) \notag \\
&\qquad  \left.\times M_{D,j}^{(t)}(\ell)\left(Q\left(\frac{\beta_{j}}{p^{\prime}} \sum_{k=t_o}^{t}\mathbf{z}_{j}^{(k)}(\ell)\right) -\frac{\beta_{j}}{p^{\prime}} \sum_{k=t_o}^{t}\mathbf{z}_{j}^{(k)}(\ell)\right)\right)\label{vanish}
%&\leq \frac{d}{4c^2}
\end{align}
The second term in \eqref{vanish} vanishes in expectation due to the unbiasedness of quantization as shown in  \eqref{unbiasedness}.
%\begin{align}
    %\mathbb{E}_Q\left[Q\left(\frac{\beta_{j}}{p} \sum_{k=t_o}^{t}\mathbf{z}_{j}^{(k)}(\ell)\right)\right]= \frac{\beta_{j}}{p} \sum_{k=t_o}^{t}\mathbf{z}_{j}^{(k)}(\ell)
%\end{align}as shown in \cite[Lemma 1]{BREA}. 
Hence,
\begin{align}
&\mathbb{E}\left[\sum_{\ell=1}^{d} \left| \sum_{i=1}^{N}M_{D,i}^{(t)}(\ell)\left(Q\left(\frac{\beta_{i}}{p^{\prime}} \sum_{k=t_o}^{t}\mathbf{z}_{i}^{(k)}(\ell)\right) -\frac{\beta_{i}}{p^{\prime}} \sum_{k=t_o}^{t}\mathbf{z}_{i}^{(k)}(\ell)\right)\right| ^2\right] \notag\\  
&=\sum_{\ell=1}^{d} \sum_{i=1}^{N}\mathbb{E}_{D}\left[\left(D_{i}^{(t)}\right)^2\right]\mathbb{E}_{M}\left[\left(M_{i}^{(t)}(\ell)\right)^{2}\right]\notag  \\
& \hspace{0.5cm}\times \mathbb{E}_{Q}\left[\left(Q\left(\frac{\beta_{i}}{p^{\prime}} \sum_{k=t_o}^{t}\mathbf{z}_{i}^{(k)}(\ell)\right)- \frac{\beta_{i}}{p^{\prime}} \sum_{k=t_o}^{t}\mathbf{z}_{i}^{(k)}(\ell)\right)^2\right]\\
& \leq \sum_{\ell=1}^{d} \sum_{i=1}^{N}(1-\theta)p\frac{1}{4c^2} \label{bounded_var}\\
&=\frac{Ndp^{\prime}}{4c^2}
\end{align}
where \eqref{bounded_var} follows from the bounded variance property of quantized gradient estimator as shown in \cite[Lemma 1]{BREA}.
Next, for the second term in \eqref{sto_bound},
\begin{align}
&\left|\sum_{i=1}^{N}\left(M_{D,i}^{(t)}(\ell)\frac{\beta_{i}}{p^{\prime}} \sum_{k=t_o}^{t}\mathbf{z}_{i}^{(k)}(\ell)-\beta_{i} \sum_{k=t_o}^{t}\mathbf{z}_{i}^{(k)}(\ell)\right)\right|^2 \notag\\
&=\sum_{i=1}^{N}\left(\left(M_{D,i}^{(t)}(\ell)\frac{\beta_{i}}{p^{\prime}} \sum_{k=t_o}^{t}\mathbf{z}_{i}^{(k)}(\ell)-\beta_{i} \sum_{k=t_o}^{t}\mathbf{z}_{i}^{(k)}(\ell)\right)^2 \notag \right. \\
& \hspace{1 cm} + \sum_{j=1,j \neq i}^{N} \left( M_{D,i}^{(t)}(\ell)\frac{\beta_{i}}{p^{\prime}} \sum_{k=t_o}^{t}\mathbf{z}_{i}^{(k)}(\ell)-\beta_{i} \sum_{k=t_o}^{t}\mathbf{z}_{i}^{(k)}(\ell)\right)  \left.\times \left(M_{D,j}^{(t)}(\ell)\frac{\beta_{j}}{p^{\prime}} \sum_{k=t_o}^{t}\mathbf{z}_{j}^{(k)}(\ell)-\beta_{j} \sum_{k=t_o}^{t}\mathbf{z}_{j}^{(k)}(\ell)\right) \right)\\
&=\sum_{i=1}^{N}\left(\left(\beta_{i}M_{D,i}^{(t)}(\ell)\frac{1}{p^{\prime}} \sum_{k=t_o}^{t}\mathbf{z}_{i}^{(k)}(\ell)\right)^2 + \left(\beta_{i}\sum_{k=t_o}^{t}\mathbf{z}_{i}^{(k)}(\ell)\right)^2   \right. - 2\beta_{i}M_{D,i}^{(t)}(\ell)\frac{1}{p^{\prime}} \left(\sum_{k=t_o}^{t}\mathbf{z}_{i}^{(k)}(\ell)\right) \beta_{i}\left(\sum_{k=t_o}^{t}\mathbf{z}_{i}^{(k)}(\ell)\right)   \notag \\
& + \sum_{j=1,j \neq i}^{N} \!\! \left(\frac{\beta_{i}\beta_{j}}{(p^{\prime})^2} M_{D,i}^{(t)}(\ell) \left(\sum_{k=t_o}^{t}\mathbf{z}_{i}^{(k)}(\ell)\right) M_{D,j}^{(t)}(\ell)  \notag \right. \left(\sum_{k=t_o}^{t}\mathbf{z}_{j}^{(k)}(\ell)\right)-  \frac{\beta_{i}\beta_{j}}{p^{\prime}} M_{D,i}^{(t)}(\ell) \left(\sum_{k=t_o}^{t}\mathbf{z}_{i}^{(k)}(\ell)\right) \left(\sum_{k=t_o}^{t}\mathbf{z}_{j}^{(k)}(\ell)\right) \\
&\hspace{1.5 cm}  - \frac{\beta_{i}\beta_{j}}{p^{\prime}}M_{D,j}^{(t)}(\ell) \left(\sum_{k=t_o}^{t}\mathbf{z}_{i}^{(k)}(\ell)\right) \left(\sum_{k=t_o}^{t}\mathbf{z}_{j}^{(k)}(\ell)\right) \left. \left. + \beta_{i}\beta_{j} \left(\sum_{k=t_o}^{t}\mathbf{z}_{i}^{(k)}(\ell)\right) \left(\sum_{k=t_o}^{t}\mathbf{z}_{j}^{(k)}(\ell)\right)\right) \right) \\ 
% & \hspace{2 cm}- \frac{\beta_{i}\beta_{j}}{p^{\prime}}M_{D,j}^{(t)}(\ell) \left(\sum_{k=t_o}^{t}\mathbf{z}_{i}^{(k)}(\ell)\right) \left(\sum_{k=t_o}^{t}\mathbf{z}_{j}^{(k)}(\ell)\right) \notag\\
% &\left. \left. + \beta_{i}\beta_{j} \left(\sum_{k=t_o}^{t}\mathbf{z}_{i}^{(k)}(\ell)\right) \left(\sum_{k=t_o}^{t}\mathbf{z}_{j}^{(k)}(\ell)\right)\right) \right)\\
& = \sum_{i=1}^{N}\left(\beta_{i}^2 \left(\sum_{k=t_o}^{t}\mathbf{z}_{i}^{(k)}(\ell)\right)^2 \left(\frac{\left(M_{D,i}^{(t)}(\ell)\right)^2}{(p^{\prime})^2}  + 1 - \frac{2}{p^{\prime}} M_{D,i}^{(t)}(\ell)\right)  \notag \right. \\
&  + \sum_{j=1,j \neq i}^{N}\beta_{i}\beta_{j} \left(\frac{M_{D,i}^{(t)}(\ell)  
 M_{j}^{(t)}(\ell)}{(p^{\prime})^2} -  \frac{M_{D,i}^{(t)}(\ell)+M_{D,j}^{(t)}(\ell)}{p^{\prime}}+1\right)  \left.  \left(\sum_{k=t_o}^{t}\mathbf{z}_{i}^{(k)}(\ell)\right) \left(\sum_{k=t_o}^{t}\mathbf{z}_{j}^{(k)}(\ell)\right)  \right) \label{line_b4_ex}
%& \leq \sum_{i=1}^{N}\left(\beta_{i}^2 \left\|\sum_{k=t_o}^{t}\mathbf{z}_{i}^{(k)}(\ell)\right\|^2 \left(\frac{1}{p^2}\left\|M_{i}^{(t)}(\ell)\right\|^2  + 1 - \frac{2}{p} M_{i}^{(t)}(\ell)\right)  \notag \right. \\
%& \hspace{1.5 cm}+ \sum_{j=1,j \neq i}^{N} \left(\frac{\beta_{i}\beta_{j}}{p^2} \left(\frac{\left\|M_{i}^{(t)}(\ell)\sum_{k=t_o}^{t}\mathbf{z}_{i}^{(k)}(\ell)\right\|^2}{2} \notag \right. \right. \\ 
%&+\left. \frac{\left\|M_{j}^{(t)}(\ell)\sum_{k=t_o}^{t}\mathbf{z}_{j}^{(k)}(\ell)\right\|^2}{2}\right) - \beta_{i}\beta_{j} \left(\frac{M_{i}^{(t)}(\ell)}{p} + \frac{M_{j}^{(t)}(\ell)}{p}-1\right) \notag\\ 
%&\hspace{1.5 cm} \times \left. \left.  \left(\frac{ \left\|\sum_{k=t_o}^{t}\mathbf{z}_{i}^{(k)}(\ell)\right\|^2 + \left\| \sum_{k=t_o}^{t}\mathbf{z}_{j}^{(k)}(\ell)\right\|^2}{2}\right) \right) \right) \label{AM-GM_inequality}
\end{align}
%where AM-GM inequality is used in %\eqref{AM-GM_inequality}.
Next, by taking the expectation of  \eqref{line_b4_ex},
\begin{align}
% &\mathbb{E}\left[\sum_{i=1}^{N}\left(\beta_{i}^2 \left(\sum_{k=t_o}^{t}\mathbf{z}_{i}^{(k)}(\ell)\right)^2 \left(\frac{\left(M_{D,i}^{(t)}(\ell)\right)^2}{(p^{\prime})^2}  + 1 - \frac{2M_{D,i}^{(t)}(\ell)}{p^{\prime}} \right)  \notag \right. \right. \\
% &  + \sum_{j=1,j \neq i}^{N}\beta_{i}\beta_{j} \left(\frac{M_{D,i}^{(t)}(\ell)  
%  M_{D,j}^{(t)}(\ell)}{(p^{\prime})^2} -  \frac{M_{D,i}^{(t)}(\ell)+M_{D,j}^{(t)}(\ell)}{p^{\prime}}+1\right) \notag \\
%  & \left. \left. \hspace{2.8cm}\times \left(\sum_{k=t_o}^{t}\mathbf{z}_{i}^{(k)}(\ell)\right) \left(\sum_{k=t_o}^{t}\mathbf{z}_{j}^{(k)}(\ell)\right)   \right) \right] \notag\\
 &\sum_{i=1}^{N}\left(\beta_{i}^2 \left(\sum_{k=t_o}^{t}\mathbf{z}_{i}^{(k)}(\ell)\right)^2 \left(\frac{1}{(p^{\prime})^2}\mathbb{E}_{D,M}\left[\left(M_{D,i}^{(t)}(\ell)\right)^2\right]  + 1    \right. \right. \left. - \frac{2}{p^{\prime}} \mathbb{E}_{D,M}\left[M_{D,i}^{(t)}(\ell)\right]\right)   \notag\\
 & + \sum_{j=1,j \neq i}^{N}\left(\beta_{i}\beta_{j} \left(\frac{1}{(p^{\prime})^2}   \right. \right. \mathbb{E}_{D,M}\left[M_{D,i}^{(t)}(\ell)  
 M_{D,j}^{(t)}(\ell)\right]   \notag \\ 
&\hspace{0.4cm}  -  \frac{\mathbb{E}_{D,M}\left[M_{D,i}^{(t)}(\ell)\right]}{p^{\prime}} \left. \left. \left. - \frac{\mathbb{E}_{D,M}\left[M_{D,j}^{(t)}(\ell)\right]}{p^{\prime}}+1\right) \left(\sum_{k=t_o}^{t}\mathbf{z}_{i}^{(k)}(\ell)\right)\left(\sum_{k=t_o}^{t}\mathbf{z}_{j}^{(k)}(\ell)\right)  \right) \right) \notag\\
&=\sum_{i=1}^{N}\left(\beta_{i}^2 \left(\sum_{k=t_o}^{t}\mathbf{z}_{i}^{(k)}(\ell)\right)^2 \left(\frac{1}{(p^{\prime})^2}p^{\prime}  + 1   \notag  - \frac{2}{p^{\prime}} p^{\prime}\right)\notag \right.\\
& \qquad + \hspace{-0.2cm}\sum_{j=1,j \neq i}^{N}\hspace{-0.2cm}\beta_{i}\beta_{j} \left(\frac{\Tilde{p}}{(p^{\prime})^2}  -  \frac{2p^{\prime}}{p^{\prime}}+1\right) \left(\sum_{k=t_o}^{t}\mathbf{z}_{i}^{(k)}(\ell)\right)  \left.  \left(\sum_{k=t_o}^{t}\mathbf{z}_{j}^{(k)}(\ell)\right)  \right) \label{E_M}\\
&=\sum_{i=1}^{N}\left(\beta_{i}^2 \left(\sum_{k=t_o}^{t}\mathbf{z}_{i}^{(k)}(\ell)\right)^2 \left(\frac{1}{p^{\prime}}  -1\right) \right. \left. + \sum_{j=1,j \neq i}^{N}\hspace{-0.2cm}\beta_{i}\beta_{j} \left(\frac{\Tilde{p}}{(p^{\prime})^2}  - 1\right) \left(\sum_{k=t_o}^{t}\mathbf{z}_{i}^{(k)}(\ell)\right) \left(\sum_{k=t_o}^{t}\mathbf{z}_{j}^{(k)}(\ell)\right)   \right)\\
& \leq \sum_{i=1}^{N}\left(\beta_{i}^2 \left(\sum_{k=t_o}^{t}\mathbf{z}_{i}^{(k)}(\ell)\right)^2 \left(\frac{1}{p^{\prime}}  -1\right) \right. \left. + \hspace{-0.2cm}\sum_{j=1,j \neq i}^{N}\hspace{-0.2cm}\beta_{i}\beta_{j} \left(\frac{\Tilde{p}}{(p^{\prime})^2}  - 1\right) \left|\left(\sum_{k=t_o}^{t}\mathbf{z}_{i}^{(k)}(\ell)\right) \left(\sum_{k=t_o}^{t}\mathbf{z}_{j}^{(k)}(\ell)\right)  \right| \right) \label{Validate}\\
&\leq \sum_{i=1}^{N}\left(\beta_{i}^2 \left(\sum_{k=t_o}^{t}\mathbf{z}_{i}^{(k)}(\ell)\right)^2 \left(\frac{1}{p^{\prime}}  -1\right)+ \sum_{j=1,j \neq i}^{N}\beta_{i}\beta_{j} \notag \right. \left.  \left(\frac{\Tilde{p}}{(p^{\prime})^2}  - 1\right) \frac{\left(\sum_{k=t_o}^{t} \mathbf{z}_{i}^{(k)}(\ell)\right)^2 + \left(\sum_{k=t_o}^{t} \mathbf{z}_{j}^{(k)}(\ell)\right)^2}{2} \right)\label{line_b4_AM-GM}
\end{align}
where \eqref{E_M} follows from the fact that
\begin{align}
\mathbb{E}_{D,M}\left[M_{D,i}^{(t)}(\ell)\right]&=
\mathbb{E}_{D,M}\left[D_{i}^{(t)}M_{i}^{(t)}(\ell)\right]\\
&=\mathbb{E}_{D}\left[D_{i}^{(t)}\right]\mathbb{E}_{M}\left[M_{i}^{(t)}(\ell)\right] \label{indep_ex}\\
&= (1-\theta)p= p^{\prime}
\end{align}
% where \eqref{indep_ex} follows from the independence of these two Bernoulli random variables. 
and,
\begin{align}
\mathbb{E}_{D,M}\left[\left(M_{D,i}^{(t)}(\ell)\right)^{2}\right]&=
\mathbb{E}_{D,M}\left[\left(D_{i}^{(t)}\right)^{2}\left(M_{i}^{(t)}(\ell)\right)^{2}\right]\\
%&=\mathbb{E}_{D}\left[\left(D_{i}^{(t)}\right)^{2}\right]\mathbb{E}_{M}\left[\left(M_{i}^{(t)}(\ell)\right)^{2}\right]\\
&= (1-\theta)p= p^{\prime}
\end{align}
$\forall i \in [N]$, and we define, 
\begin{align}
\Tilde{p} &= \mathbb{E}_{M,D}\left[M_{D,i}^{(t)}(\ell)  
 M_{D,j}^{(t)}(\ell)\right]\\
%  &=\mathbb{E}_{D}\left[D_{i}^{(t)}  
%  D_{j}^{(t)}\right]\mathbb{E}_{M}\left[M_{i}^{(t)}(\ell)  
%  M_{j}^{(t)}(\ell)\right]\\
 &=\mathbb{E}_{D}\left[D_{i}^{(t)}\right]  
\mathbb{E}_{D}\left[D_{j}^{(t)}\right]\mathbb{E}_{M}\left[M_{i}^{(t)}(\ell)M_{j}^{(t)}(\ell)\right] \label{indep_dropout}\\
&=(1-\theta)^2\mathbb{E}_{M}\left[M_{i}^{(t)}(\ell)  
 M_{j}^{(t)}(\ell)\right]
 \end{align}
which represents the probability that both user $i$ and $j$ (where $i \neq j$) participate in the aggregation phase at location $\ell$. 
%where \eqref{indep_dropout} follows from the fact that dropout event of each user is independent of other users. Now,
Then, 
\begin{align}
&\mathbb{E}_{M}\left[M_{i}^{(t)}(\ell)  
 M_{j}^{(t)}(\ell)\right] \notag\\
 %&=P\left[M_{i}^{(t)}(\ell)=1, M_{j}^{(t)}(\ell)=1\right]\\
 &=P\left[M_{i}^{(t)}(\ell)=1, M_{j}^{(t)}(\ell)=1|\mathbf{b}_{ij}(\ell)=0\right]P\left[\mathbf{b}_{ij}(\ell)=0\right] \notag\\
 &+P\left[M_{i}^{(t)}(\ell)=1, M_{j}^{(t)}(\ell)=1|\mathbf{b}_{ij}(\ell)=1\right]P\left[\mathbf{b}_{ij}(\ell)=1\right] \notag\\
 &=P\left[M_{i}^{(t)}(\ell)=1\right] P\left[M_{j}^{(t)}(\ell)=1\right]\left(1-\frac{\alpha}{N-1}\right) \notag \\
% &+P\left[M_{i}^{(t)}(\ell)=1, M_{j}^{(t)}(\ell)=1|\mathbf{b}_{ij}(\ell)=1\right]\frac{\alpha}{N-1} \label{indep}\\
&=\left(1-\left(1-\frac{\alpha}{N-1}\right)^{N-2}\right)^{2}\left(1-\frac{\alpha}{N-1}\right)+1 \times \frac{\alpha}{N-1}\\ 
%&=\left(1-2\left(1-\frac{\alpha}{N-1}\right)^{N-2}+\left(1-\frac{\alpha}{N-1}\right)^{2N-4}\right)\notag\\
%& \hspace{3.5cm} \times \left(1-\frac{\alpha}{N-1}\right)+1 \times \frac{\alpha}{N-1}\\
&= 1-2\left(1-\frac{\alpha}{N-1}\right)^{N-1}+\left(1-\frac{\alpha}{N-1}\right)^{2N-3}
\end{align}
% where \eqref{indep} follows from the independence of those two users' participation events when $\mathbf{b}_{ij}=0$.
Since,
\begin{align}
p^2 %&=\left(1-\left(1-\frac{\alpha}{N-1}\right)^{N-1}\right)^2\\
&=1-2\left(1-\frac{\alpha}{N-1}\right)^{N-1}+\left(1-\frac{\alpha}{N-1}\right)^{2N-2}
\end{align}
it follows that $(p^{\prime})^2 \leq \Tilde{p}$, and hence,
\begin{align}
\beta_{i}\beta_{j} \left(\frac{\Tilde{p}}{(p^{\prime})^2}  - 1\right) \geq 0
\end{align}
from which \eqref{Validate} follows. Finally, \eqref{line_b4_AM-GM} follows  from the AM-GM inequality. 
Next,
\begin{align}
&\mathbb{E}\left[\sum_{\ell=1}^{d}\left|\sum_{i=1}^{N}\frac{\beta_{i}M_{i}^{(t)}(\ell)}{p^{\prime}} \sum_{k=t_o}^{t}\mathbf{z}_{i}^{(k)}(\ell)-\sum_{i=1}^{N}\beta_{i} \sum_{k=t_o}^{t}\mathbf{z}_{i}^{(k)}(\ell)\right|^2\right] \notag\\
&\leq \mathbb{E}_{\xi}\left[\sum_{\ell=1}^{d}\sum_{i=1}^{N}\left(\beta_{i}^2 \left(\sum_{k=t_o}^{t}\mathbf{z}_{i}^{(k)}(\ell)\right)^2 \left(\frac{1}{p^{\prime}}  -1\right) \notag \right. \right. \\
& + \sum_{j=1,j \neq i}^{N}\beta_{i}\beta_{j}\left. \left.  \left(\frac{\Tilde{p}}{(p^{\prime})^2}  - 1\right) \frac{\left(\sum_{k=t_o}^{t} \mathbf{z}_{i}^{(k)}(\ell)\right)^2 + \left(\sum_{k=t_o}^{t} \mathbf{z}_{j}^{(k)}(\ell)\right)^2}{2}\right) \right] \notag \\
%&\leq \mathbb{E}_{\xi} \left[  \sum_{\ell=1}^{d}\sum_{i=1}^{N}\left(\beta_{i}^2\left\|\sum_{k=t_o}^{t}\mathbf{z}_{i}^{(k)}(\ell)\right\|^2 \left( \frac{1}{p^2}p  + 1 - 2 \frac{1}{p} p \right) \notag \right. \right.\\
%&  + \sum_{j=1,j \neq i}^{N} \left( \frac{\beta_{i}\beta_{j}}{p^2} \frac{p \left\|\sum_{k=t_o}^{t}\mathbf{z}_{i}^{(k)}(\ell)\right\|^2 + p \left\|\sum_{k=t_o}^{t}\mathbf{z}_{j}^{(k)}(\ell)\right\|^2}{2} \notag \right. \\ 
%&  \left. \left. \left.-  \beta_{i}\beta_{j}\left(\frac{p}{p}  + \frac{p}{p}-1\right)\frac{\left\|\sum_{k=t_o}^{t} \mathbf{z}_{i}^{(k)}(\ell)\right\|^2 + \left\|\sum_{k=t_o}^{t} \mathbf{z}_{j}^{(k)}(\ell)\right\|^2}{2}  \right) \right) \right] \label{expectation}\\
%&= \mathbb{E}_{\xi} \left[ \sum_{\ell=1}^{d}\sum_{i=1}^{N} \left( \beta_{i}^2\left\|\sum_{k=t_o}^{t}\mathbf{z}_{i}^{(k)}(\ell)\right\|^2 \left(\frac{1}{p}-1\right)   \notag \right. \right.  \\
%&\left. \left. +\sum_{j=1,j \neq i}^{N} \beta_{i}\beta_{j} \frac{\left\|\sum_{k=t_o}^{t}\mathbf{z}_{i}^{(k)}(\ell)\right\|^2 + \left\|\sum_{k=t_o}^{t}\mathbf{z}_{j}^{(k)}(\ell)\right\|^2}{2}\left(\frac{1}{p}-1  \right)  \right)  \right]\\
&= \mathbb{E}_{\xi} \left[ \sum_{i=1}^{N} \left( \beta_{i}^2\left\|\sum_{k=t_o}^{t}\mathbf{z}_{i}^{(k)}\right\|^2 \left(\frac{1}{p^{\prime}}-1\right) \right. \right. \left. \left.  + \sum_{j=1,j \neq i}^{N}  \beta_{i}\beta_{j}\frac{\left\|\sum_{k=t_o}^{t}\mathbf{z}_{i}^{(k)}\right\|^2 + \left\|\sum_{k=t_o}^{t}\mathbf{z}_{j}^{(k)}\right\|^2}{2}\left(\frac{\Tilde{p}}{(p^{\prime})^2}-1  \right)  \right) \right] \notag \\ 
& \leq \sum_{i=1}^{N} \left( \beta_{i}^2  \mathbb{E}_{\xi} \left[\left\|\eta^{(t_o)}\sum_{k=t_o}^{t}\nabla F_{i} \left(\mathbf{w}_{i}^{(k)}, \xi_{i}^{(k)}\right)\right\|^2\right]\left(\frac{1}{p^{\prime}}-1\right) \notag  \right. \\
&  + \hspace{-0.2cm}\sum_{j=1,j \neq i}^{N} \! \left( \frac{\beta_{i}\beta_{j}}{2}\left(\frac{\Tilde{p}}{(p^{\prime})^2}\!-\!1\right) \left(\mathbb{E}_{\xi} \left\|\eta^{(t_o)}\sum_{k=t_o}^{t}\nabla F_{i} \left(\mathbf{w}_{i}^{(k)},  \! \left. \left. \xi_{i}^{(k)}\right)\right\|^2 \right. \right. \right. \right. \left. \left. \left. + \mathbb{E}_{\xi} \left\|\eta^{(t_o)}\sum_{k=t_o}^{t}\nabla F_{j} \left(\mathbf{w}_{j}^{(k)}, \xi_{j}^{(k)}\right)\right\|^2\right)\right) \right) \label{learning rate}\\
& \leq \!\left(\eta^{(t_o)}\right)^{2}\sum_{i=1}^{N} \left( \beta_{i}^2 \left(\frac{1}{p^{\prime}}-1\right)  E\sum_{k=t_o}^{t}\mathbb{E}_{\xi} \left[\left\|\nabla F_{i} \left(\mathbf{w}_{i}^{(k)}, \xi_{i}^{(k)}\right)\right\|^2\right]  \notag  \right. \\
&  + \sum_{j=1,j \neq i}^{N} \left( \frac{\beta_{i}\beta_{j}}{2}\left(\frac{\Tilde{p}}{(p^{\prime})^2}-1\right) E \left(\sum_{k=t_o}^{t}\mathbb{E}_{\xi} \left\|\nabla F_{i} \left(\mathbf{w}_{i}^{(k)}, \xi_{i}^{(k)}\right)\right\|^2  \right. \right.  \left. \left. \left.  + \sum_{k=t_o}^{t}\mathbb{E}_{\xi} \left[\left\|\nabla F_{j} \left(\mathbf{w}_{j}^{(k)}, \xi_{j}^{(k)}\right)\right\|^2 \right]\right) \right) \right) \label{rule}\\
% & \leq \leq 4\left(\eta^{(t)}\right)^{2}E^2G^2\sum_{i=1}^{N} \left( \beta_{i}^2 \left(\frac{1}{p^{\prime}}-1\right) \notag \right.\\
% &\left. \hspace{3.5cm}+ \sum_{j=1,j \neq i}^{N}  \beta_{i}\beta_{j}\left(\frac{\Tilde{p}}{(p^{\prime})^2}-1\right) \right) \label{assumption 4}\\
& \leq 4\left(\eta^{(t)}\right)^{2}\!\!E^2G^2\sum_{i=1}^{N} \left( \beta_{i}^2 \left(\frac{1}{p^{\prime}}-1\right)  \right. \left. +\hspace{-0.2cm} \sum_{j=1,j \neq i}^{N}  \hspace{-0.2cm} \beta_{i}\beta_{j}\left(\frac{\Tilde{p}}{(p^{\prime})^2}-1\right) \right) \label{assumption 4}
%&= \left(\frac{1}{p}-1\right)4\left(\eta^{(t)}\right)^{2}E^2G^2 \label{weight_sum}
\end{align}
%  where in \eqref{norm}, we have used: \begin{equation}
% \left\|\sum_{k=t_o}^{t} \mathbf{z}_{i}^{(k)}\right\|^2=\sum_{\ell=1}^{d}\left|\sum_{k=t_o}^{t} \mathbf{z}_{i}^{(k)}(\ell)\right|^2
% \end{equation}
where \eqref{learning rate} follows from  $\eta^{(t_o)} \geq \eta^{(k)}$  $\forall k \geq t_o$ and  \eqref{rule} holds since $\left\|\sum_{i=1}^{s} \mathbf{a}_{i}\right\|^{2} \leq s \sum_{i=1}^{s}\left\|\mathbf{a}_{i}\right\|^{2}$ for any $\mathbf{a} \in \mathbb{R}^d$ \cite{Stich2019localsgd}. 
% \begin{equation}
% \left\|\sum_{i=1}^{s} \mathbf{a}_{i}\right\|^{2} \leq s \sum_{i=1}^{s}\left\|\mathbf{a}_{i}\right\|^{2}
% \end{equation}
% for any vector $\mathbf{a} \in \mathbb{R}^d$ as mentioned in \cite{Stich2019localsgd}. 
Finally,  \eqref{assumption 4} follows from \eqref{eqassumpt:4}. 
% Assumption 4.
%and \eqref{weight_sum} follows from the fact that 
%\begin{equation}
%\sum_{i=1}^{N} \left\{ \beta_{i}^2  + \sum_{j=1,j \neq i}^{N}  \beta_{i}\beta_{j} \right\}=1
%\end{equation}
\end{proof} 
Now we can proceed with the convergence proof. Note that,
\begin{align}
\left\|\overline{\mathbf{w}}^{(t+1)}-\mathbf{w}^{*}\right\|^{2} &=\left\|\overline{\mathbf{w}}^{(t+1)}-\overline{\mathbf{v}}^{(t+1)}+\overline{\mathbf{v}}^{(t+1)}-\mathbf{w}^{*}\right\|^{2} \\
&=\left\|\overline{\mathbf{w}}^{(t+1)}-\overline{\mathbf{v}}^{(t+1)}\right\|^{2} +\left\|\overline{\mathbf{v}}^{(t+1)}-\mathbf{w}^{*}\right\|^{2} +2\left\langle\overline{\mathbf{w}}^{(t+1)}-\overline{\mathbf{v}}^{(t+1)}, \overline{\mathbf{v}}^{(t+1)}-\mathbf{w}^{*}\right\rangle \label{terms}
\end{align}
The last term on the right hand side of \eqref{terms} vanishes in expectation due to  Lemma \ref{lemma 1}. 
The remainder of the proof follows standard induction steps such as in  \cite{li2019convergence}. When $t+1 \notin \tau_{E}, \overline{\mathbf{w}}_{t+1}=\overline{\mathbf{v}}_{t+1}$, whereas when $t+1 \in \tau_{E}$,
\begin{align}
\mathbb{E}\left\|\overline{\mathbf{w}}^{(t+1)}-\mathbf{w}^{*}\right\|^{2} &=\mathbb{E}\left\|\overline{\mathbf{w}}^{(t+1)}-\overline{\mathbf{v}}^{(t+1)}\right\|^{2}+\mathbb{E}\left\|\overline{\mathbf{v}}^{(t+1)}-\mathbf{w}^{*}\right\|^{2} \\
& \leq\left(1-\eta^{(t)} \mu\right) \mathbb{E}\left\|\overline{\mathbf{w}}^{(t)}-\mathbf{w}^{\star}\right\|^{2}+\left(\eta^{(t)}\right)^{2}(B+C)
\end{align}
where the last inequality follows from \cite{li2019convergence}. 
%Where $C=\frac{d}{4c^2}+\sum_{i=1}^{N} \left\{ \beta_{i}^2  + \sum_{j=1,j \neq i}^{N}  \beta_{i}\beta_{j} \right\}\left(\frac{1}{p}-1\right)4E^2G^2$.
Thus, from the definition of strong convexity and by following the steps of \cite{li2019convergence}, one can show that:
\begin{equation}
\mathbb{E}\left[F\left(\overline{\mathbf{w}}^{(J)}\right)\right]-F^{*} \leq \frac{2 \frac{L}{\mu}}{\nu+J}\left(\frac{B+C}{\mu}+2 L\left\|\mathbf{w}^{(0)}-\mathbf{w}^{*}\right\|^{2}\right)
\end{equation}
which concludes the proof.

\end{document}